\documentclass[10pt,twocolumn,letterpaper]{article}

\usepackage[pagenumbers]{cvpr} 
\usepackage{xcolor}
\usepackage{amsmath}

\usepackage{bm}
\usepackage{amsthm}
\usepackage{amsmath}
\usepackage{algorithmic}
\usepackage[linesnumbered,ruled,vlined]{algorithm2e}
\usepackage{graphicx}
\usepackage{multirow} 
\usepackage{makecell}
\usepackage{tabularx}
\theoremstyle{definition}

\newtheorem{theorem}{Theorem}[section]

\usepackage{colortbl}
\usepackage{tabularx}
\newcommand{\bmmc}[1]{\bm{\mathcal{#1}}}
\newcommand{\comments}[1]{\textcolor[RGB]{0,150,0}{#1}}

\usepackage{amssymb}  
\usepackage{pifont}  

\newcommand{\cmark}{\ding{51}}  
\newcommand{\xmark}{\ding{55}}

\definecolor{cvprblue}{rgb}{0.21,0.49,0.74}
\usepackage[pagebackref,breaklinks,colorlinks,allcolors=cvprblue]
{hyperref}

\def\paperID{2263} 
\def\confName{CVPR}
\def\confYear{2025}
\title{COAP: Memory-Efficient Training with Correlation-Aware Gradient Projection}

\author{%
  Jinqi Xiao$^{1,2,*}$, Shen Sang$^{1}$, Tiancheng Zhi$^{1}$, Jing Liu$^{1}$,   \\ Qing Yan$^{1}$,  Yuqian Zhang$^{2}$, Linjie Luo$^{1}$, Bo Yuan$^{2}$
 \\
 \\
  $^{1}$ ByteDance Inc., $^{2}$ Rutgers University
  }
\begin{document}
\maketitle
\footnotetext[0]{$^{*}$Work done during internship at ByteDance.}

\begin{abstract}

Training large-scale neural networks in vision, and multimodal domains demands substantial memory resources, primarily due to the storage of optimizer states. While LoRA, a popular parameter-efficient method, reduces memory usage, it often suffers from suboptimal performance due to the constraints of low-rank updates. Low-rank gradient projection methods (e.g., GaLore, Flora) reduce optimizer memory by projecting gradients and moment estimates into low-rank spaces via singular value decomposition or random projection. However, they fail to account for inter-projection correlation, causing performance degradation, and their projection strategies often incur high computational costs. In this paper, we present \textbf{COAP} (\underline{CO}rrelation-\underline{A}ware Gradient \underline{P}rojection), a memory-efficient method that minimizes computational overhead while maintaining training performance. Evaluated across various vision, language, and multimodal tasks, COAP outperforms existing methods in both training speed and model performance. For LLaMA-1B, it reduces optimizer memory by 61\% with only 2\% additional time cost, achieving the same PPL as AdamW. With 8-bit quantization, COAP cuts optimizer memory by 81\% and achieves 4x speedup over GaLore for LLaVA-v1.5-7B fine-tuning, while delivering higher accuracy. \url{https://byteaigc.github.io/coap/}
\end{abstract}    
\section{Introduction}

Deep neural networks have achieved remarkable success across vision~\cite{dosovitskiy2020image,rombach2022high, podell2023sdxl,peebles2023scalable,blackforestlabs2024flux}, language~\cite{achiam2023gpt,touvron2023llama, touvron2023llama-2, dubey2024llama}, and multi-modality domains~\cite{driess2023palm, team2023gemini, liu2024visual, liu2024improved}, driven by the increasing scale of these models. While scaling up model size significantly contributes to these advancements~\cite{kaplan2020scaling}, it also leads to considerable memory constraints, particularly due to the optimizer states. For instance, training an LLaVA-7B~\cite{liu2024improved} model with the Adam~\cite{kingma2014adam} optimizer at BF16 numerical format requires at least 63.8GB of GPU memory, with optimizer states consuming 40\% of the total, while the model and gradients each take up 20\%. Combining engineering and system efforts, \emph{e.g.}, activation checkpointing~\cite{chen2016training} and LOMO~\cite{lv2023full} can significantly reduce the memory usage of activations and gradients. This makes optimizer a critical bottleneck that needs to be addressed. To tackle this issue, low-rank training has proven effective in minimizing the memory footprint of optimizer states by performing training within a low-rank subspace~\cite{zhao2024galore,hao2024flora,hu2021lora}.

\begin{figure}[t]
    \centering
    \includegraphics[width=\linewidth]{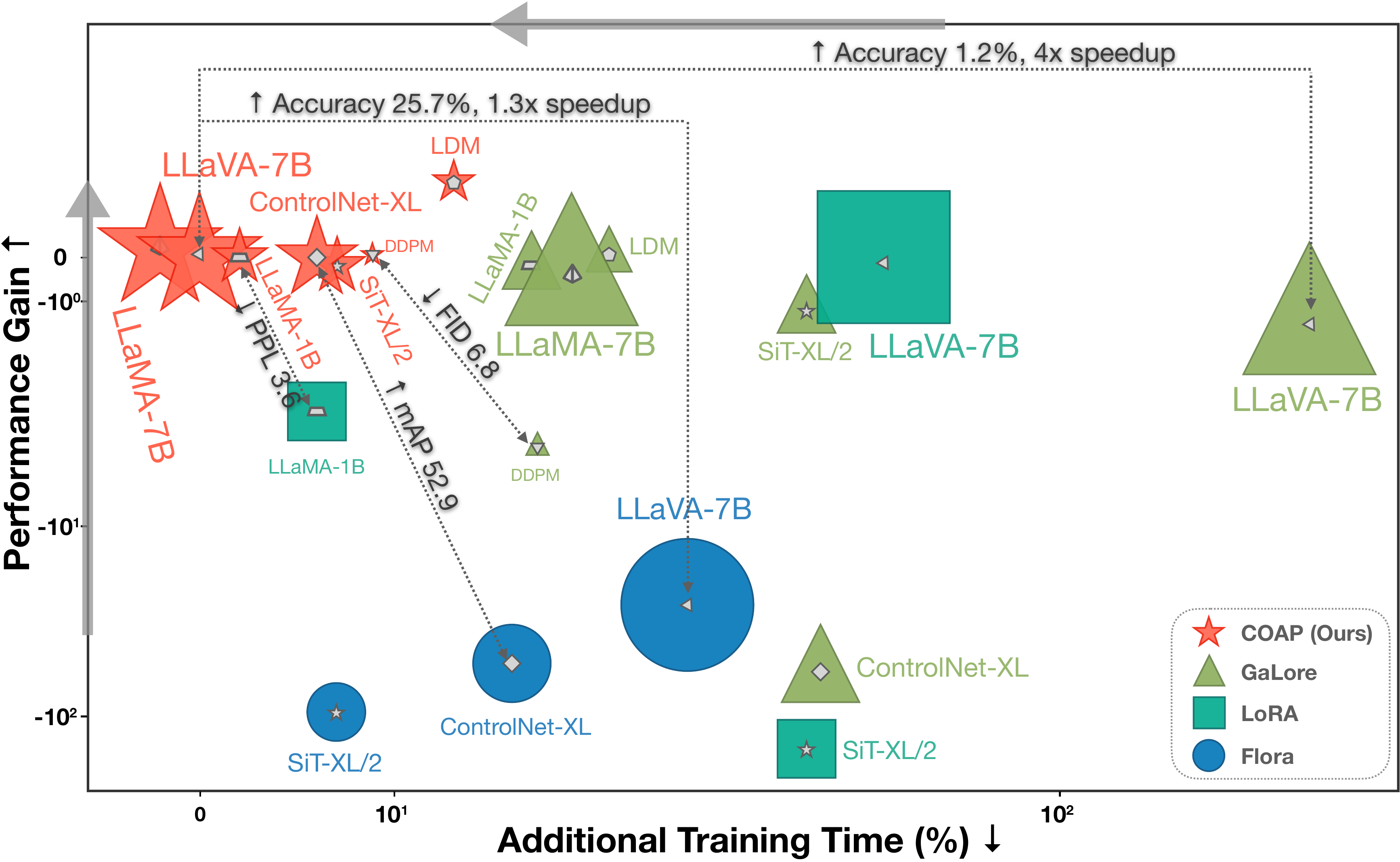}
    \vspace{-6mm}
    \caption{Comparison between COAP and other low-rank-based methods. The X-axis shows additional training time, with lower values being better. The Y-axis shows performance (\emph{e.g.}, FID, PPL) changes compared to the original optimizer (\emph{e.g.}, Adam~\cite{kingma2014adam}, Adafactor~\cite{shazeer2018adafactor}), with higher values indicating better performance. }
    \label{fig:overall_performance}
    \vspace{-5mm}
\end{figure}

Current low-rank training approaches can generally be categorized into two main types: parameter-efficient fine-tuning (PEFT) and training with low-rank gradients. The most common method in PEFT is Low-Rank Adaptation (LoRA)~\cite{hu2021lora}, which adds trainable low-rank matrices to a frozen pre-trained model's weights, allowing for efficient model adaptation with significantly fewer parameters than full fine-tuning. While low-rank updates can reduce memory consumption, extensive empirical evidence~\cite{xia2024chain,zhang2023adalora,liu2024dora,wang2024lora,ding2023parameter,biderman2024lora} shows that the low-rank constraint diminishes the model's representation capacity, resulting in sub-optimal performance compared to full fine-tuning. In addition, its pre-training alternative ReLoRA~\cite{lialin2023relora}, which periodically updates the model using previously learned low-rank adapters, still requires a full-rank warm-up phase, thereby negating its memory-saving benefits. For training with low-rank gradient~\cite{zhao2024galore, hao2024flora}, these methods leverage the inherent low-rank structure of the gradients rather than constraining model parameters to a low-rank space. Projecting gradients into low-rank space can significantly reduce memory usage during optimization, making it feasible for scenarios with limited memory resources or extremely large models. For example, GaLore~\cite{zhao2024galore} periodically applies Singular Value Decomposition (SVD) to decompose gradients, creating a fixed low-rank projection matrix for subsequent steps. However, the high computational cost of SVD can severely slow down training, especially for models with large weight matrices. In contrast, Flora~\cite{hao2024flora} resamples a random projection matrix at each iteration to address the low-rank limitation of LoRA. Despite this, generating new random projection matrices at each iteration still incurs computational overhead, particularly for large models. Besides the significant time costs required to achieve memory savings, these methods also face the challenge of lacking inter-projection correlation. When the projection is determined solely by a single step’s gradient or random sampling, it may disrupt the continuity of the optimization trajectory, leading to abrupt shifts and adversely affecting training dynamics.

To address these challenges, we propose a memory-efficient training method, \textbf{COAP} (\underline{CO}rrelation-\underline{A}ware Gradient \underline{P}rojection), that minimizes additional computational overhead while retaining model performance (see Fig.~\ref{fig:overall_performance}). \underline{1)} Unlike methods that rely solely on a single step's gradient or random generation to determine the low-rank projection matrix, COAP introduces an \emph{inter-projection correlation-aware update strategy}, by solving an optimization problem with gradient descent. It tracks the previous low-rank projection matrix and combines the direction of the first-order moments to continuously update the low-rank projection matrix, thereby avoiding abrupt shifts in training dynamics. \underline{2)} Additionally, we propose to use \emph{occasional low-cost SVD} to recalibrate the low-rank projection matrices, preventing the gradient descent-based updates from getting trapped in local optima. Through the implementation of the criterion that prioritizes gradient reconstruction capability and training direction smoothness, combined with occasional low-cost SVD to continuously update the low-rank projection matrix, we achieve a significant reduction in the computational cost of vanilla SVD in large-scale models while simultaneously enhancing model performance.

We demonstrate the effectiveness of our method in both pre-training and fine-tuning across various vision, language, and multi-modal tasks. For pre-training, our method matches AdamW's performance while reducing optimizer memory by 61\% for LLaMA-1B~\cite{touvron2023llama}, with only 2\% increase in time overhead. In 8-bit optimizers, our approach outperforms 8-bit Adam and GaLore in both training speed and performance on LLaMA-7B. For fine-tuning LLaVA-v1.5-7B~\cite{liu2024improved}, our method with 8-bit optimization reduces optimizer memory by 81\%, achieves 4$\times$ speedup over GaLore, and increases accuracy by 1.2\%.

\section{Related Works}

\noindent\textbf{Momentum-based Optimizers}, for instance Adam~\cite{kingma2014adam}, AdamW~\cite{loshchilov2017decoupled} and Adadelta~\cite{zeiler2012adadelta}, have largely replaced traditional SGD due to their adaptiveness and faster convergence. However, this adaptiveness increases memory usage, as they require storing first and second moment estimates, effectively tripling the memory footprint. Adafactor~\cite{shazeer2018adafactor} addresses this by employing low-rank approximations for the second moments, reducing memory requirements while retaining adaptive learning. In this work, we explore projecting both gradients and moments into a low-rank space during training, enabling efficient large-scale model training with commonly-used momentum-based optimizers.

\noindent\textbf{Low-rank Adaptation} (LoRA)~\cite{hu2021lora} enables parameter-efficient fine-tuning by applying low-rank updates to pre-trained models, reducing memory and computation costs. DoRA~\cite{liu2024dora} improves this by decoupling weights into magnitude and direction, using low-rank updates only for directional adjustments. ReLoRA~\cite{lialin2023relora} adapts LoRA to pre-training but requires a full-rank warm-up, increasing memory usage with model size. Variants such as LyCORIS~\cite{yeh2024navigating}, AdaLoRA~\cite{zhang2023adalora}, HALOC~\cite{xiao2023haloc} and COMCAT~\cite{pmlr-v202-xiao23e} step further in efficiency and flexibility. However, these methods often result in suboptimal performance~\cite{xia2024chain} due to the low-rank limit. Conversely, our approach preserves full-rank learning while leveraging low-rank gradient structures, supporting both pre-training and fine-tuning without compromising performance.

\noindent\textbf{Training with Low-rank Gradient}, inspired by projected gradient descent methods~\cite{chen2015fast, chen2019non}, aims to reduce the memory usage of optimizers during training by applying low-rank gradient projections. GaLore~\cite{zhao2024galore} computes projection matrices via per-layer gradient decomposition using inefficient SVD, In contrast, Flora~\cite{hao2024flora} proposes to generate projection matrices randomly on the fly to eliminate this overhead. However, a major limitation of these methods is the lack of inter-projection correlation. As the optimizer transitions to a new subspace, the projection matrix is either based on the current gradient or random selection, leading to abrupt shifts. In this work, we address this issue by incorporating prior optimization directions, ensuring smoother transitions in the projection space. This approach improves training dynamics and enhances overall optimization performance.

\noindent\textbf{Memory-efficient Training Techniques} are widely employed to reduce the overall memory footprint in large-scale model training, including activation checkpointing~\cite{chen2016training}, 8-bit optimizers~\cite{dettmers20218}, and mixed-precision training~\cite{micikevicius2017mixed}, zeroth-order optimization~\cite{balasubramanian2022zeroth,zhang2022robustify,chen2023deepzero,zhang2024revisiting}, etc. The popular approach, ZeRO~\cite{rajbhandari2020zero}, optimizes memory usage in multi-GPU settings. ZeRO contains three stages, progressively sharding optimizer states, gradients, and model weights across devices, synchronizing only when necessary to minimize memory usage. Our method specifically targets memory reduction for optimizer states and can be effectively combined with these techniques, offering additional memory savings and further improving training efficiency.
\section{Method}
\label{sec:method}
\subsection{Background}
\label{subsec:background}

\textbf{Notation.} We use the following notation conventions: matrices and vectors are indicated with boldface capital and lowercase letters, such as $\bm{X}$ and $\bm{x}$, respectively. Non-boldface letters with indices, \emph{e.g.}, $X(i,j)$, and $x(i)$, denote the entries of a matrix $\bm{X}$, and a vector $\bm{x}$, respectively.

\textbf{Full-Rank Training.} 
For a time-varying weight matrix $\bm{W} \in \mathbb{R}^{m \times n}$ with objective function as $\mathcal{L}(\bm{W})$, the corresponding gradient matrix at time step $t$ can be denoted as $\bm{G}_t = \nabla_{\bm{W}} \mathcal{L}_t(\bm{W}_t) \in \mathbb{R}^{m \times n}$. Then, the general weight update process can be formulated as:
\begin{equation}
    \bm{W}_{t+1} = \bm{W}_t - \eta \rho_t(\bm{G}_t),
\end{equation}
where $\eta$ is the learning rate, and $\rho_t$ is the optimizer-dependent gradient regularizer that adaptively adjusts raw gradients $\bm{G}_t$ via incorporating additional optimizer states. 
For instance, Adam~\cite{kingma2014adam} uses bias-corrected first and second moment $\bm{M}_t$ and $\bm{V}_t$ to regularize the gradients as follows:
\begin{equation}
\begin{aligned}
    &\bm{M}_t = \beta_1 \bm{M}_{t-1} + (1 - \beta_1) \bm{G}_t, \\
    &\bm{V}_t = \beta_2 \bm{V}_{t-1} + (1 - \beta_2) \bm{G}_t^2, \\
    &\rho_t(\bm{G}_t) = \frac{\bm{M}_t/(1-\beta_1^t)}{\sqrt{\bm{V}_t/(1-\beta_2^t)} + \epsilon},
\end{aligned}
\label{eqn:adam}
\end{equation}
where $\beta_1$ and $\beta_2$ are hyper-parameters, $\epsilon$ is a small constant to ensure numerical stability, and the matrix operations are element-wise. Considering Adam requires extra $2mn$ memory to store $\bm{M}_t$ and $\bm{V}_t$, Adafactor~\cite{shazeer2018adafactor} proposes 
to use factorization to estimate $\bm{V}_t$ for higher memory efficiency:
\begin{equation}
\begin{aligned}
    &\bm{R}_t = \beta_2 \bm{R}_{t-1} + (1-\beta_2)\cdot \mathrm{Sum}(\bm{G}_t^2, 1), \\
    &\bm{C}_t = \beta_2 \bm{C}_{t-1} + (1-\beta_2)\cdot  \mathrm{Sum}(\bm{G}_t^2, 0), \\
    &\hat{\bm{V}_t} = \sqrt{\frac{\mathrm{Mean}(\bm{R}_t, -1)}{\bm{R}_t\bm{C}_t}} , \\
\end{aligned}
\label{eqn:adafac}
\end{equation}
where $\mathrm{Sum}(\cdot)$ returns the sum of each row of the input matrix in the given dimension ($0$ or $1$). Here by using $\bm{R}_t \in \mathbb{R}^{m \times 1}$ and $\bm{C}_t \in \mathbb{R}^{1 \times n}$ to estimate $\bm{V}_t$ as $\hat{\bm{V}_t} \in \mathbb{R}^{m \times n}$, the memory consumption for storing second moments is reduced from $mn$ to $m+n$. 

\textbf{Training with Low-rank Weight Update.} Due to the high cost of updating the entire full-rank $ \bm{W}_t$, LoRA~\cite{hu2021lora} and its variants ~\cite{zhang2023adalora, lialin2023relora}, as the memory-efficient solutions that only adjust the change of $ \bm{W}_t$ in the low-rank subspace, are popularly used in practice as follows:
\begin{equation}
    \bm{W}_t = \bm{W}_0 + \bm{B}_t \bm{A}_t,
\end{equation}
where $ \bm{W}_0 \in \mathbb{R}^{m \times n}$ is the fixed pre-trained or initialized weights, and $\bm{A}_t \in \mathbb{R}^{r \times n}$ and $\bm{B}_t \in \mathbb{R}^{m \times r}$ are the learnable low-rank adapters with $r \ll \min(m, n)$.

\textbf{Training with Low-rank Gradient.} Motivated by the observations that 1) updating weight in the low-rank subspace may limit the representation capacity, and 2) gradient matrix tends to exhibit a certain degree of low-rankness during training, recent works~\cite{zhao2024galore, hao2024flora} propose to explore memory-efficient training via projecting gradient into low-rank format. For instance, given a projection matrix $\bm{P}_t$ that defines a low-rank subspace, the \textit{projected} gradient, first order moments and second order moments in Adam optimizer are calculated as  $\bm{G}_t^{\rm proj}= \bm{G}_t\bm{P}_t$, $\bm{M}_t^{\rm proj} = \beta_1\bm{M}_{t-1}^{\rm proj}+(1-\beta_1)\bm{G}_t^{\rm proj}$ and $\bm{V}_t^{\rm proj} = \beta_2\bm{V}_{t-1}^{\rm proj}+(1-\beta_2)(\bm{G}_t^{\rm proj})^2$, respectively. Then, the weight update is performed as follows:
\begin{equation}
\label{eq:galore}
\begin{aligned}
&\bm{W}_{t+1} = \bm{W}_t - \eta \rho_t(\bm{G}^{\rm proj}_t), \\
&\rho_t(\bm{G}^{\rm proj}_t)=\frac{\bm{M}_t^{\rm proj}/(1-\beta_1^t)}{\sqrt{\bm{V}_t^{\rm proj}/(1-\beta_2^t)}+\epsilon } \bm{P}_t^\top.
\end{aligned}
\end{equation}
Here, $\bm{W}_t$, $\bm{G}_t$, $\rho_t(\bm{G}^{\rm proj}_t)$ $\in \mathbb{R}^{m \times n}~(m \geq n)$, $\bm{P}_t \in \mathbb{R}^{n \times r}$, and $\bm{G}_t^{\rm proj}$, $\bm{M}_t^{\rm proj}$, $\bm{V}_t^{\rm proj}$ $\in \mathbb{R}^{m \times r}$.

\vspace{1mm}
\subsection{Challenge Analysis of Gradient Projection}
As discussed in Section \ref{subsec:background} and demonstrated in~\cite{zhao2024galore, hao2024flora}, projecting gradients onto low-rank subspace enables efficient memory-saving training while maintaining the essential full-rank structure of model weights, which is crucial for preserving the model's expressive capacity and avoiding information loss. However, despite these successes, we argue that state-of-the-art gradient projection methods still face some fundamental challenges as follows.


\textbf{1) Lack of Inter-Projection Correlation Awareness.} As highlighted in~\cite{zhao2024galore,hao2024flora}, weight updates should not be restricted to a single low-rank subspace; therefore, periodically switching the projection subspace (i.e., recalculating $\bm{P}_t$) is essential in the training process. Specifically, GaLore  \cite{zhao2024galore} performs SVD on $\bm{G}_t$ every set period of batches, selecting either the truncated left or right singular vector matrix as $\bm{P}_t$. In contrast, Flora~\cite{hao2024flora} proposes to directly use random matrix as $\bm{P}_t$ for gradient projection at each batch.

\begin{figure}[t]
    \centering
    \includegraphics[width=\linewidth]{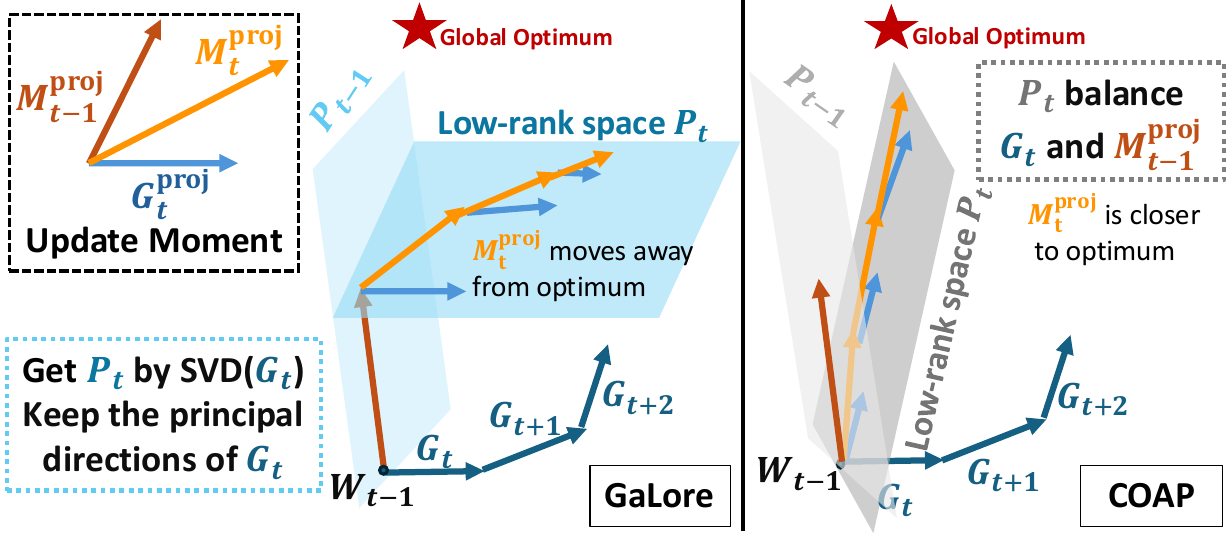}
    \vspace{-5mm}
    \caption{Comparison of optimization trajectories between COAP and GaLore. $\bm{P}_{t-1}$ represents the low-rank projection space from the previous cycle, and $\bm{P}_t$ represents the current low-rank projection space. The symbol ``$\star$'' denotes the global optimum. GaLore updates $\bm{P}_t$ based on a batch of stochastic data, which can lead to suboptimal updates if the data significantly deviates from the overall data distribution. In contrast, our method uses the more stable first-order moment $\bm{M}_{t-1}$ as guidance, mitigating this issue.}
    \label{fig:opt_traj}
    \vspace{-3mm}
\end{figure}

However, a fundamental limitation for these existing $\bm{P}_t$ update strategies is the lack of inter-projection correlation. Specifically, each time when the optimizer switches to a new low-rank subspace, the updated projection matrix $\bm{P}_t$ is determined either solely by $\bm{G}_t$ (as in GaLore) or through randomly selection (as in Flora), without leveraging directional information from previous projections. For instance, the first-order moment $\bm{M}_{t-1}^{proj}$, which reflects the moving average of projected gradients, is not incorporated into the $\bm{P}_t$ update. From an optimization trajectory perspective, this lack of continuity leads to abrupt shifts in the projection space without accounting for prior optimization direction (see Fig. \ref{fig:opt_traj}). Such discontinuities, if not properly addressed, can destabilize the training dynamics, potentially compromising model performance.  

\begin{figure}[t]
    \centering
    \includegraphics[width=\linewidth]{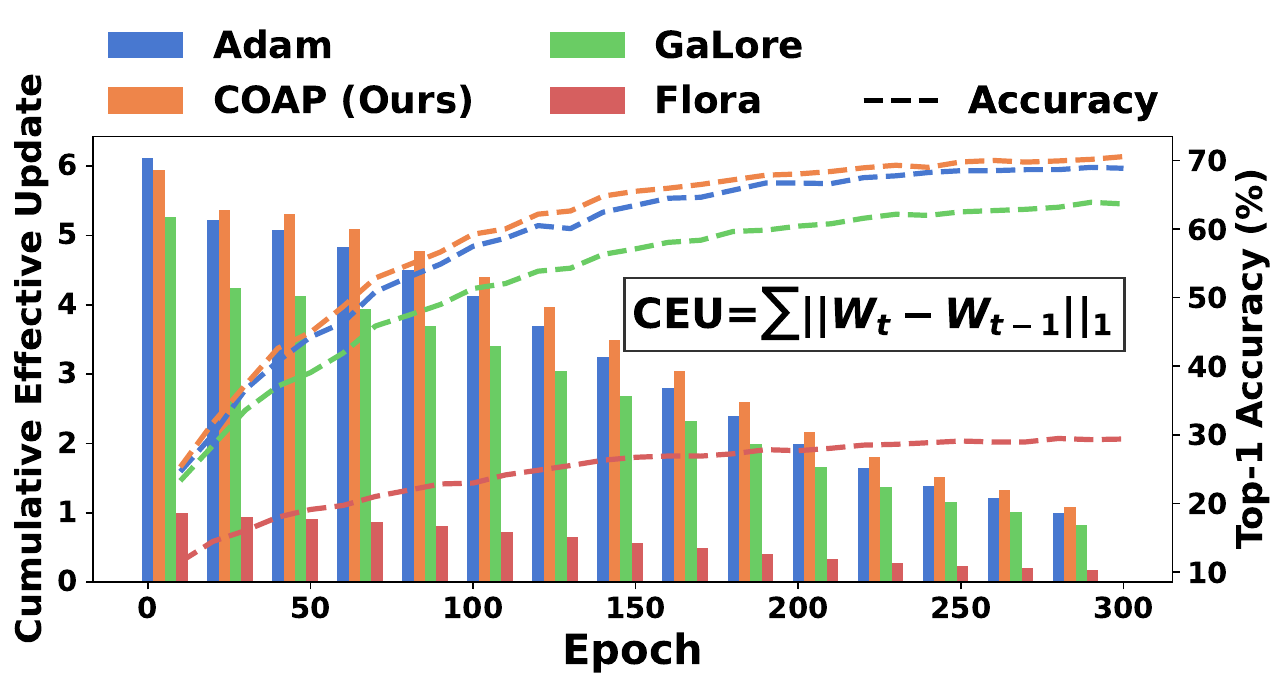}
    \vspace{-5mm}
    \caption{Cumulative effective update (CEU) and Top-1 accuracy over 300 epochs for different optimization methods on the CIFAR-100 dataset using the DeiT-Base model trained from scratch. The rank of $\bm{P}_t$ is 192, the learning rate is $5\times10^{-5}$, and all models are trained on a single A100 GPU with a batch size of 256. The magnitude of CEU indicates the extent to which the optimizer influences the model.}
    \vspace{-3mm}
    \label{fig:snr}
\end{figure}

To validate our analysis on the importance of inter-projection correlation, we perform an empirical experiment by pre-training DeiT-Base~\cite{touvron2021training} model on the CIFAR-100~\cite{krizhevsky2009learning} dataset. Fig. \ref{fig:snr} shows the \textit{cumulative effective update} (CEU) as training progresses. Here CEU, defined as $\sum\left \| W_t-W_{t-1}\right \|_1=\eta \sum\left \| \rho_t(\bm{G}^{\rm proj}_t)\right \|_1$, serves as a metric to assess optimization sufficiency, with a value close to that of the original optimizer indicating minimal loss due to the low-rank projection. From Fig. \ref{fig:snr}, it is seen that the inter-projection correlation-unaware GaLore and Flora yield a CEU that significantly deviates from that of the original optimizer, resulting in a notable drop in test accuracy. Particularly, because Flora selects $\bm{P}_t$ in a purely random way, it produces a CEU that is very different from Adam's, leading to severe performance degradation. On the other hand, our proposed inter-projection-aware approach achieves a higher CEU than the existing solutions (sometimes even surpassing that of Adam), bringing higher test accuracy. It suggests that the projection directions may have effectively eliminated noise or irrelevant information. This could potentially enable the model to converge more efficiently towards the optimal solution. Experiments in Section \ref{sec:exp} further demonstrate that our approach empirically outperforms the state-of-the-art methods across various pre-training and fine-tuning tasks for different model types.

\textbf{2) High SVD-incurred Projection Cost.} Another challenge of current projection gradient methods, especially GaLore, is the high computational overhead caused by the costly SVD\footnote{While Flora avoids the intensive SVD, it still incurs significant overhead in large-scale model training due to regenerating random projection matrices each iteration.},  which has a complexity of $O(mn^2)$ for an $m\times n$ matrix. Specifically, since each $\bm{P}_t$ update requires performing SVD on $\bm{G}_t$, this process can become prohibitively expensive, especially for large gradient matrices. For instance, when using GaLore to pre-train LLaVA-7B~\cite{liu2024improved} model under a typical configuration -- updating $\bm{P}_t$ every 200 batches with a batch size of 16 -- it takes approximately 540 seconds on a single A100 GPU to compute all $\bm{P}_t$ with a rank of 512 for each projection update. On the other hand,  the total time required for weight updates over these 200 batches is only 600 seconds. This means that calculating the $\bm{P}_t$ in GaLore incurs a $90\%$ training overhead\footnote{GaLore claims that the total computational overhead induced by SVD is negligible ($<10\%$) compared to other memory-efficient training techniques such as CPU-offload~\cite{ren2021zero}. However, this overhead cannot be ignored when compared to methods that do not use these techniques}, making it extremely computationally intensive.



\subsection{Proposed Method}

To overcome these challenges and make the low-rank gradient projection solution more practical and feasible for model training, we propose COAP, which is detailed below.

\textbf{Inter-projection Correlation-aware $\bm{P}_t$ Update.} Considering the importance of incorporating the information from the previous projection into the current update, we propose calculating $\bm{P}_t$ via solving the optimization problem:
\begin{equation}
\label{eq:obj}
\min\limits_{\bm{P}_t} \underbrace{\mathrm{MSE}(\hat{\bm{G}}_t, \bm{G}_t)}_{\text{reconstruction term}} \cdot \underbrace{(1-\mathrm{CosSim}(\hat{\bm{M}}_{t-1}, \bm{G}_t))}_{\text{direction term}},
\end{equation}
where $\mathrm{CosSim}(\cdot, \cdot)$ and  $\mathrm{MSE}(\cdot,\cdot)$ return the cosine similarity and mean squared error, respectively. $\hat{\bm{G}}_t \in \mathbb{R}^{m \times n} = \bm{G}_t\bm{P}_{t-1}\bm{P}_{t-1}^\top$ and $\hat{\bm{M}}_{t-1} \in \mathbb{R}^{m \times n} = \bm{M}_{t-1}^{\rm proj}\bm{P}_{t-1}^\top$ are the full-rank estimates of gradient and first-order moment projected back from the low-rank subspace, respectively. Notably, the \textit{reconstruction term} is introduced to minimize the reconstruction error for gradients incurred by projection -- achieving the similar goal that SVD essentially aims for, and \textit{direction term} encourages the consistency of optimization direction after restoring from low-rank subspace. To solve Eqn. \ref{eq:obj} as a non-convex optimization problem, we use stochastic gradient descent (SGD) to iteratively update $\bm{P}_t$ (see theoretical analysis and implementation details in Appendix).

\textbf{Occasional Low-cost SVD to recalibrate $\bm{P}_t$ Update.} In principle, unlike GaLore, our proposed $\bm{P}_t$ update only requires to perform SGD-based iterative update without performing SVD on $\bm{G}_t$. However, considering 1) SGD may get stuck in local optimum in the optimization process; and 2) theoretically truncated SVD provides minimal approximation error for low-rank decomposition, we propose performing infrequent low-cost SVD to redirect the calculation of $\bm{P}_t$ as follows:
\begin{equation}
\label{eq:rsvd}
\begin{aligned}
&\bm{Q}_{\text{red}}, \_  = \mathrm{QR}_{\text{red}}(\bm{G}_t\bm{P}_{t-1}),\\
&\bm{U},\bm{\Sigma},\bm{Z}^\top = \mathrm{SVD}(\bm{Q}^\top_{\text{red}} \bm{G}_t), \\
&\bm{P}_{t} = \bm{Z}
\end{aligned}
\end{equation}
where $\bm{U},\bm{\Sigma} \in \mathbb{R}^{r \times r}, \bm{Z} \in \mathbb{R}^{n \times r}$ and $\mathrm{QR}_{\text{red}}(\cdot)$ is the reduced QR decomposition, which returns $\bm{Q}_{\text{red}} \in \mathbb{R}^{m \times r}$ consisting of orthogonal columns. 
Our key idea is to first project $\bm{G}_t$ into the $\bm{P}_{t-1}$-defined low-rank subspace and obtain $\bm{Q}_{\text{red}}$. Then, because all the columns of $\bm{Q}_{\text{red}}$ are orthogonal, we have $\bm{G}_t \approx \bm{Q}_{\text{red}}\bm{Q}^\top_{\text{red}} \bm{G}_t \approx \bm{Q}_{\text{red}}\bm{U}\Sigma \bm{Z}^\top$ to determine $\bm{P}_t$ as $\bm{Z}$ . Notably, here the SVD on $\bm{Q}^\top_{\text{red}} \bm{G}_t$ can be viewed as the approximated version for SVD on $\bm{G}_t$, but the much smaller size of $\bm{Q}^\top_{\text{red}}$ brings significant reduction in computational complexity from  $\mathcal{O}(mn^2)$ to $\mathcal{O}(mr^2)$. Evaluation on single A100 GPU shows that this low-cost method only consumes 23 seconds when updating all $\bm{P}_t$ for LLaVA-7B model, bringing over $20\times$ speedup than the SVD operation in GaLore.

\textbf{Overall Training Procedure.} The proposed gradient projection update schemes, due to their generality, can be seamlessly integrated into existing optimizers such as Adam and Adafactor. Algorithm \ref{alg:AGLP} outlines the overall Adam-based training procedure incorporating the proposed $\bm{P}_t$ update approach. In this setup,  $\bm{P}_t$ is updated at regular intervals every $T_{\rm u}$ steps following Eqn. \ref{eq:obj}, with additional recalibrating based on Eqn.~\ref{eq:rsvd} every $\lambda \times T_{\rm u}$ steps. Therefore, the weight update process can be formulated as:
\begin{equation}
\bm{W}_t = \bm{W}_0 + \sum_{i=0}^{\lambda-1} \sum_{j=1}^{T_{\rm u}} \Delta \bm{W}_{i\times T_{\rm u}+j}^{\rm proj} \bm{P}_i^\top,
\end{equation}
where $t=\lambda \times T_{\rm u}$ and $\Delta \bm{W}_i^{\rm proj}$ represent the weight update within the low-rank subspace defined by $\bm{P}_i$, with gradients over each $T_{\rm u}$ steps sharing the same $\bm{P}_i$. Notably, since $T_{\rm u}$ and $\lambda$ are hyper-parameters that control the update interval of $\bm{P}_t$, the corresponding ablation study on their effects is discussed in Section \ref{sec:ablation}. Additionally, we provide details on the CNN-specific handling and the Adafactor-based algorithm in Appendix.

\vspace{-1mm}
\begin{algorithm}
\begin{algorithmic}
\setlength{\itemindent}{-1em} 
\small
   \caption{Adam with COAP}
   \label{alg:AGLP}
   \STATE {\bfseries Input:} Weight matrix $\bm{W} \in \mathbb{R}^{m \times n}$, Learning rate $\eta$, Rank $r$, \\
   \hspace{2em} Betas $[\beta_1, \beta_2]$, Update interval $[\lambda, T_{\rm u}]$.\\
   \STATE {\bfseries Initialize:} $\bm{M}_0^{\rm proj} \in \mathbb{R}^{m \times r} \leftarrow 0$, $\bm{V}_0^{\rm proj}\in \mathbb{R}^{m \times r} \leftarrow 0$, 
   $t\leftarrow 0$ \\
   \STATE {\bfseries Randomly Initialize:} $\bm{P}_0 \in \mathbb{R}^{n \times r} $\\
   \STATE \textbf{Compute:} $\bm{P}_0\leftarrow (\bm{P}_0,\bm{G}_0)$$\quad\quad\quad\quad\quad\quad\quad\quad\triangleright$ Eqn.~\ref{eq:rsvd} \\
   \hspace{-1em}\For{ $t ~{\rm in}~ [1, 2, \cdots]$}{
        \STATE \hspace{1em}\textbf{Compute:} gradient $\bm{G}_t$ of $\bm{W}_t$ in the loss function. \\
        \If {$t \mod T_{\rm u} = 0$} {
            \If {$t \mod (\lambda\times T_{\rm u}) = 0$} {
                \STATE\hspace{1em} \textbf{Compute:} $\bm{P}_t\leftarrow (\bm{P}_{t-1},\bm{G}_t) $ \hfill $\triangleright$ Eqn.~\ref{eq:rsvd}
            }
            \Else {
                \STATE\hspace{1em} \textbf{Update:}$\bm{P}_{t}\leftarrow(\bm{P}_{t-1},\bm{G}_t,\bm{M}_{t-1}) $\hfill$\triangleright$ Eqn.~\ref{eq:obj}\\
            }
        }
        \Else {
            $\bm{P}_t\leftarrow \bm{P}_{t-1}$
        }
        \comments{$\triangleright$Project gradient and moments into low-rank space.} \\
        $\bm{G}_t^{\rm proj} \leftarrow \bm{G}_t\bm{P}_t$ \\
        $\bm{M}_t^{\rm proj} \leftarrow \beta_1\bm{M}_{t-1}^{\rm proj}+(1-\beta_1)\bm{G}_t^{\rm proj}$ \\
        $\bm{V}_t^{\rm proj} \leftarrow \beta_2\bm{V}_{t-1}^{\rm proj}+(1-\beta_2)(\bm{G}_t^{\rm proj})^2$ \\
        \comments{$\triangleright$Calculate the bias correction term in low-rank space.} \\
        $\Delta \bm{W}_t^{\rm proj}\leftarrow \frac{\bm{M}_t^{\rm proj}/(1-\beta_1^t)}{\sqrt{\bm{V}_t^{\rm proj}/(1-\beta_2^t)}+\epsilon }$ \\
        \comments{$\triangleright$Restore $\Delta \bm{W}_t^{\rm proj}$ to original space and update $\bm{W}$.} \\
        $\bm{W}_t \leftarrow \bm{W}_{t-1} - \eta  \Delta \bm{W}_t^{\rm proj}\bm{P}_t^\top$ \\
       }
    \STATE {\bfseries Return:} updated $\bm{W}$ \\
\end{algorithmic}
\end{algorithm}
\vspace{-3mm}
\section{Experiments}
\label{sec:exp}

We evaluate COAP over various scales of datasets and models, including both pre-training and fine-tuning. The models assessed encompass architectures that contain convolutional layers, \emph{e.g.}, Latent Diffusion Models (LDM)~\cite{rombach2021high}, ControlNet-XL~\cite{podell2023sdxl, zhang2023adding}, as well as architectures that based only on transformers, including Scalable Interpolant Transformers (SiT)~\cite{ma2024sit}, LLaVA~\cite{liu2024improved, liu2024visual} and LLaMA~\cite{touvron2023llama}.

\textbf{Rank Ratio} Given the varying sizes of weight matrices in models, \emph{e.g.}, LDM, and SDXL, we define the rank ratio as $c$ to unify the calculation of ranks for different matrices. For an $m \times n$ matrix, the rank $r$ is given by $r = \frac{\min(m, n)}{c}$.

In our experiments, we use various measures, and ``$\uparrow$" indicates that a higher value is better for this metric, while ``$\downarrow$" indicates that a lower value is better for this metric. We provide the detailed hyper-parameter settings for all experiments in Section 1.3 of Appendix.

\subsection{Pre-training LDM}
\textbf{Experimental Settings.} We follow the configuration of LDM\footnote{\url{https://github.com/CompVis/latent-diffusion}} and train the U-Net model of LDM from scratch using different optimizers on 8$\times$V100 GPUs with a batch size of 32. We generate 50 images for each of the 1000 classes with 250 DDIM steps and compute the Fréchet Inception Distance (FID)~\cite{heusel2017gans} between the generated images and the validation dataset of ImageNet-1K~\cite{imagenet_cvpr09}. 

\textbf{Comparison Results.} As shown in Table~\ref{tbl:ldm}, our method reduces optimizer memory by 40\%, and decreases the FID by 1.8 and 20.4 compared to the baseline. When integrated into AdamW, our method reduces the FID by 1.6 compared to GaLore. In Adafactor, our method requires only 70\% of the memory that GaLore needs, reduces the extra training time by 60\%, and decreases the FID score by 5.0.

\begin{table}[t]
\centering
\small
\caption{Pre-training LDM on the ImageNet-1K dataset for 690K steps using 8$\times$V100 GPUs. FID is reported, along with training time and GPU memory usage of optimizer states in FP32 format.}
\vspace{-3mm}
\label{tbl:ldm}
\setlength{\tabcolsep}{3.5pt}
\resizebox{1\linewidth}{!}{\begin{tabular}{lcccc}
\toprule
Method & Rank Ratio & Optimizer Mem. (GB)$\downarrow$  & Training Time$\downarrow$ & FID$\downarrow$   \\ 
\midrule
AdamW    &- & 3.0     &310.3 h        & 18.0    \\
GaLore   &2 & 2.0 (-33\%)     & +21\%      & 17.8    \\
\rowcolor{gray!15}\textbf{COAP}   & 2 & \textbf{1.8 (-40\%) } & \textbf{+13\%}    & \textbf{16.2 }  \\
\midrule
Adafactor &-& 2.2     &336.9 h       & 38.7    \\
GaLore    &2& 1.8 (-18\%)      &+18\%       & 23.3    \\
\rowcolor{gray!15}\textbf{COAP}     &2.2 & \textbf{1.3 (-41\%) }     & \textbf{+7\%}    & \textbf{18.3 }  \\
\bottomrule
\vspace{-7mm}
\end{tabular}}
\end{table}

\subsection{Pre-training SiT-XL/2} 
\textbf{Experimental Settings.} We perform Sit-XL/2~\cite{ma2024sit} training with Representation Alignment (REPA)~\cite{yu2024representation} on the ImageNet-1K~\cite{imagenet_cvpr09} dataset, preprocessing each image to a resolution of 256×256 pixels. Both the training and evaluation settings follow those outlined in REPA. All models are trained for 400K iterations with CFG (classifier-free guidance). For FID computation, we generate 50K samples using the Euler-Maruyama sampler, with the number of inference steps set to 250.

\textbf{Comparison Results.} As shown in Table~\ref{tbl:sit}, our method outperforms other low-rank methods, achieving comparable performance to full-rank training while reducing optimizer memory by more than 40\%. In contrast, using LoRA for pre-training reduces optimizer memory by 29\%, but the adapter increases model size by 48\%, leading to a 33\% increase in training time and a significant FID increase of 149.6. Flora, which uses random projection, also incurs additional training time to generate low-rank mapping matrices per iteration, increasing FID by 113.3. Compared to GaLore, our method in AdamW achieves a lower FID with 58\% less extra training time. In Adafactor, our FID is 0.9 lower than GaLore's, with 78\% less extra training time.

\begin{table}[t]
\centering
\caption{Pre-training SiT-XL/2 with REPA on the ImageNet-1K dataset for 400K steps using 8$\times$H100 GPUs. We set the rank as 512 for all methods. FID scores are reported, along with training time and GPU memory usage of optimizer states in FP32 format.}
\vspace{-3mm}
\label{tbl:sit}
\footnotesize
\resizebox{1\linewidth}{!}{\begin{tabular}{lcccc}
\toprule
Method    & \begin{tabular}[c]{@{}c@{}}Optimizer\\  Mem. (GB)$\downarrow$\end{tabular} & \begin{tabular}[c]{@{}c@{}} Model\\ Mem. (GB)$\downarrow$\end{tabular} & \begin{tabular}[c]{@{}c@{}}Training \\Time$\downarrow$\end{tabular} & \begin{tabular}[c]{@{}c@{}}FID$\downarrow$ \end{tabular}\\
\midrule
AdamW     & 5.1           & 2.5  &20.8 h       &1.9    \\
\midrule
GaLore    & 2.6 (-49\%)          & 2.5   &+38\%     &2.3  \\
LoRA    &  3.6 (-29\%)        &  3.7 (+48\%)  &+33\%       &151.9   \\
ReLoRA      & 3.6 (-29\%)      & 3.7 (+48\%)  &+33\%  & 151.8  \\
\rowcolor{gray!15}\textbf{COAP}      & \textbf{2.6 (-49\%) }   & 2.5  &\textbf{+14\% }     &\textbf{2.2}   \\
\midrule
\midrule
Adafactor & 2.5          & 2.5  & 25.5 h & 1.9  \\
\midrule
GaLore    & 1.5 (-40\%)        & 2.5   &+33\%   &3.0  \\
Flora    & 1.6 (-36\%)        & 2.5   & +7\%   &115.2   \\
\rowcolor{gray!15}\textbf{COAP}     &\textbf{1.5 (-40\%)} & 2.5   &\textbf{+7\%}   &\textbf{2.1}   \\
\bottomrule
\end{tabular}}
\end{table}

\subsection{Training ControlNet with Stable Diffusion XL}
\begin{table}[t]
\centering
\caption{Training ControlNet based on SDXL for 80K steps using 8$\times$H100 GPUs in BF16 format, conditioned on human poses. }
\vspace{-3mm}
\label{tbl:sdxl}
\resizebox{1\linewidth}{!}{\begin{tabular}{lcccccccc}
\toprule
\multirow{2}{*}{Method} & Rank  & Optimizer    & \multicolumn{3}{c}{mAP$\uparrow$ @ training steps}   &\multirow{2}{*}{Converged}   & \multicolumn{1}{c}{Training}\\ 
\cline{4-6}
                        & Ratio & Mem. (GB)$\downarrow$                             & 20K           & 40K           & 80K       &    & Time (80K)$\downarrow$ \\ 
\midrule
AdamW &-&9.3          &     18.3 &      19.1&    19.9 & \xmark &  19.3 h\\
Adafactor&- &5.1                & 19.2     & 70.0   &72.7 & \cmark & 22.3 h \\
\midrule
Flora &2 & 3.9 (-24\%)         &    18.0&      18.9&   19.6  & \xmark &  +16\%\\
GaLore   &2 & 4.7 (-8\%)             &18.6      &67.0   &72.7  & \cmark & +39\% \\
GaLore-8bit   &2 & 3.1 (-39\%)          &19.6      &20.8   &20.9 & \xmark   & +49\%   \\
\rowcolor{gray!15}\textbf{COAP}     &2 &3.6 (-29\%)              &\textbf{66.6}      &\textbf{71.6}   &\textbf{73.4} & \cmark  & \textbf{+4\%}\\
\rowcolor{gray!15}\textbf{8-bit COAP}     &2 &\textbf{1.9 (-63\%)}             &18.7      &66.9   &72.2 & \cmark & +17\%\\
\midrule
GaLore   &4   & 3.5 (-31\%)             &18.9      &19.7   &19.5 & \xmark & +39\%\\
8-bit GaLore   &4 & 3.1 (-39\%)             &18.8      &19.7   &19.8 & \xmark & +50\% \\
\rowcolor{gray!15}\textbf{COAP}    &4 & 1.8 (-65\%)           &\textbf{50.9}      &\textbf{70.4}  &\textbf{72.1} & \cmark  &  \textbf{+5\%} \\
\rowcolor{gray!15}\textbf{8-bit COAP}     &4 &\textbf{1.0 (-80\%)}            &19.4      &19.2   &71.5& \cmark& +15\%   \\
\midrule
GaLore   &8   & 2.7 (-47\%)            &18.6      &18.2   &19.7 & \xmark  & +35\% \\
8-bit GaLore   &8 & 2.3 (-55\%)              &18.6     &18.2   &19.7  & \xmark& +45\% \\
\rowcolor{gray!15}\textbf{COAP}    &8  & 0.9 (-82\%)             &\textbf{25.8 }     &\textbf{70.2}  &\textbf{72.6} & \cmark &  \textbf{+6\%}\\
\rowcolor{gray!15}\textbf{8-bit COAP}     &8 &\textbf{0.5 (-90\%)}             &19.3      &18.9   &69.9 & \cmark& +13\% \\
\bottomrule

\end{tabular}}
\end{table}

\begin{table}[ht]
    \centering
    \small
    \setlength{\tabcolsep}{2pt}
     \caption{Comparison of generated images at different training steps (20K, 40K, 80K). We use the DDIM with a guidance scale of 5.0 for generating images. The number of inference steps is 50.}
     \vspace{-3mm}
    \begin{tabular}{ccc|ccc}
        \toprule
        \multicolumn{6}{c}{Prompt: a man in a shirt and tie holding a cup of coffee} \\
        \midrule
        \multirow{2}{*}{\begin{tabular}[c]{@{}c@{}}Human\\ pose\end{tabular}} & \multirow{2}{*}{\begin{tabular}[c]{@{}c@{}}Reference\end{tabular}} && 20K & 40K & 80K\\
        \cmidrule{4-6}
        &&&
        \multicolumn{3}{c}{\begin{tabular}[c]{@{}c@{}}GaLore (2.7 GB)\end{tabular}} \\
         \includegraphics[width=0.07\textwidth]{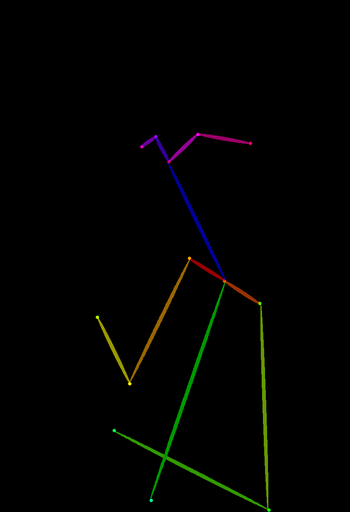} 
        & \includegraphics[width=0.07\textwidth]{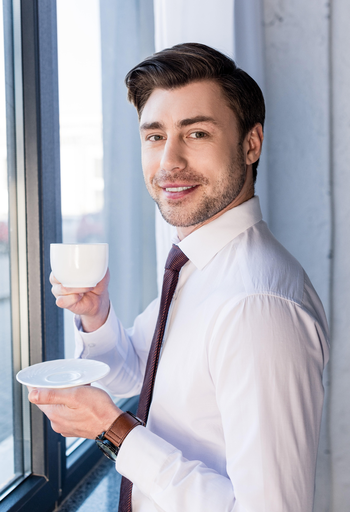} &
        &\includegraphics[width=0.07\textwidth]{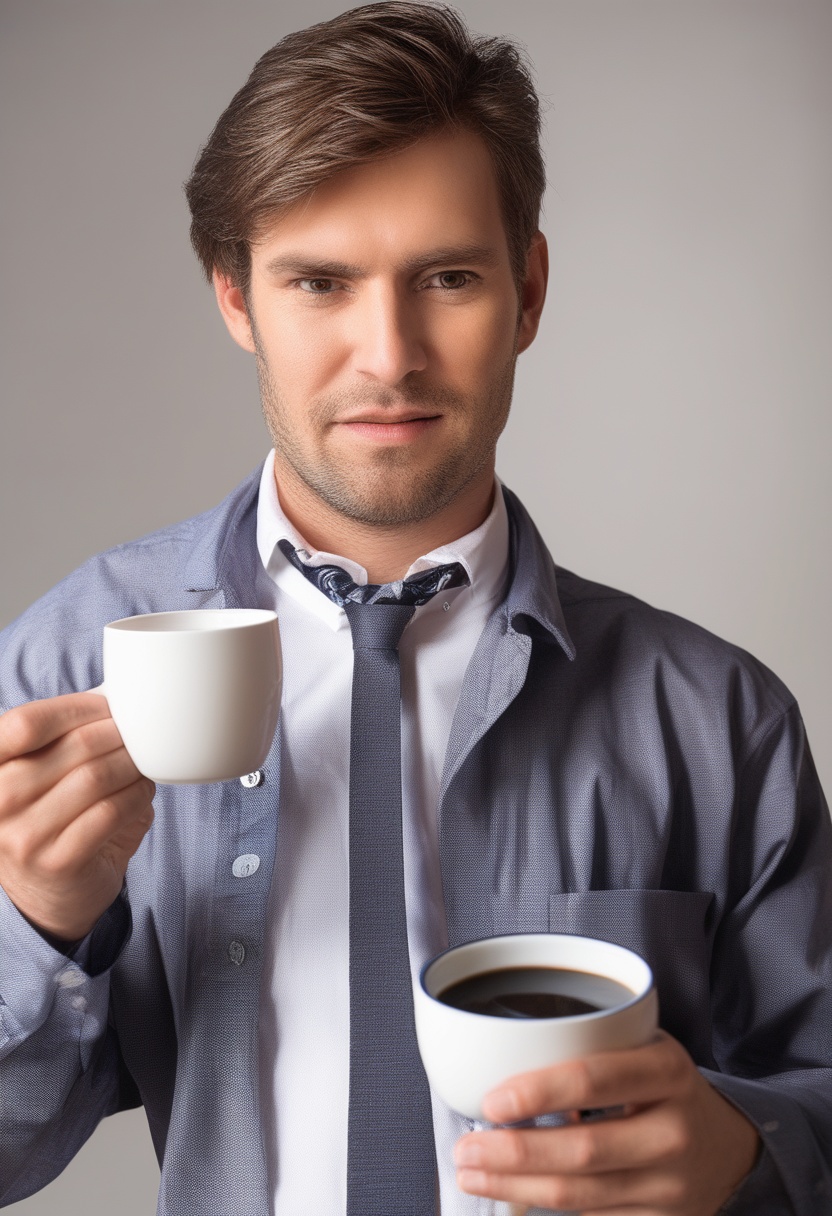} 
        & \includegraphics[width=0.07\textwidth]{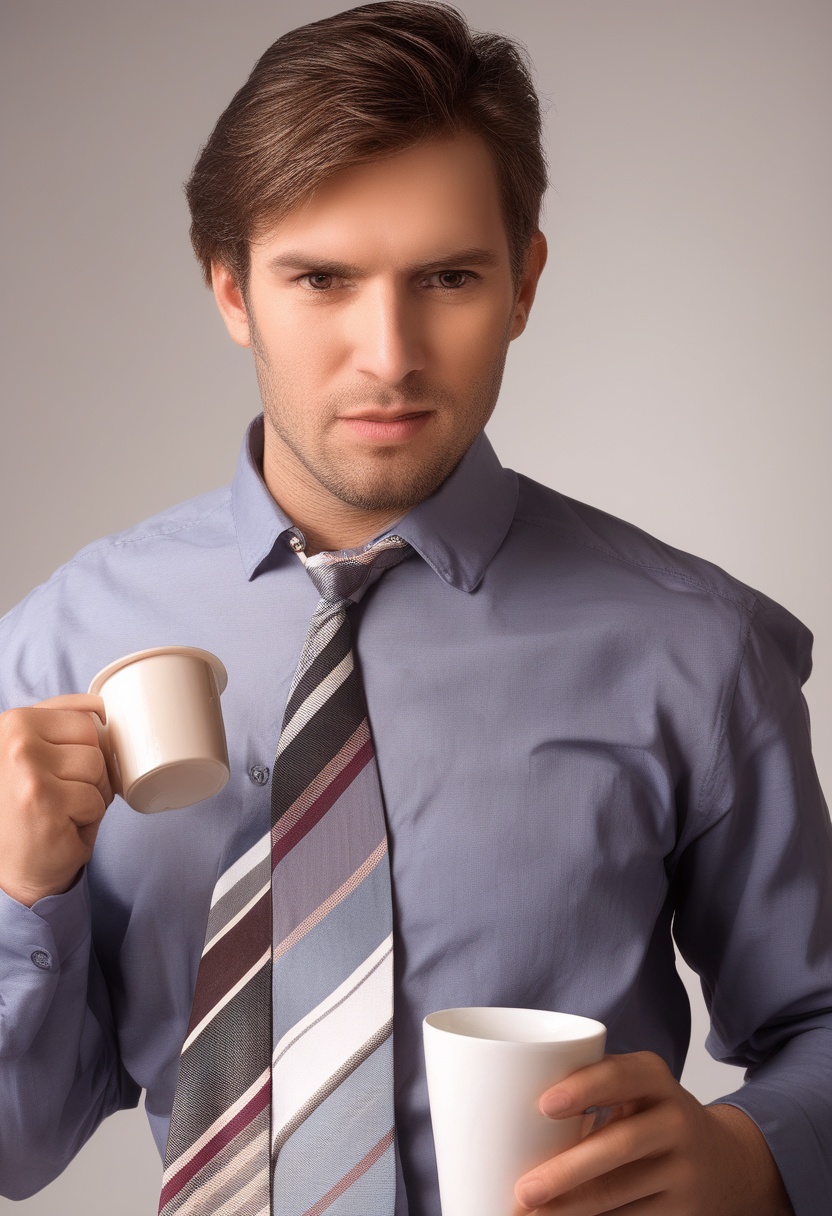} 
        & \includegraphics[width=0.07\textwidth]{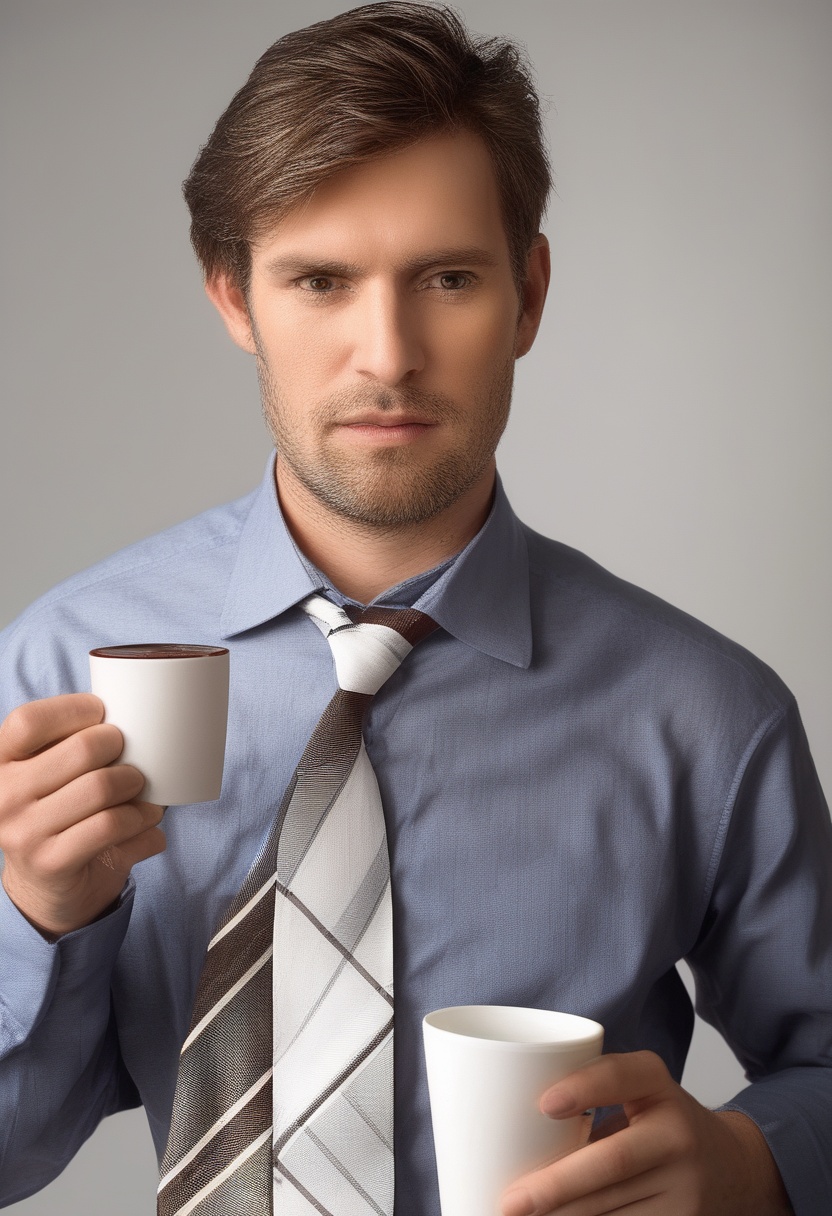}\\
        \midrule
        \multicolumn{3}{c|}{\begin{tabular}[c]{@{}c@{}}\textbf{COAP} (0.9 GB)\end{tabular}} &\multicolumn{3}{c}{\begin{tabular}[c]{@{}c@{}}\textbf{8-bit COAP (0.5 GB)}\end{tabular}} \\
        \includegraphics[width=0.07\textwidth]{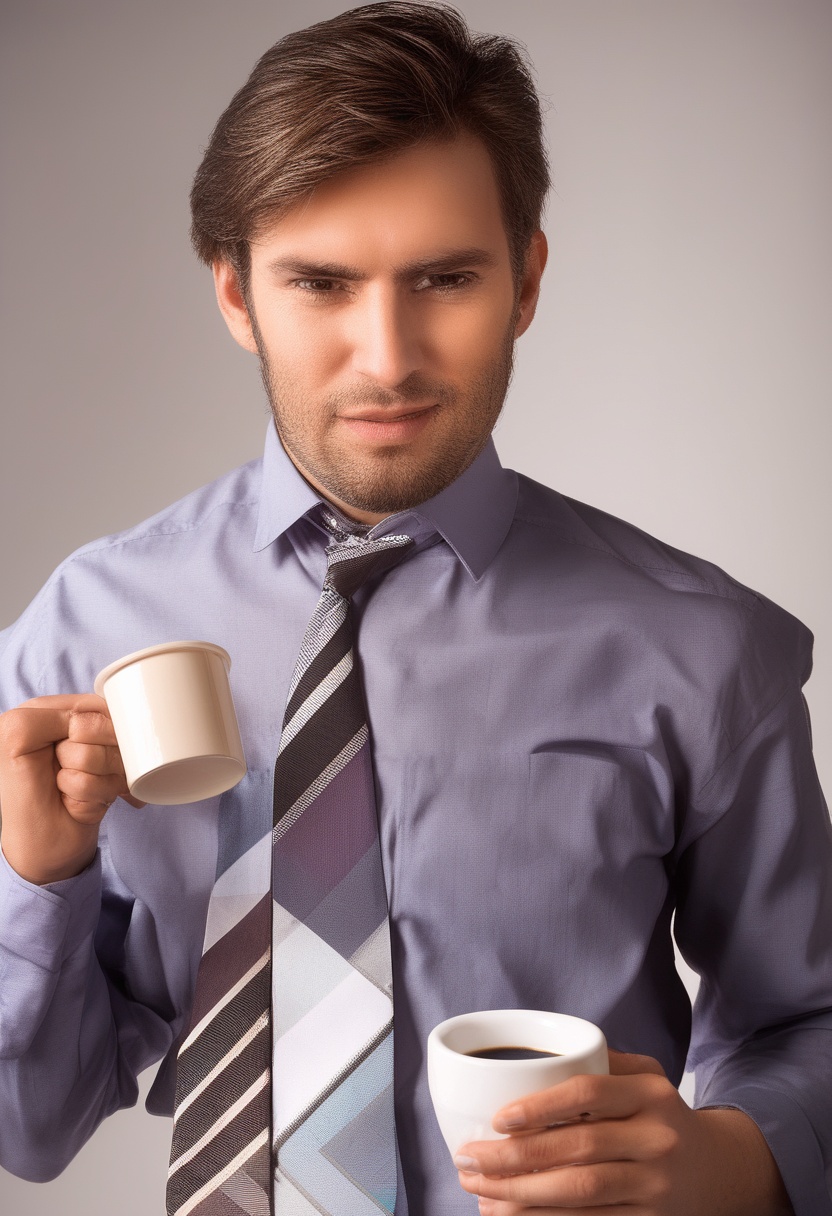} 
        & \includegraphics[width=0.07\textwidth]{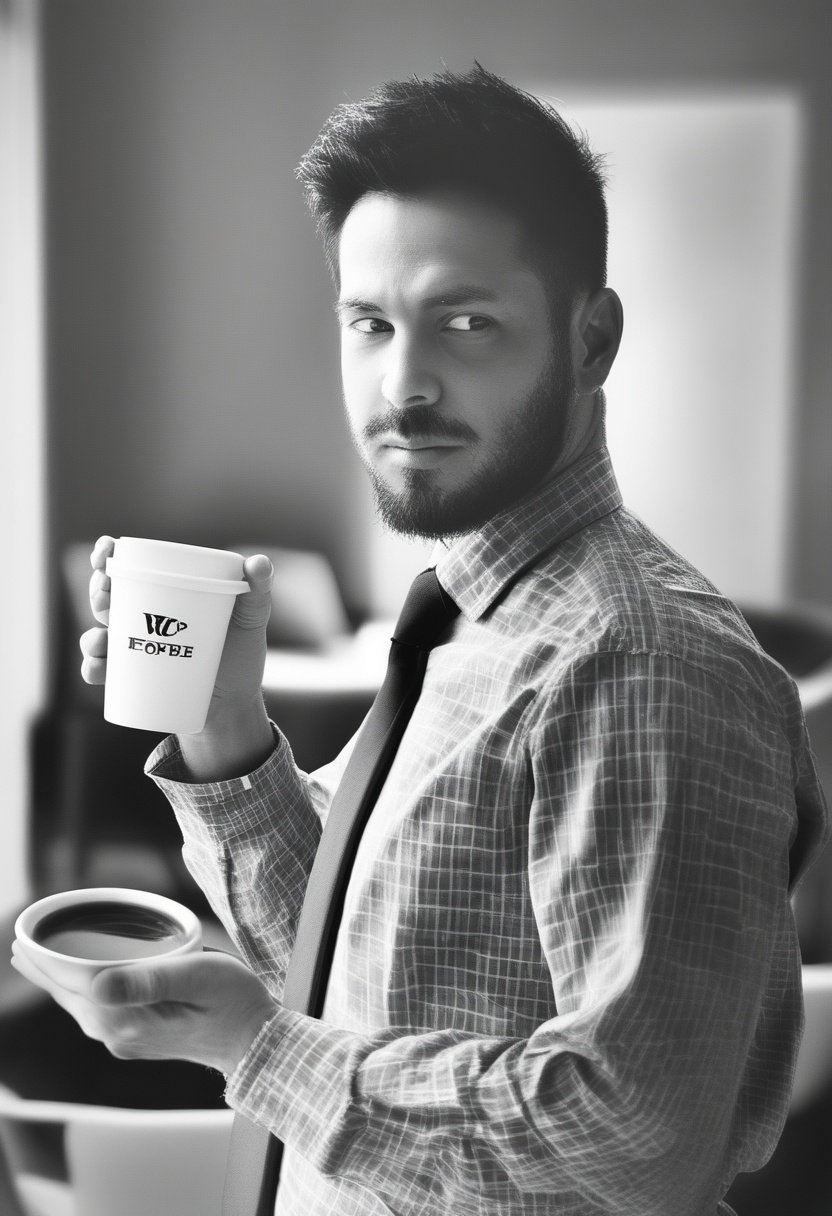} 
        & \includegraphics[width=0.07\textwidth]{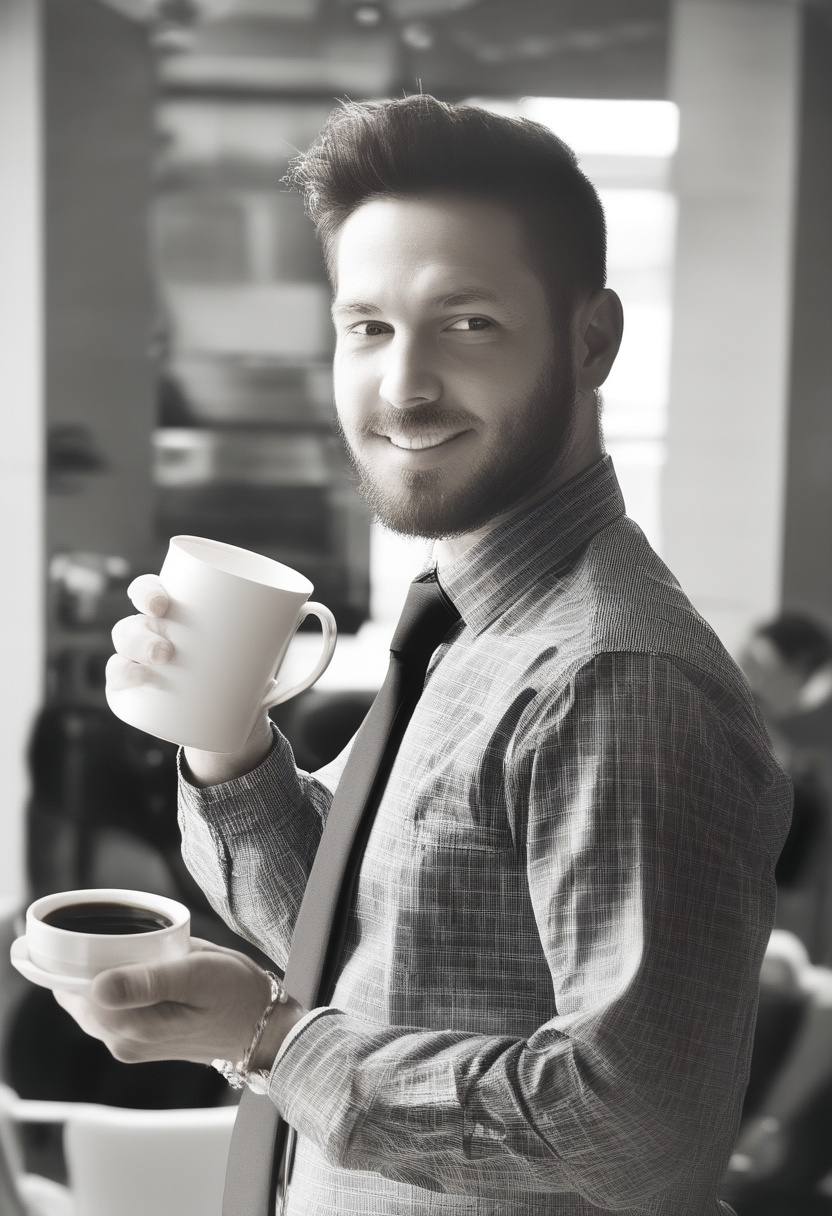}
        &\includegraphics[width=0.07\textwidth]{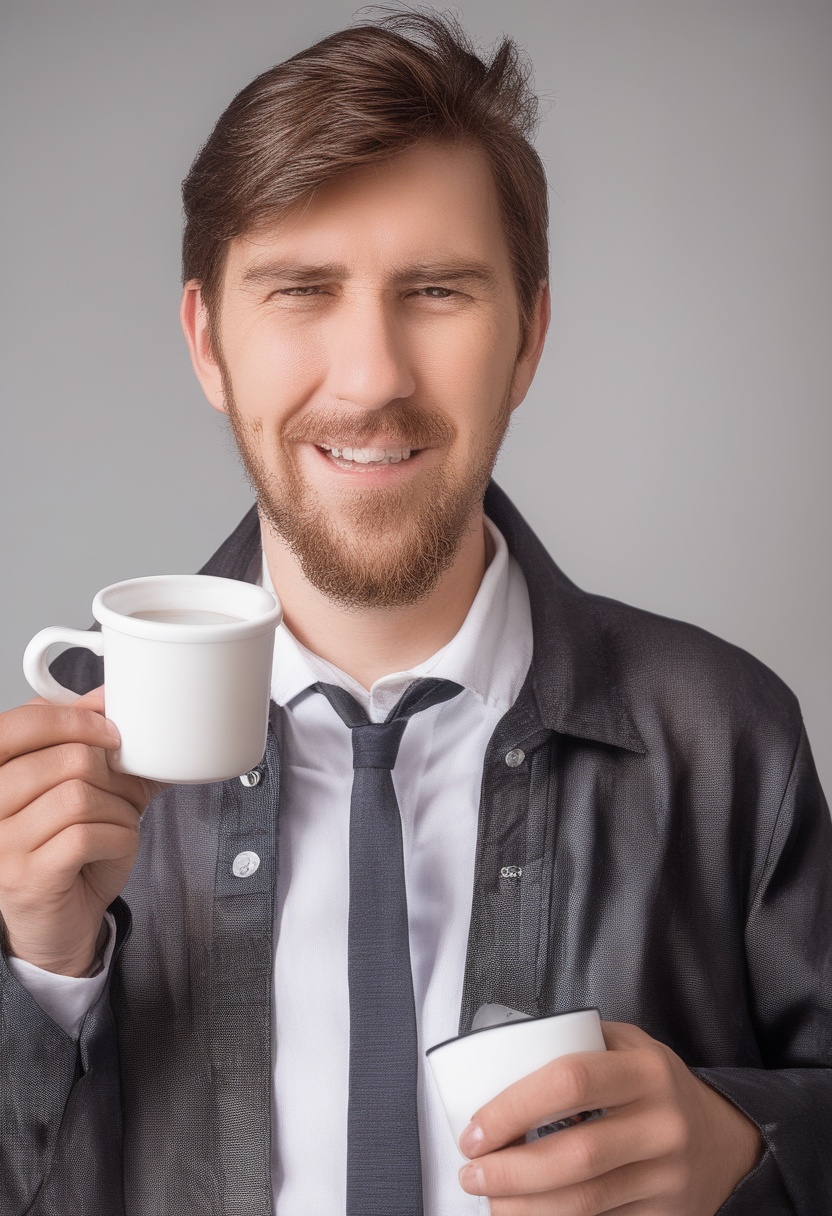} 
        & \includegraphics[width=0.07\textwidth]{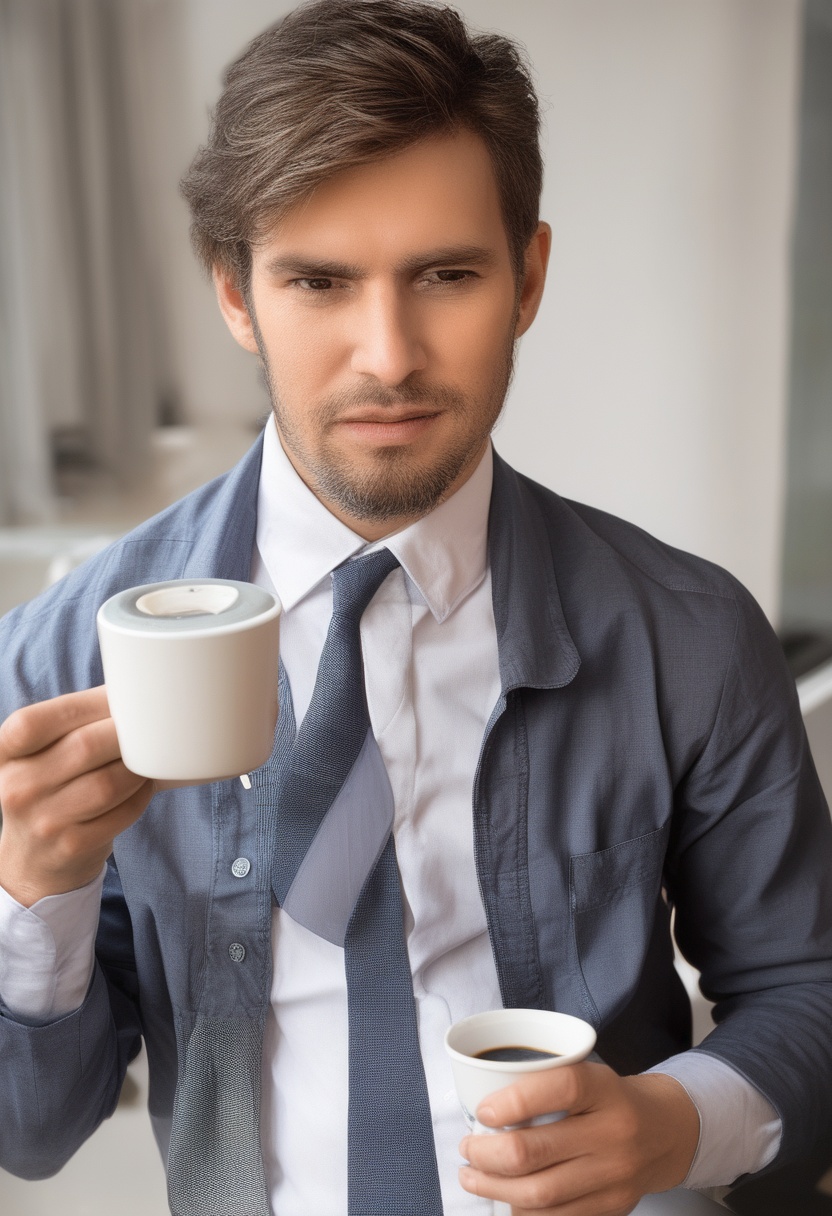} 
        & \includegraphics[width=0.07\textwidth]{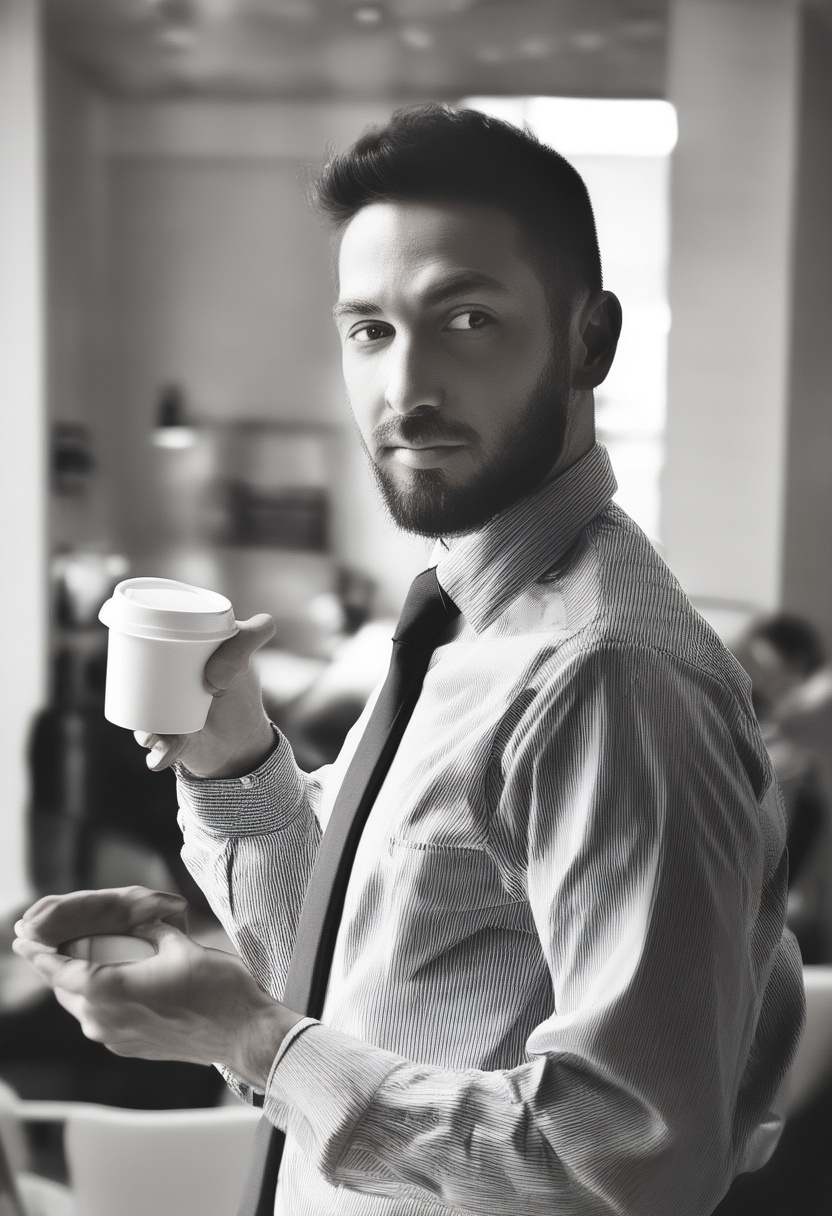}
 \\
        \bottomrule
    \end{tabular}
    \vspace{-5mm}
    \label{tbl:sdxl-imgs}
\end{table}

\textbf{Experimental Settings.} We train a ControlNet~\cite{zhang2023adding} based on Stable Diffusion XL~\cite{podell2023sdxl} for text-to-image generation using human poses as additional control images. Single-person images are collected as training data, with RTMPose~\cite{jiang2023rtmpose} used for pose estimation and BLIP2~\cite{li2023blip} for generating text descriptions and prompts. The estimated poses serve as ground truth, and 1000 samples are randomly selected for evaluation. Specifically, we use human pose and BLIP2-generated prompts as inputs, perform pose estimation on the generated images using RTMPose, and calculate the mean Average Precision (mAP) to evaluate the match between the generated poses and the original poses. The mAP is computed by setting Intersection over Union (IoU) thresholds from 0.5 to 0.95 in increments of 0.05, calculating the IoU between predicted and ground truth key points to determine the Average Precision (AP), and averaging these AP values. We train the model for 80K steps using 8$\times$H100 GPUs in BF16 format, with total batch size of 32, learning rate of $1\times10^{-5}$, and weight decay of 0.

\textbf{Comparison Results.} Table~\ref{tbl:sdxl} presents the quantitative evaluation results of models trained with different optimizers. The convergence speed of Adafactor is significantly faster than AdamW, so we chose Adafactor for our experiments. Our method achieves results comparable to Adafactor while reducing optimizer memory usage by 80\%, with only a 6\% increase in additional training time. In contrast, methods like Flora and GaLore fail to converge at the same compression rate. Additionally, our method is also applicable to 8-bit, reducing optimizer memory usage by 90\%. Table~\ref{tbl:sdxl-imgs} compares images generated by models trained with different methods. Our method is able to generate human poses that align with the input at just 20K steps, while GaLore fails to converge even after 80K steps, leading to noticeable mismatches between the generated and input poses. 

\subsection{Pre-training LLaMA-1B and LLaMA-7B}
\textbf{Experimental Settings.} For large language models, we follow the training settings described in~\cite{zhao2024galore} to train LLaMA-1B from scratch on the C4 dataset. Our training employs an initial learning rate of 0.01 and a total batch size of 512.

\textbf{Comparison Results.} As shown in Table~\ref{tbl:llama-1b}, our method outperforms other methods. In LLaMA-1B, compared to other low-rank methods, we reduce the optimizer memory by 61\% with only a 2\% increase in additional training time, which is significantly lower than the training time increase of other methods, and achieve performance comparable to AdamW. In contrast, LoRA and ReLoRA require an additional 36\% model size and result in perplexity (PPL) increases of 3.65 and 2.77, respectively. In LLaMA-7B using 8-bit optimizers, our method outperforms 8-bit Adam~\cite{dettmers20218}, reducing training time by 2\% and lowering PPL by 0.11. Meanwhile, GaLore increases training time by 19\% and suffers a PPL increase of 0.12.

\begin{table}[t]
\centering
\caption{Pre-training LLaMA-1B and LLaMA-7B on the C4 dataset using 8$\times$H100 GPUs. PPL is reported along with GPU memory usage of optimizer states and the model in BF16 format. The rank is 512 for LLaMA-1B and 1024 for LLaMA-7B.  }
\vspace{-3mm}
\label{tbl:llama-1b}
\resizebox{1\linewidth}{!}{\begin{tabular}{clcccc}
\toprule
Model &Method    & \begin{tabular}[c]{@{}c@{}}Optimizer\\  Mem. (GB)$\downarrow$\end{tabular} & \begin{tabular}[c]{@{}c@{}} Model\\  Mem. (GB)$\downarrow$\end{tabular} & \begin{tabular}[c]{@{}c@{}}Training \\Time$\downarrow$\end{tabular} & \begin{tabular}[c]{@{}c@{}}PPL$\downarrow$ \end{tabular}\\
\midrule
\multirow{5}{*}{\begin{tabular}[c]{@{}c@{}}LLaMA\\1B\\ (100K)\end{tabular}}&AdamW &4.99  &2.49 &28.50 h &15.56\\
\cmidrule{2-6}
&GaLore   &1.94 (-61\%)  &2.49&+17\%&15.64\\
&LoRA &2.27 (-55\%)  &3.38 (+36\%) &+6\%&19.21\\
&ReLoRA &2.27 (-55\%)  &3.38 (+36\%) &+6\%&18.33\\
& \textbf{COAP}     &\textbf{1.94 (-61\%)} &2.49 &\textbf{+2\%} &\textbf{15.56}\\
\midrule
\multirow{3}{*}{\begin{tabular}[c]{@{}c@{}}LLaMA\\7B\\ (80K)\end{tabular}}&8-bit Adam &12.55  &12.55 &52.01 h &15.39\\ 
\cmidrule{2-6}
&8-bit GaLore   &5.25 (-58\%)  &12.55&+19\%&15.47\\ 
& \textbf{8-bit COAP}    &\textbf{5.25 (-58\%)} &12.55 &\textbf{-2\%} &\textbf{15.28}\\ 
\bottomrule
\end{tabular}}
\end{table}

\subsection{Fine-tuning LLaVA-7B}

\begin{table}[t]
\small
\setlength{\tabcolsep}{2.5pt}
\centering
\caption{Fine-tuning LLaVA-v1.5-7B on the ScienceQA dataset using 1$\times$A100. ``OOM" means out-of-memory.}
\vspace{-3mm}
\label{tbl:llava}
\resizebox{1\linewidth}{!}{\begin{tabular}{lcccc}
\toprule
Method          & \begin{tabular}[c]{@{}c@{}}Training \\Time$\downarrow$ \end{tabular} & \begin{tabular}[c]{@{}c@{}}Optimizer \\ Mem. (GB)$\downarrow$\end{tabular} &\begin{tabular}[c]{@{}c@{}} Model\\  Mem. (GB)$\downarrow$\end{tabular} & \begin{tabular}[c]{@{}c@{}}ScienceQA\\ IMG-Acc(\%)$\uparrow$\end{tabular} \\
\midrule
AdamW           & -       & OOM &OOM  & -     \\
DeepSpeed & 47.1 h     & 26.4  & 13.2       & 82.4  \\
GaLore       & 30.2 h     & 8.1 (-49\%)   & 13.2        & 91.1      \\
LoRA        &  11.1 h        &8.1 (-49\%)  & 17.2 (+30\%)     & \textbf{92.3} \\
Flora   &9.5 h &8.1 (-49\%) &13.2 & 66.6 \\
\rowcolor{gray!15} \textbf{COAP}          & \textbf{7.6 h}       & 8.1 (-49\%)    & 13.2       & \textbf{92.3} \\
\midrule
8-bit GaLore       &30.0 h      & 4.9 (-81\%)   &13.2         & 90.7      \\
\rowcolor{gray!15} \textbf{8-bit COAP}          & \textbf{7.8 h}       &4.9 (-81\%)     &13.2        & 92.0 \\
\bottomrule
\vspace{-5mm}
\end{tabular}}
\end{table}

\textbf{Experimental Settings.} We apply our method to a recent state-of-the-art Large Multimodal Model (LMM): LLaVA-v1.5-7B~\cite{liu2024improved}. LLaVA connects the pre-trained CLIP ViT-L/14~\cite{radford2021learning} visual encoder and the large language model Vicuna~\cite{zheng2023judging} using a simple projection matrix for feature alignment. We perform fine-tuning on LLaVA based on this setup. We train the model using task-specific fine-tuning on the ScienceQA~\cite{lu2022learn} dataset for 12 epochs with a batch size of 16 and a learning rate of $2\times10^{-5}$ on 1$\times$A100 GPU. 

\textbf{Comparison Results.} Table~\ref{tbl:llava} compares the results of fine-tuning LLaVA-7B on the ScienceQA dataset using different optimizers. In a single GPU environment, training large models directly with the AdamW optimizer can lead to Out-of-memory (OOM) errors due to the significant GPU memory consumption by optimizer states. Although DeepSpeed's CPU-offload~\cite{ren2021zero} feature can alleviate GPU memory pressure by migrating optimizer states to CPU memory, it significantly increases training latency due to frequent data transfers. Compared to DeepSpeed and GaLore, our training speed is improved by 6$\times$ and 4$\times$, respectively, with accuracy improvements of 9.9\% and 1.2\%. Compared to LoRA, our training speed is 1.4$\times$ faster while achieving the same performance without increasing the model size.

\section{Ablation Study}
\label{sec:ablation}
\subsection{Hyper-parameters of COAP}
In algorithm~\ref{alg:AGLP}, we introduce three key hyper-parameters: $r$, $\lambda$, and $T_u$. Rank $r$ represents the compression level of the optimizer states, with a smaller rank indicating a higher compression rate. $T_u$ denotes the update interval of the SGD-based iterative update (Eqn. \ref{eq:obj}), while $\lambda \times T_u$ represents the update interval of Occasional Low-cost SVD (Eqn. \ref{eq:rsvd}).

\begin{figure}[t]
    \centering
    \includegraphics[width=\linewidth]{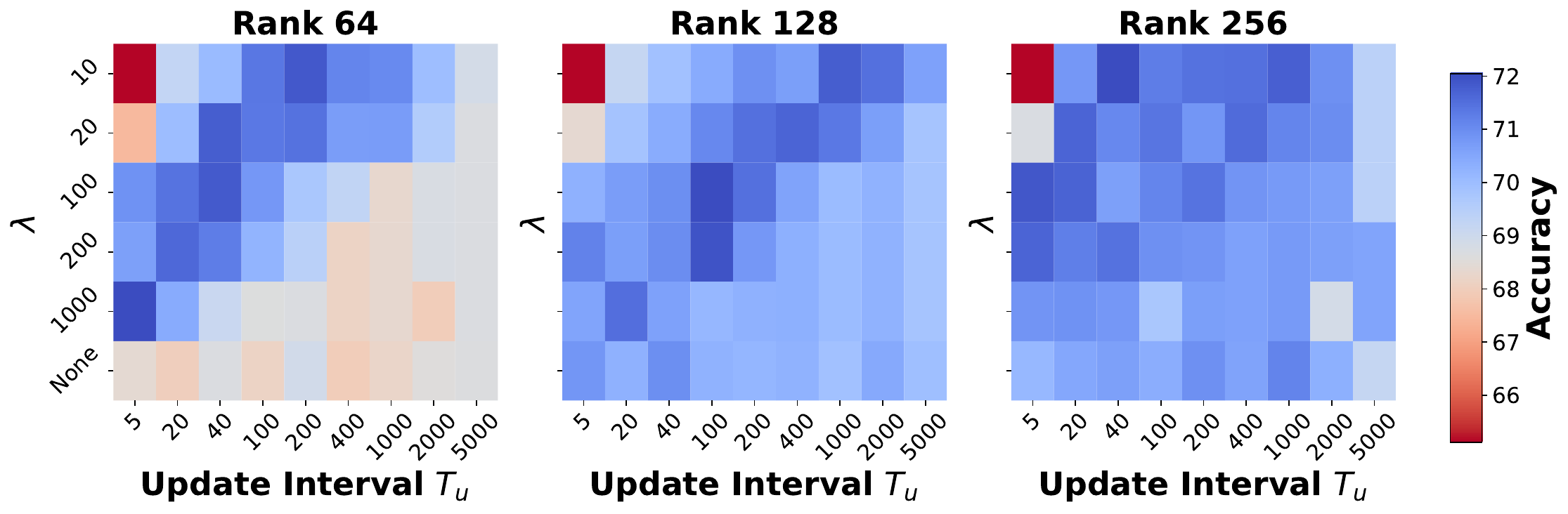}
    \vspace{-6mm}
    \caption{Ablation study on hyper-parameters $\lambda$, $r$, and $T_u$ for DeiT-Base on CIFAR-100 over 300 epochs. Here, $\lambda=\rm{None}$ means that occasional low-cost SVD does not participate in updating the low-rank projection matrix. }
    \label{fig:T_Tu}
    \vspace{-1mm}
\end{figure}

As shown in Fig.~\ref{fig:T_Tu}, it is evident that all listed ranks can achieve baseline performance (68.75\%). For the DeiT-Base model with a full-rank matrix of 768, when the rank is 256 or 128, the performance difference is minimal when $\lambda \geq 100$. However, when $\lambda$ is too small, \emph{e.g.}, 10, excessively small $T_u$ values can lead to a drop in model performance. When the compression rate exceeds 10$\times$, \emph{i.e.}, $r=64$, the model becomes highly sensitive to hyper-parameters. Additionally, it can be observed that selecting $\lambda$ and $T_u$ near the diagonal yields better performance.

\subsection{Low-rank Projection Matrix Updates}
\begin{table}[t]
\caption{Ablation study of low-rank projection matrix updates in DeiT-Base model on 1$\times$A100 GPU.}
\vspace{-3mm}
\label{tbl:upd_porj_method}
\setlength{\tabcolsep}{3.5pt}
\resizebox{1\linewidth}{!}{\begin{tabular}{lccccc}
\toprule
Method                &\begin{tabular}[c]{@{}c@{}}w/ Eqn.\ref{eq:rsvd}  \end{tabular} & \begin{tabular}[c]{@{}c@{}}w/ Eqn.\ref{eq:obj} \\ CosSim.\end{tabular} & \begin{tabular}[c]{@{}c@{}}w/ Eqn.\ref{eq:obj} \\ MSE\end{tabular}   & \begin{tabular}[c]{@{}c@{}}Top-1 Acc\\ Fine-tuning\end{tabular} & \begin{tabular}[c]{@{}c@{}}Top-1 Acc\\ Pre-training\end{tabular} \\
\midrule
AdamW (655.2 MB)             & -       & -       & -   & 90.41       & 68.75       \\
GaLore (169.2 MB)              & -       & -       & -  & 91.09       & 65.58       \\
\midrule
\multirow{8}{*}{\begin{tabular}[c]{@{}@{}l@{}}COAP\\ Rank=192 \\ (169.2 MB) \end{tabular}} 
    & \cmark       & \cmark  & \cmark  & \textbf{91.42}  & \textbf{70.39}\\
    \cmidrule{2-6}
    & \xmark       & \cmark       & \cmark      &91.25      & 63.28      \\
    & \xmark       & \cmark       & \xmark       &91.29       &63.31  \\
       & \xmark       & \xmark       & \cmark  &91.30    &63.27    \\
       \cmidrule{2-6}
     & \cmark       & \xmark       & \xmark     & 90.04      &69.83    \\
    & \cmark       & \cmark       & \xmark     & 91.29   &70.23    \\
    & \cmark       & \xmark       & \cmark    &91.26   & 69.90     \\
\bottomrule
\vspace{-10mm}
\end{tabular}}
\end{table}
Our proposed method for updating the low-rank projection matrix involves two key components: Inter-projection Correlation-aware $\bm{P}_t$ Update (Eqn.~\ref{eq:obj}) and Occasional Low-cost SVD (Eqn.~\ref{eq:rsvd}). To evaluate the benefits of these two update strategies on model performance, we conduct experiments using the DeiT-Base model. We train the model on the CIFAR-100 dataset using two strategies: training from scratch and fine-tuning a pre-trained model from ImageNet-1K, both for 300 epochs.

As shown in Table~\ref{tbl:upd_porj_method}, for the task of training from scratch, the Occasional Low-cost SVD provides the most significant benefits. In contrast, the gains are minimal for the fine-tuning task. Regarding the reconstruction term and direction term in Eqn.~\ref{eq:obj}, the direction term offers greater benefits. This indicates that during training from scratch, the model has not yet found an optimal update direction, thus requiring frequent adjustments through Occasional Low-cost SVD. On the other hand, for the fine-tuning task, the model is likely already in a good update direction, making the Inter-projection Correlation-aware update more beneficial. Combining both update strategies yields the best results for both training from scratch and fine-tuning.

\subsection{GPU Memory Profiling}
\begin{figure}[t]
    \centering
    \includegraphics[width=\linewidth]{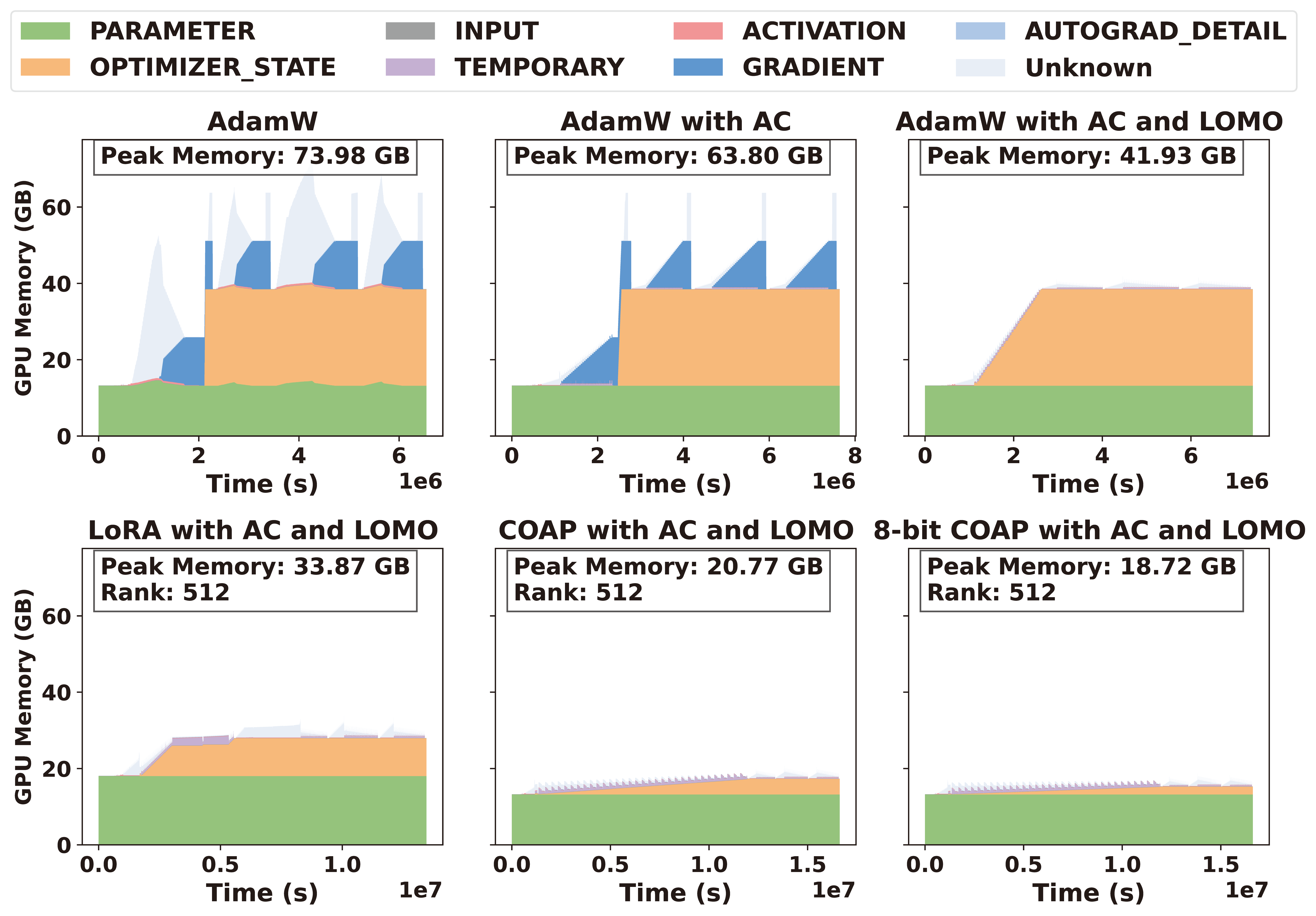}
     \vspace{-7mm}
    \caption{Profiling the GPU memory usage during the training stage of LLaVA-v1.5-7B on 1$\times$A100.}
    \vspace{-2mm}
    \label{fig:mem}
\end{figure}
To validate the effectiveness of our method on GPU memory usage, we utilize PyTorch's Memory Profiler~\cite{pytorchGPUblog} to analyze the memory consumption of LLaVA-v1.5-7B during training. We use the AdamW optimizer as the baseline with batch size of 4 and categorize the memory usage according to PyTorch’s convention. Since our method aims to reduce the optimizer states, we combined it with other complementary memory-efficient training techniques, \emph{e.g.}, activation checkpointing (AC)~\cite{chen2016training}, LOMO~\cite{lv2023full}, and quantization to reduce the overall memory usage. As shown in the memory profile of AdamW, the optimizer states occupy a significant portion (36\%) of the memory during training. with the remaining memory being used by gradients, activations, and model parameters. By enabling LOMO and AC, the memory used by gradients and activations can be reduced, but these techniques alone only reduce the peak memory usage to 41.93 GB. Finally, by incorporating our 8-bit COAP, peak memory usage is further reduced to 18.72 GB, achieving a total reduction of 75\%. As shown in Table~\ref{tbl:llava}, this significant memory reduction has minimal impact on performance.
\vspace{-0.4mm}
\section{Conclusion}

We introduce COAP, a memory-efficient approach that integrates seamlessly with momentum-based optimizers (\emph{e.g.}, AdamW, Adafactor) for large-scale training across language, vision, and multimodal domains. We analyze the limitations of current low-rank gradient projection methods, and develop a correlation-aware projection update rule that reduces computational costs and minimizes the performance gap with standard optimizers, while significantly cutting memory usage. Extensive experiments show that COAP outperforms competing methods in both training efficiency and model performance. We believe that, when combined with other memory-efficient techniques, COAP has strong potential to advance large-scale model training.

\section{Acknowledgments}
Special thanks are extended to Prof. Yuqian Zhang for providing the theoretical analysis that strengthened this research.

{
    \small
    \bibliographystyle{ieeenat_fullname}
    \bibliography{main}
}

\clearpage
\setcounter{section}{0}
\setcounter{equation}{0}
\setcounter{figure}{0}
\setcounter{table}{0}
\clearpage
\setcounter{page}{1}
\setcounter{equation}{0}
\maketitlesupplementary
\section{Detailed Proposed Method}
\textbf{Notation.} In this paper, we use the following notation conventions: Matrices and vectors are indicated with boldface capital and lowercase letters, \emph{e.g.}, $\bm{X}$ and $\bm{x}$, respectively. Tensors are represented using boldface calligraphic script, denoted as $\bmmc{X}$. 
\subsection{Inter-projection Correlation-aware \texorpdfstring{$\bm{P_t}$}~~Update. } 
Considering the importance of incorporating the information from the previous projection into the current update, we propose calculating $\bm{P}_t$ via solving the optimization problem:
\begin{equation}
\min\limits_{\bm{P}_t} \underbrace{\mathrm{\rm MSE}(\hat{\bm{G}}_t, \bm{G}_t)}_{\text{reconstruction term}} \underbrace{(1-\mathrm{CosSim}(\hat{\bm{M}}_{t-1}, \bm{G}_t))}_{\text{direction term}},
\end{equation}
where $\mathrm{CosSim}(\cdot, \cdot)$ and  $\mathrm{\rm MSE}(\cdot,\cdot)$ return the cosine similarity and mean squared error, respectively. To simplify notation, the notation without the subscript $t$ represents a general form of the optimization problem, \emph{i.e.},
\begin{equation}
\label{eq:obj_general}
\min\limits_{\bm{P}} \underbrace{\mathrm{\rm MSE}(\hat{\bm{G}}, \bm{G})}_{\text{reconstruction term}}  \underbrace{(1-\mathrm{CosSim}(\hat{\bm{M}}, \bm{G}))}_{\text{direction term}},
\end{equation}
where $\bm{P} \in \mathbb{R}^{n \times r}$, $\hat{\bm{G}} \in \mathbb{R}^{m \times n} = \bm{G}\bm{P}\bm{P}^\top$ and $\hat{\bm{M}} \in \mathbb{R}^{m \times n} = \bm{M}^{\rm proj}\bm{P}^\top$ are the full-rank estimates of gradient and first-order moment projected back from the low-rank subspace, respectively. Notably, the \textit{reconstruction term} is introduced to minimize the reconstruction error for gradients incurred by projection -- achieving the similar goal that SVD essentially aims for, and \textit{direction term} encourages the consistency of optimization direction after restoring from low-rank subspace. 

To solve Eqn. \ref{eq:obj_general} as a non-convex optimization problem, we propose using stochastic gradient descent to iteratively update $\bm{P}_t$ as follows:
\begin{equation}
\label{eq:update_proj}
\begin{aligned}
\bm{P} := &\bm{P}- \eta (\frac{\partial \mathrm{\rm MSE}(\hat{\bm{G}}, \bm{G})}{\partial \bm{P}}(1-\mathrm{CosSim}(\hat{\bm{M}}, \bm{G}) + \\  
          & \quad\frac{\partial \mathrm{CosSim}(\hat{\bm{M}}, \bm{G})}{\partial \bm{P}} \mathrm{\rm MSE}(\hat{\bm{G}}, \bm{G})).
\end{aligned}
\end{equation}

Here, $\eta$ represents learning rate, set to 0.1 by default. The gradient expressions for the reconstruction term and the direction term are derived as follows:

\textbf{Gradient of Reconstruction term (MSE).}
\begin{equation}
\begin{aligned}
\label{eq:mse}
\frac{\partial \mathrm{\rm MSE}(\hat{\bm{G}}, \bm{G})}{\partial \bm{P}}&=\frac{\partial}{\partial \bm{P}} \left( \frac{1}{mn} \mathrm{tr}((\hat{\bm{G}} - \bm{G})^\top (\hat{\bm{G}} - \bm{G})) \right)\\
&=\frac{2}{m n} (\hat{\bm{G}}^\top \bm{G} \bm{P}-2\bm{G}^\top \bm{G}\bm{P}+\bm{G}^\top \hat{\bm{G}}\bm{P}),
\end{aligned}
\end{equation}
where $\mathrm{tr}(\cdot)$ represents the trace of a matrix.

\textbf{Gradient of Direction Term (CosSim).} 

Given the cosine similarity between matrices $\hat{\bm{M}}$ and $\bm{G}$, defined as:
\begin{equation}
\begin{aligned}
\mathrm{CosSim}(\hat{\bm{M}}, \bm{G})&=\frac{1}{m} \sum_{i=1}^{m} \mathrm{CosSim}(\hat{\bm{M}}_i, \bm{G}_i)\\
&=\frac{1}{m} \sum_{i=1}^{m}\frac{\langle \hat{\bm{M}}_i, \bm{G}_i \rangle}{\|\hat{\bm{M}}_i\| \|\bm{G}_i\|}.
\end{aligned}
\end{equation}
Applying the chain rule to compute the gradient of $\mathrm{CosSim}(\hat{\bm{M}}, \bm{G})$ with respect to $\bm{P}$:
\begin{equation}
\begin{aligned}
\label{eq:cos}
&\frac{\partial \mathrm{CosSim}(\hat{\bm{M}}, \bm{G})}{\partial \bm{P}} = (\frac{\partial \mathrm{CosSim}(\hat{\bm{M}}, \bm{G})}{\partial \hat{\bm{M}}})^\top \frac{\partial \hat{\bm{M}}}{\partial \bm{P}} \\
&\quad=\frac{1}{m} \sum_{i=1}^{m} (\frac{\partial \mathrm{CosSim}(\hat{\bm{M}}, \bm{G})}{\partial \hat{\bm{M}}_i} )^\top \bm{M}^{\rm proj}_i \\
&\quad= \frac{1}{m} \sum_{i=1}^{m} (\frac{\bm{G}_i}{\|\hat{\bm{M}}_i\| \|\bm{G}_i\|} - \frac{\hat{\bm{M}}_i \langle \hat{\bm{M}}_i, \bm{G}_i \rangle}{\|\hat{\bm{M}}_i\|^3 \|\bm{G}_i\|})^\top \bm{M}^{\rm proj}_i,
\end{aligned}
\end{equation}
where $\|\cdot\|$ represents the Euclidean norm, $\langle \cdot, \cdot \rangle$ denotes the inner product, $\hat{\bm{M}}_i, \bm{G}_i, \bm{M}^{\rm proj}_i$ denotes the $i$-th row of $\hat{\bm{M}}, \bm{G}, \bm{M}^{\rm proj}$. 

Incorporating the gradient expressions above, we derive the final update formula for $\bm{P}$ as follows:
\begin{equation}
\label{eq:update_proj_final}
\begin{aligned}
\bm{P} := &\bm{P}- \eta (\frac{2}{m n} (\hat{\bm{G}}^\top \bm{G} \bm{P}-2\bm{G}^\top \bm{G}\bm{P}+\bm{G}^\top \hat{\bm{G}}\bm{P})\\
        &(1-\frac{1}{m} \sum_{i=1}^{m}\frac{\langle \hat{\bm{M}}_i, \bm{G}_i \rangle}{\|\hat{\bm{M}}_i\| \|\bm{G}_i\|}) + \\
        &\frac{1}{m} \sum_{i=1}^{m} (\frac{\bm{G}_i}{\|\hat{\bm{M}}_i\| \|\bm{G}_i\|} - \frac{\hat{\bm{M}}_i \langle \hat{\bm{M}}_i, \bm{G}_i \rangle}{\|\hat{\bm{M}}_i\|^3 \|\bm{G}_i\|})^\top \bm{M}^{\rm proj}_i \\
        &\frac{1}{mn} \mathrm{tr}((\hat{\bm{G}} - \bm{G})^\top (\hat{\bm{G}} - \bm{G}))).
\end{aligned}
\end{equation}

\subsection{Theory Analysis.}
COAP proposes to approximately keep the same low-rank subspace across iterations. Intuitively, this requires the low-rank subspace to remain stable across iterations, which can be bounded in terms of the learning rate and Lipchitz constant. Theoretical justification composed of the following steps:
\begin{itemize}
    \item Bounded update will lead to bounded change of the subspace.
    \item Equation \ref{eq:obj_general} finds a subspace not much worse than the true subspace.
\end{itemize}
We aim to demonstrate that this procedure guarantees an approximation to the underlying true subspace.
Let $\bm{P}_t^{\natural}$ be the true subspace. Let $\bm{P}_t^{\rm MSE}$ and $\bm{P}_t^{\rm SIM}$ be the solutions of the MSE loss and CosSim loss respectively, then
\begin{equation}
\begin{aligned}
\bm{P}_t^{\rm MSE} &= \bm{P}_{t-1}\\
\bm{P}_t^{\rm SIM} &= \text{argmax}\left \langle\bm{G}_{t}\bm{P}_{t}, \bm{M}_{t-1}^{proj} \right \rangle ,
\end{aligned}
\end{equation}
where $\bm{M}_{t-1}^{\rm proj}=\beta\bm{M}_{t-2}^{\rm proj}+(1-\beta)\bm{P}_{t}^T\bm{G}_t=\sum_{i=0}^{t-1}\beta^{i}(1-\beta)\bm{P}_{t-i}^T\bm{G}_{t-i}$, $\text{argmax}\left \langle\cdot,\cdot\right \rangle $ maximizes an inner product between two entities. Therefore,
\begin{equation}
\begin{aligned}
\bm{P}_t^{\rm SIM} &= \sum_{i=0}^{t-1}\beta^{i}(1-\beta)\text{argmax}\left \langle \bm{P}_t^T\bm{G}_{t},\bm{P}_{t-i}^T\bm{G}_{t-i} \right \rangle \\
&=\sum_{i=0}^{t-1}\beta^{i}(1-\beta)\bm{P}_{t-i}
\end{aligned}
\end{equation}
\begin{theorem}
Assume that the gradient is Lipchitz with respect to the weight with Lipchitz constant $L$. Assume $\kappa_1\doteq\frac{\sigma_1}{\sigma_{r+1}}>1$ and $\kappa_r\doteq\frac{\sigma_r}{\sigma_{r+1}}>1$ to be the ratio between the $1$-th and $r$-th to the $r+1$-th singular value of the gradient $\bm{G}$ respectively. Then the proposed procedure in Equation \ref{eq:obj_general} recovers a subspace well approximating the true subspace.
\begin{equation} \begin{aligned}
&\left \|  \bm{P}_t-\bm{P}_{t}^{\natural}\right \|_F\\
&~~\le c(t\mod(\lambda\times T_u))\frac{r\kappa_1+\min(m,n)-r}{\kappa_r-1} L \eta.
\end{aligned} \end{equation}
\end{theorem}
With learning rate $\eta$, Lipchitz property of the gradient gives
\begin{equation} \begin{aligned}
\left \| \bm{G}_{t-1}-\bm{G}_{t}\right \|_2&\le L\left \|\bm{W}_{t-1}-\bm{W}_{t}\right \|_2\\
&=L\left \|\eta\bm{G}_{t-1}^{proj}\right \|_2\le L\eta\left \| \bm{G}_{t-1}\right \|_2.
\end{aligned} \end{equation}
Consequently, the optimal subspace $\bm{P}_t^{\natural}$ is also close enough to the previous optimal subspace $\bm{P}_{t-1}^{\natural}$:
\begin{equation} \begin{aligned}
\left \|\bm{P}_{t-1}^{\natural}-\bm{P}_{t}^{\natural}\right \|_F\le\frac{r\kappa_1+\min(m,n)-r}{\kappa_r-1}L\eta.
\end{aligned} \end{equation}
When $t\mod(\lambda\times T_u)=0$, the proposed algorithm updates $\bm{P}_t=\bm{P}_{t}^{\natural}$. Otherwise, we show that $\bm{P}_t$ highly approximates $\bm{P}_{t}^{\natural}$,
\begin{equation} \begin{aligned}
&\left \|\bm{P}_t-\bm{P}_{t}^{\natural}\right \|_F\le\left \|\bm{P}_{t-1}^{\natural}-\bm{P}_{t}^{\natural}\right \|_F+\left \|\bm{P}_t-\bm{P}_{t-1}^{\natural}\right \|_F\\
&\le\left \|\bm{P}_{t-1}^{\natural}-\bm{P}_{t}^{\natural}\right \|_F+\left \|\bm{P}_t^{\rm MSE}-\bm{P}_{t-1}^{\natural}\right \|_F+\left \|\bm{P}_t^{\rm SIM}-\bm{P}_{t-1}^{\natural}\right \|_F\\
&\le\left \|\bm{P}_{t-1}^{\natural}-\bm{P}_{t}^{\natural}\right \|_F+(2-\beta) \left \|\bm{P}_{t-1}-\bm{P}_{t-1}^{\natural}\right \|_F\\
&\quad+(1-\beta)\sum_{i=2}\beta^i\left \|\bm{P}_{t-i}-\bm{P}_{t-1}^{\natural}\right \|_F\\
&\le\left \|\bm{P}_{t-1}^{\natural}-\bm{P}_{t}^{\natural}\right \|_F+2\left \|\bm{P}_{t-1}-\bm{P}_{t-1}^{\natural}\right \|_F\\
&\quad+(1-\beta)\sum_{i=2}\beta^i\left \|\bm{P}_{t-i}-\bm{P}_{t-1}\right \|_F\\
&\le 2^{(t\mod(\lambda\times T_u))}\frac{r\kappa_1+\min(m,n)-r}{\kappa_r-1} L \eta
\end{aligned} \end{equation}
\subsection{Hyper-parameter Settings.}
Table~\ref{tbl:hyper-param} presents the detailed hyper-parameters used in all experiments of COAP. Except for the Update Interval ($T_u$) and Re-project Factor ($\lambda$) specific to COAP, all other parameters remain consistent with the corresponding baselines. The optimizer, rank settings, learning rate, GPU configuration, and mixed precision are listed for each experiment. Hyper-parameters for different models, including LDM, SiT-XL/2, ControlNet SDXL, and LLaMA series, are provided to ensure reproducibility.
\begin{table*}[t]
\centering
\caption{Table 1: Hyper-parameter settings for all experiments.}
\resizebox{1\linewidth}{!}{\begin{tabular}{lcccccccc}
\toprule
Optimizer                          & Exprement                                   & \begin{tabular}[c]{@{}c@{}}Rank\\ $r$\end{tabular} & \begin{tabular}[c]{@{}c@{}}Rank Ratio\\ $\alpha$\end{tabular} & \begin{tabular}[c]{@{}c@{}}Update Interval\\ $T_u$\end{tabular} & \begin{tabular}[c]{@{}c@{}}Re-project Factor\\ $\lambda$\end{tabular} & \begin{tabular}[c]{@{}c@{}}Learning Rate\\ $\eta$\end{tabular} & GPU    & Mixed-precision \\
\midrule
\multirow{5}{*}{\begin{tabular}[c]{@{}c@{}}Adafactor  \\ 
(COAP)\end{tabular}}   & Table 1. {\href{https://github.com/CompVis/latent-diffusion}{Pre-training LDM}}                   & -    & 2          & 16              & 10              & $2\times10^{-5}$      & 8$\times$V100 & FP32            \\
\cmidrule{2-9}
                                   & Table 2. \href{https://github.com/sihyun-yu/REPA}{Pre-training SiT-XL/2}& 512  & -          & 200             & 5               & $1\times10^{-4}$       & 8$\times$H100 & FP16            \\
\cmidrule{2-9}
                                   & \multirow{3}{*}{Table 3. ControlNet   SDXL} & -    & 2          & 8               & 10              & $1\times10^{-5}$     & 8$\times$H100 & BF16            \\
                                   &                                             & -    & 4          & 8               & 10              &$1\times10^{-5}$       & 8$\times$H100 & BF16            \\
                                   &                                             & -    & 8          & 8               & 10              & $1\times10^{-5}$       & 8$\times$H100 & BF16            \\
\midrule
\midrule
\multicolumn{1}{l}{\multirow{5}{*}{\begin{tabular}[c]{@{}l@{}}AdamW \\ (COAP)\end{tabular}}}     & Table 1. {\href{https://github.com/CompVis/latent-diffusion}{Pre-training LDM}}
                  & -    & 2          & 16              & 10              & $2\times10^{-5}$     & 8$\times$V100 & FP32            \\
\cmidrule{2-9}
                                   & Table 2. \href{https://github.com/sihyun-yu/REPA}{Pre-training SiT-XL/2}                & 512  & -          & 30              & 10              & $1\times10^{-4}$       & 8$\times$H100 & FP16            \\
\cmidrule{2-9}
                                   & Table 5. \href{https://github.com/jiaweizzhao/GaLore}{Pre-training LLaMA-1B}               & 512  & -          & 40              & 5               & $1\times10^{-2}$          & 8$\times$H100 & BF16            \\
\cmidrule{2-9}
                                   & Table 5. \href{https://github.com/jiaweizzhao/GaLore}{Pre-training LLaMA-7B}              & 1024 & -          & 100             & 1               & $5\times10^{-3}$         & 8$\times$H100 & BF16            \\
\cmidrule{2-9}
                                   & Table 6. \href{https://github.com/haotian-liu/LLaVA}{Fine-tuning LLaVA-v1.5-7B}         & -    & 4          & 32              & 1               & $2\times10^{-5}$       & 1$\times$A100 & BF16     \\    
\bottomrule
\end{tabular}
}
\label{tbl:hyper-param}
\end{table*}

\subsection{Adafactor with COAP.}
Adafactor is an adaptive gradient optimization algorithm designed to reduce memory usage while maintaining training efficiency. Unlike Adam, which stores full-matrix second-moment estimates, Adafactor factorizes the second-moment accumulation to save memory, making it particularly suitable for training large-scale models.
However, first-moment remains crucial for stabilizing training and accelerating convergence. Without it, optimization can become unstable, especially in deep networks where gradients vary significantly. The first moment helps smooth updates and improves the overall training dynamics.
Algorithm~\ref{alg:COAP_Adafactor} describes the Adafactor-based training procedure, by projecting the gradient and its first-moment estimate into a lower-dimensional subspace, we effectively compress the momentum while retaining its benefits. This approach allows us to maintain training stability with reduced memory overhead, making it highly efficient for large-scale models.
\begin{algorithm}
\begin{algorithmic}
\setlength{\itemindent}{-1em} 
\small
   \caption{Adafactor with COAP}
   \label{alg:COAP_Adafactor}
   \STATE {\bfseries Input:} Weight matrix $\bm{W} \in \mathbb{R}^{m \times n}$, Learning rate $\eta$, Rank $r$, \\
   \hspace{2em} Betas $[\beta_1, \beta_2]$, Update interval $[\lambda, T_{\rm u}]$, Decay rate $\gamma$.\\
   \STATE {\bfseries Initialize:} $\bm{M}_0^{\rm proj} \in \mathbb{R}^{m \times r} \leftarrow 0$, $\bm{V}_0^{\rm proj}\in \mathbb{R}^{m \times r} \leftarrow 0$, 
   $t\leftarrow 0$ \\
   \quad\quad\quad~$\bm{R}_0\in \mathbb{R}^{m \times 1} \leftarrow 0$, $\bm{C}_0\in \mathbb{R}^{1 \times r} \leftarrow 0$.
   \STATE {\bfseries Randomly Initialize:} $\bm{P}_0 \in \mathbb{R}^{n \times r} $\\
   \STATE \textbf{Compute:} $\bm{P}_0\leftarrow (\bm{P}_0,\bm{G}_0)$$\quad\quad\triangleright$ Occasional Low-cost SVD\\
   \hspace{-1em}\For{ $t ~{\rm in}~ [1, 2, \cdots]$}{
        \STATE \hspace{1em}\textbf{Compute:} gradient $\bm{G}_t$ of $\bm{W}_t$ in the loss function. \\
        \If {$t \mod T_{\rm u} = 0$} {
            \If {$t \mod (\lambda\times T_{\rm u}) = 0$} {
                \STATE\hspace{1em} 
                $\triangleright$ Occasional Low-cost SVD\\
                \textbf{Compute:} $\bm{P}_t\leftarrow (\bm{P}_{t-1},\bm{G}_t) $ \\
                
            }
            \Else {
                \STATE\hspace{1em} \textbf{Update:}$\bm{P}_{t}\leftarrow(\bm{P}_{t-1},\bm{G}_t,\bm{M}_{t-1}) $\hfill$\triangleright$ Eqn.~\ref{eq:obj_general}\\
            }
        }
        \Else {
            $\bm{P}_t\leftarrow \bm{P}_{t-1}$
        }
        \comments{$\triangleright$Project gradient and moments into low-rank space.} \\
        $\beta_2 \leftarrow 1-t^\gamma$\\
        $\bm{G}_t^{\rm proj} \leftarrow \bm{G}_t\bm{P}_t$ \\
        $\bm{M}_t^{\rm proj} \leftarrow \beta_1\bm{M}_{t-1}^{\rm proj}+(1-\beta_1)\bm{G}_t^{\rm proj}$ \\
        $\bm{R}_t = \beta_2 \bm{R}_{t-1} + (1-\beta_2)\cdot \mathrm{Sum}({\bm{G}_t^{\rm proj}}^2, -1)$\\
        $\bm{C}_t = \beta_2 \bm{C}_{t-1} + (1-\beta_2)\cdot  \mathrm{Sum}({\bm{G}_t^{\rm proj}}^2, -2)$ \\
        $\hat{\bm{V}_t} = \sqrt{\frac{\mathrm{Mean}(\bm{R}_t, -1)}{\bm{R}_t\bm{C}_t}} $\\
       
        \comments{$\triangleright$Calculate the bias correction term in low-rank space.} \\
        $\Delta \bm{W}_t^{\rm proj}\leftarrow \beta_1\bm{M}_t^{\rm proj}+(1-\beta_1)\eta\hat{\bm{V}_t}\odot\bm{G}_t^{\rm proj}  $ \\
        \comments{$\triangleright$Restore $\Delta \bm{W}_t^{\rm proj}$ to original space and update $\bm{W}$.} \\
        $\bm{W}_t \leftarrow \bm{W}_{t-1} -   \Delta \bm{W}_t^{\rm proj}\bm{P}_t^\top$ \\
       }
    \STATE {\bfseries Return:} updated $\bm{W}$ \\
\end{algorithmic}
\end{algorithm}
\subsection{Extension to CONV Layer.}
To enhance the generality and applicability of our algorithm, it is essential to extend support to higher-dimensional weight tensors, which are prevalent in architectures such as Convolutional Neural Networks (CNNs). While the most straightforward approach would be to reshape CNN weights into matrices and apply the same low-rank space construction method used for matrix operations, this naive strategy would inevitably result in the loss of intrinsic spatial characteristics inherent to CNNs~\cite{liu2012tensor}. 

For a convolutional layer with a weight tensor $\bmmc{W} \in \mathbb{R}^{O \times I \times K_1 \times K_2}$, where $I$ and $O$ are the number of input channels and output channels, respectively, and $K_1$ and $K_2$ are the kernel sizes, we use Tucker-2 decomposition \cite{tucker1966some} as the factorization method. In this scenario, $\bmmc{W}$ can be represented with a core tensor $\bmmc{C}$ and two factor matrices ($\bm{U}_1$ and $\bm{U}_2$) along each mode as follows:
\begin{equation}
\label{eq:tucker2}
    \bmmc{W} = \bmmc{C} \times_1 \bm{U}_1 \times_2 \bm{U}_2,
\end{equation}
where ``$\times_n$" denotes the $n$-mode product. Specifically, $\bm{U}_1 \in \mathbb{R}^{O \times r_O}$ represents the left singular vectors of the mode-1 unfolding of $\bmmc{W}$, denoted as $\bm{W}_{(1)}$, \emph{i.e.}, $\bm{W}_{(1)} = \bm{U}_1 \bm{\Sigma}_1 \bm{V}_1^\top$. Similarly, $\bm{U}_2 \in \mathbb{R}^{I \times r_I}$ represents the left singular vectors of the mode-2 unfolding of $\bmmc{W}$, denoted as $\bm{W}_{(2)}$, \emph{i.e.}, $\bm{W}_{(2)} = \bm{U}_2 \bm{\Sigma}_2 \bm{V}_2^\top$. The core tensor $\bmmc{C}$ is obtained by projecting $\bmmc{W}$ onto the subspaces spanned by $\bm{U}_1$ and $\bm{U}_2$, i.e.,
\begin{equation}
\label{eq:tensor_core}
    \bmmc{C} = \bmmc{W} \times_1 \bm{U}_1^\top \times_2 \bm{U}_2^\top,
\end{equation}
where $\bmmc{C} \in \mathbb{R}^{r_O \times r_I \times K_1 \times K_2}$. Here, $r_O$ and $r_I$ are the Tucker-2 tensor ranks, determining the dimensionality of the factor matrices and the core tensor. Initially, the values for $\bm{U}_1$, $\bm{U}_2$, and $\bmmc{C}$ are typically obtained using Higher-Order Singular Value Decomposition (HOSVD)~\cite{de2000multilinear}. To refine these initial values and achieve the final decomposition, the Alternating Least Squares (ALS)~\cite{kroonenberg1980principal} method is employed. This iterative optimization technique alternates between updating the core tensor $\bmmc{C}$ and the factor matrices $\bm{U}_1$ and $\bm{U}_2$ to minimize the reconstruction error. 

Algorithm~\ref{alg:COAP_CNN} outlines the Adam-based training procedure, integrating the proposed $\bm{P}_t$ update method for convolutional layers.

Typically, for a convolutional layer with a weight tensor $\bmmc{W} \in \mathbb{R}^{O \times I \times K_1 \times K_2}$, $I$ and $O$ are significantly larger than the kernel sizes $K_1$ and $K_2$ (i.e., $I, O \gg K_1, K_2$). Therefore, we propose using the format of Tucker-2 decomposition to handle CNNs while employing our own decomposition method for the actual factorization. According to Eq. \ref{eq:tucker2} and Eq. \ref{eq:tensor_core}, the low-rank projection space of $\bmmc{G}_t$ becomes $[\bm{P}_{O_t} \in \mathbb{R}^{O \times r_O}, \bm{P}_{I_t} \in \mathbb{R}^{I \times r_I}]$. The gradient $\bmmc{G}_t$ in the low-rank space is $\bmmc{G}_t^{\rm proj} = \bmmc{G}_t \times_1 \bm{P}_{O_t}^\top \times_2 \bm{P}_{I_t}^\top$, and the restored tensor from the low-rank space is $\hat{\bmmc{G}_t} = \bmmc{G}_t^{\rm proj} \times_1 \bm{P}_{O_t} \times_2 \bm{P}_{I_t}$. Here, $\bm{P}_{O_t}$ and $\bm{P}_{I_t}$ can be updated according to Eqn. \ref{eq:update_proj_final} and Occasional Low-cost SVD on the mode-1 and mode-2 unfolding of tensor $\bmmc{G}$, respectively.

\begin{algorithm}
\begin{algorithmic}
\small
   \caption{Adam with COAP (CONV)}
   \label{alg:COAP_CNN}
   \STATE {\bfseries Input:} Weight tensor $\bmmc{W} \in \mathbb{R}^{O \times I \times K_1 \times K_2}$, Learning rate $\eta$,  \\
   \hspace{2.8em} Rank ratio $\alpha$, Betas $[\beta_1, \beta_2]$, Update interval $[\lambda, T_{\rm u}]$.\\
   \STATE {\bfseries Initialize:} $r_O = O^{\frac{1}{\sqrt{\alpha}}}$, $r_I = I^{\frac{1}{\sqrt{\alpha}}}$, $t\leftarrow 0$,\\
   $\quad\quad\quad\quad~~\bmmc{M}_0^{\rm proj} \in \mathbb{R}^{r_O \times r_I \times K_1 \times K_2} \leftarrow 0$,\\ $\quad\quad\quad\quad~~\bmmc{V}_0^{\rm proj}\in \mathbb{R}^{r_O \times r_I \times K_1 \times K_2} \leftarrow 0$ \\
   \STATE {\bfseries Randomly Initialize:} $\bm{P}_O \in \mathbb{R}^{O \times r_O}, \bm{P}_I \in \mathbb{R}^{I \times r_I} $\\
   \STATE \textbf{Define:} $\bm{G}_{O_t} \leftarrow \text{reshape}(\bmmc{G}_t, [O, IK_1K_2])$,\\
                            $\quad\quad\quad~~\bm{G}_{I_t} \leftarrow \text{reshape}(\bmmc{G}_t, [I, OK_1K_2])$\\
   \STATE \textbf{Compute:} $\bm{P}_{O_0}\leftarrow (\bm{P}_O,\bm{G}_{O_0}), \bm{P}_{I_0}\leftarrow (\bm{P}_I,\bm{G}_{I_0})$  \\ 
   \hfill$\triangleright$ Occasional Low-cost SVD  \\
   \For{ $t ~{\rm in}~ [1, 2, \cdots]$}{
        \STATE \textbf{Compute:} gradient $\bm{G}_t$ of $\bm{W}_t$ in the loss function. \\
        \If {$t \mod T_{\rm u} = 0$} {
            \If {$t \mod (\lambda\times T_{\rm u}) = 0$} {
                \STATE \textbf{Compute:} $\bm{P}_{O_t}\leftarrow (\bm{P}_{O_{t-1}},\bm{G}_{O_t})$,\\
                $\quad\quad\quad\quad~~\bm{P}_{I_t}\leftarrow (\bm{P}_{I_{t-1}},\bm{G}_{I_t})$ \\
                \hfill$\triangleright$ Occasional Low-cost SVD  \\
            }
            \Else {
                \STATE \textbf{Update:} $\bm{P}_{O_t}\leftarrow (\bm{P}_{O_{t-1}},\bm{G}_{O_t},\bm{M}_{O_{t-1}}) $ \hfill $\triangleright$ Eqn.~\ref{eq:update_proj_final}\\
                \STATE \textbf{Update:} $\bm{P}_{I_t}\leftarrow (\bm{P}_{I_{t-1}},\bm{G}_{I_t},\bm{M}_{I_{t-1}}) $ \hfill $\triangleright$ Eqn.~\ref{eq:update_proj_final}\\
            }
        }
        \Else {
            $\bm{P}_{O_t}\leftarrow \bm{P}_{O_{t-1}}$, $\bm{P}_{I_t}\leftarrow \bm{P}_{I_{t-1}}$
        }
        \comments{$\triangleright$~ Project gradient and moments into low-rank space.} \\
        $\bmmc{G}_t^{\rm proj} \leftarrow \bmmc{G}_t\times_1 \bm{P}_{O_t}^\top \times_2 \bm{P}_{I_t}^\top$ \\
        $\bmmc{M}_t^{\rm proj} \leftarrow \beta_1\bmmc{M}_{t-1}^{\rm proj}+(1-\beta_1)\bmmc{G}_t^{\rm proj}$ \\
        $\bmmc{V}_t^{\rm proj} \leftarrow \beta_2\bmmc{V}_{t-1}^{\rm proj}+(1-\beta_2)(\bmmc{G}_t^{\rm proj})^2$ \\
        \comments{$\triangleright$~ Calculate the bias correction term in low-rank space.} \\
        $\Delta \bmmc{W}_t^{\rm proj}\leftarrow \frac{\bmmc{M}_t^{\rm proj}/(1-\beta_1^t)}{\sqrt{\bmmc{V}_t^{\rm proj}/(1-\beta_2^t)}+\epsilon }$ \\
        \comments{$\triangleright$~ Restore $\Delta \bmmc{W}_t^{\rm proj}$ to original space and update $\bmmc{W}$.} \\
        $\bmmc{W}_t \leftarrow \bmmc{W}_{t-1} - \eta  \Delta \bmmc{W}_t^{\rm proj}\times_1 \bm{P}_{O_t} \times_2 \bm{P}_{I_t}$ \\
       }
    \STATE {\bfseries Return:} updated $\bmmc{W}$ \\
\end{algorithmic}
\end{algorithm}
\subsection{Impact of Low-rank Matrix Projection Formats on CNN Models Performance.}
\begin{figure}[t]
    \centering
    \includegraphics[width=\linewidth]{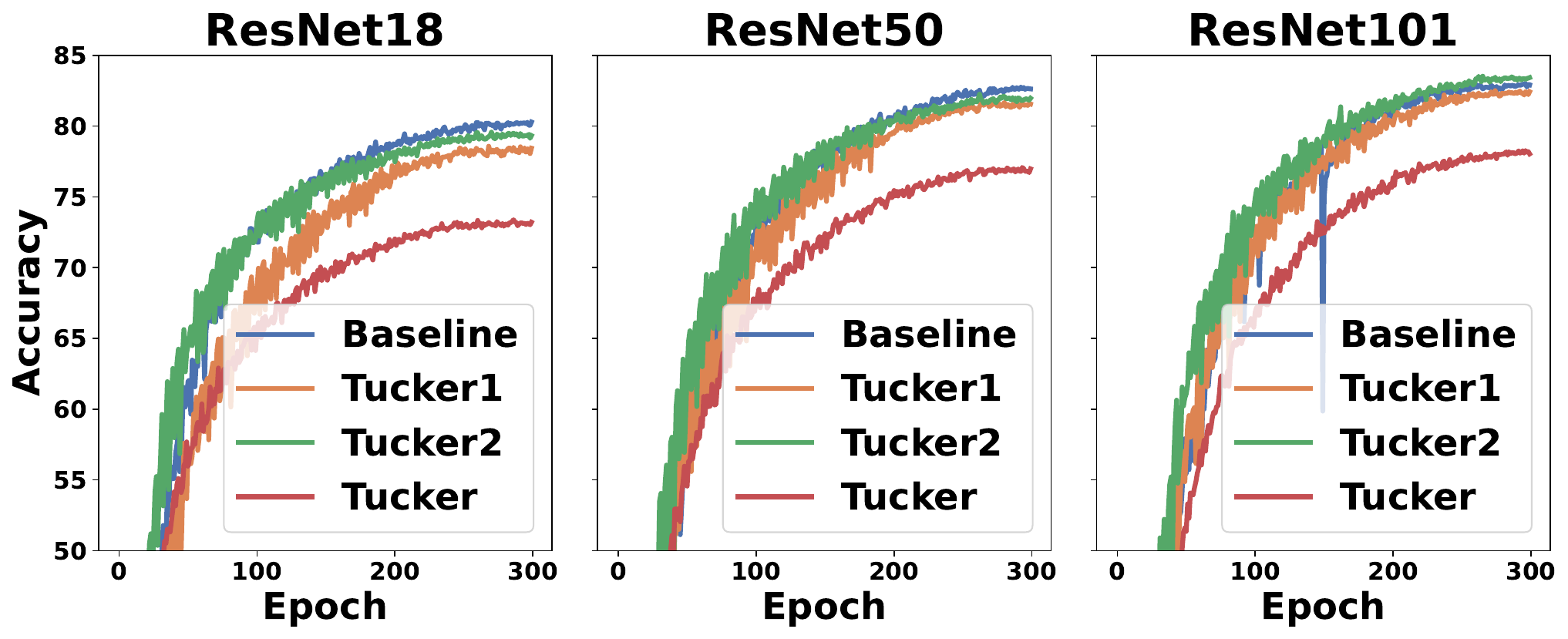}
    \caption{The comparison of Top-1 accuracy for ResNet under different low-rank projection formats is conducted with a rank ratio of 4. 
The models are trained for 100 epochs on the CIFAR-100 dataset.}
    \label{fig:tucker}
\end{figure}

We compare different formats of Tucker decomposition, \emph{i.e.}, Tucker-1, Tucker-2, and Tucker. In this context, the default Tucker format applies projections along all dimensions of a tensor. For instance, if the tensor is 4-dimensional, it requires 4 projection matrices. Tucker-2, on the other hand, uses only two projection matrices, while Tucker-1 requires just one, making it a variant of SVD. 

As shown in the Fig.~\ref{fig:tucker}, Tucker-2 achieves performance closest to the baseline across different scales of ResNet models. Thus, we select Tucker-2 as the primary format for computing convolution projection matrices.

\section{Experimental Results}
The COAP optimizer is versatile and can be applied to various model training scenarios. We have provided training results for DDPM and ControlNet, and this method is also suitable for other application scenarios~\cite{zhang2024defensive,zhang2024generate,zhang2024id,yang2024moei2compressingmixtureexperts}.
\begin{table}[t]
\centering
\caption{Pre-training DDPM on CIFAR-10 and CelebA-HQ datasets on 8$\times$V100. FID scores are reported, along with the GPU memory usage of optimizer states in FP32 format.}
\label{tbl:ddpm}
\small
\begin{tabular}{lcccc}
\toprule
Dataset                    & Method   & \begin{tabular}[c]{@{}c@{}}Rank \\ Ratio\end{tabular}    & \begin{tabular}[c]{@{}c@{}}Optimizer \\ Mem. (MB)$\downarrow$\end{tabular} & FID$\downarrow$            \\ 
\midrule
\multirow{6}{*}{\begin{tabular}[l]{@{}@{}l@{}}CIFAR-10 \\ ($32 \times 32$) \\ 800K steps\end{tabular}}
                           & AdamW    &-     & 272.72           & 5.42           \\ 
                           & GaLore   &1.5    &302.43         & 6.09           \\ 
                           & \textbf{COAP} & 1.5  & \textbf{214.66}  & 5.66           \\ 
\cmidrule{2-5}
                           & Adafactor   &- & 222.71           & 5.43           \\ 
                           & GaLore   &1.5  & 196.11           & 7.14           \\ 
                           & \textbf{COAP}  &1.5 &  \textbf{180.87}           & \textbf{5.41}  \\ 
\midrule                         
\multicolumn{1}{l}{\multirow{6}{*}{\begin{tabular}[l]{@{}@{}l@{}}CelebA-HQ \\ ($256 \times 256$) \\460K steps\end{tabular}}} 
                         & AdamW   &-       & 867.26           & 12.82          \\ 
\multicolumn{1}{l}{}     & GaLore   &2    & 562.56           & 27.95          \\ 
\multicolumn{1}{l}{}     & \textbf{COAP}  &2  & \textbf{525.18}           & 17.37          \\ 
\cmidrule{2-5}
\multicolumn{1}{l}{}     & Adafactor   &- & 714.64           & 12.38          \\ 
\multicolumn{1}{l}{}     & GaLore   &2   & 549.27           & 19.12          \\ 
\multicolumn{1}{l}{}      & \textbf{COAP} &2  & \textbf{447.59}  & \textbf{12.30} \\ 
\bottomrule
\end{tabular}
\end{table}
\subsection{Pre-training DDPM}
\textbf{Experimental Settings.} We implement DDPM based on the Diffusers~\footnote{\url{https://github.com/huggingface/diffusers/tree/main/examples/unconditional_image_generation}} from Hugging Face and conduct experiments on 8$\times$V100 GPUs following the training and evaluation settings in~\cite{ho2020denoising}. For CIFAR-10~\cite{krizhevsky2009learning}, the model is trained with a batch size of 128 for 800K steps. For CelebA-HQ~\cite{liu2015faceattributes}, the model is trained with a batch size of 64 for 460K steps. We generate 50K images to compute the FID (Frechet Inception Distance) with respect to the training dataset and the images generated with 1000 DDPM steps.

\textbf{Comparison Results.} Table~\ref{tbl:ddpm} presents the performance of our method and GaLore when compressing AdamW and Adafactor optimizers on the DDPM model. Our approach consistently outperforms GaLore on CIFAR-10 and CelebA-HQ datasets. Specifically, using the Adafactor optimizer, our method reduces FID by 1.7 and 6.8 compared to GaLore. Additionally, at compression rates of 1.2$\times$ and 1.6$\times$, our approach surpasses the baseline performance.


\subsection{Qualitative Comparisons}

We present qualitative results, showcasing images generated by models trained with COAP and other optimizers. These results provide a visual comparison that complements the quantitative analysis in the main paper. Comparisons for DDPM (CIFAR-10), DDPM (CelebA-HQ), LDM, SiT-XL/2, and ControlNet-XL are detailed in Tables~\ref{tbl:ddpm-cifar},~\ref{tbl:ddpm-celeba},~\ref{tbl:ldm-imagenet},~\ref{tbl:sit-imagenet}, and~\ref{tbl:controlnet-xl-pose}, respectively.


\begin{minipage}{\textwidth}
    \setlength{\tabcolsep}{2pt}
    \vspace{+5mm}
     \captionof{table}{Comparison of images generated by DDPM trained on CIFAR-10 with different Adam-based optimizers.}
    \label{tbl:ddpm-cifar}
    \resizebox{0.95\linewidth}{!}{\begin{tabular}{c|c|c}
        \toprule
       \multicolumn{1}{c|}{\begin{tabular}[c]{@{}c@{}}Adam (272.7 MB)\end{tabular}} &\multicolumn{1}{c|}{\begin{tabular}[c]{@{}c@{}}GaLore (302.4 MB)\end{tabular}} &\multicolumn{1}{c}{\begin{tabular}[c]{@{}c@{}}\textbf{COAP (214.7 MB)}\end{tabular}} \\
       \midrule
       \includegraphics[width=0.32\textwidth]{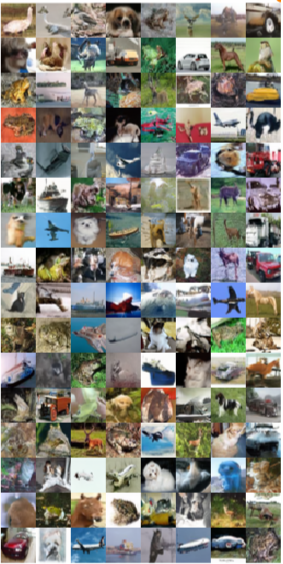} 
        & \includegraphics[width=0.32\textwidth]{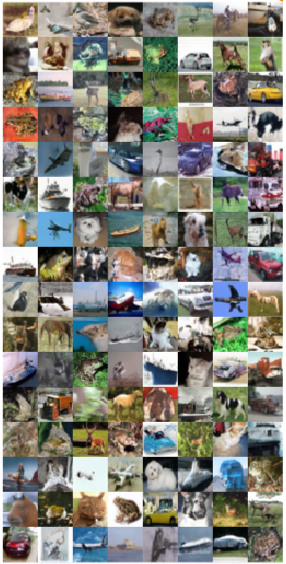} 
        &\includegraphics[width=0.32\textwidth]{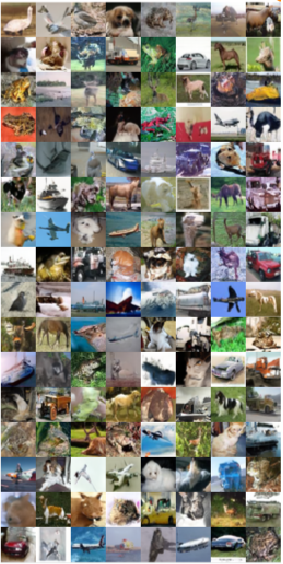} \\ 
        \bottomrule
    \end{tabular}}
\end{minipage}

\begin{table*}[ht]
    \centering
    \caption{Comparison of images generated by DDPM trained on CelebA-HQ with different Adafactor-based optimizers.}
    \label{tbl:ddpm-celeba}
    \begin{tabular}{c|c}
        \toprule
       \rotatebox{90}{Adafactor~~(714.6 MB)}&\includegraphics[width=0.9\textwidth]{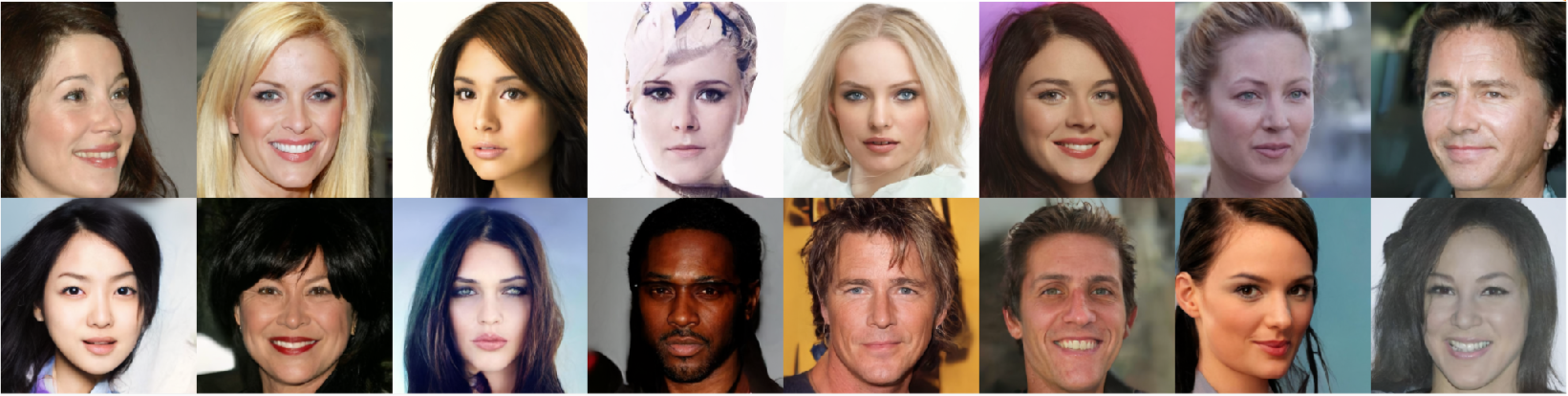} \\
       \midrule
        \rotatebox{90}{GaLore~~(549.3 MB)}&\includegraphics[width=0.9\textwidth]{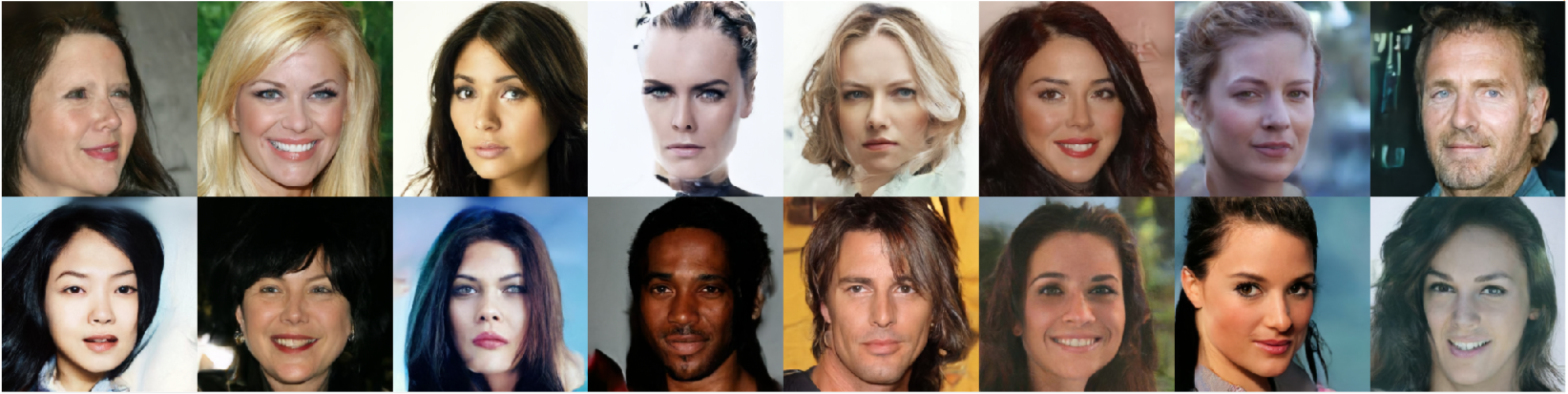} \\
        \midrule
        \rotatebox{90}{\textbf{COAP~~(447.6 MB)}}&\includegraphics[width=0.9\textwidth]{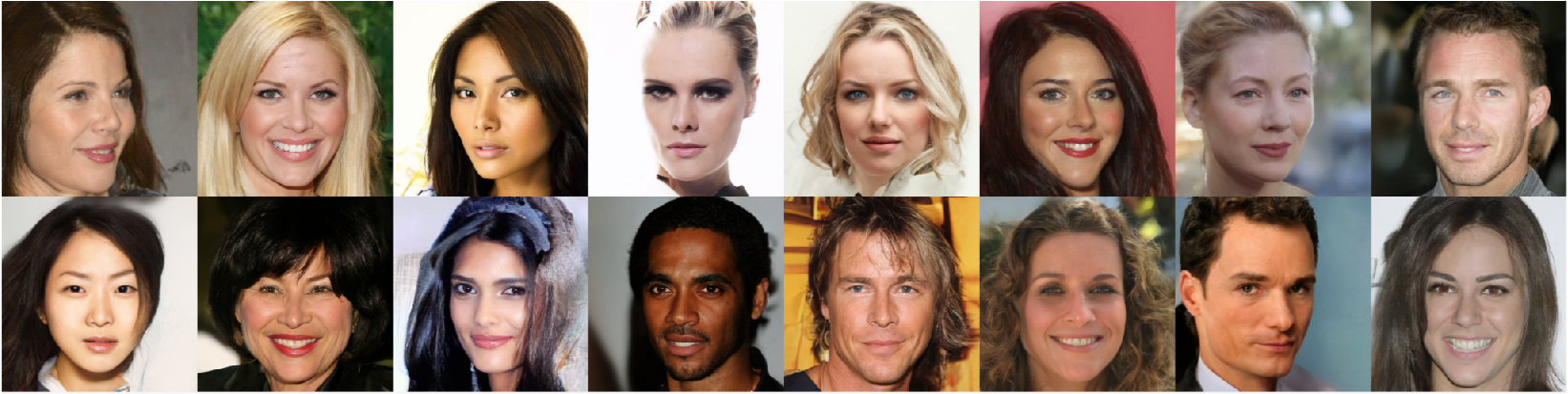} \\ 
        \bottomrule
    \end{tabular}
\end{table*}

\begin{table*}[ht]
    \centering
    \small
    \caption{Random class-conditional samples generated by LDM trained on the ImageNet dataset using the COAP optimizer.}
    \label{tbl:ldm-imagenet}
    \setlength{\tabcolsep}{2pt}
    \begin{tabular}{ccc|ccc|ccc}
        \toprule
        \multicolumn{3}{c|}{\begin{tabular}[c]{@{}c@{}}Loggerhead sea turtle (33)\end{tabular}} 
        &\multicolumn{3}{c|}{\begin{tabular}[c]{@{}c@{}}Sulphur-crested cockatoo (89)\end{tabular}} 
        &\multicolumn{3}{c}{\begin{tabular}[c]{@{}c@{}}Golden retriever (207)\end{tabular}} \\
        \includegraphics[width=0.1\textwidth]{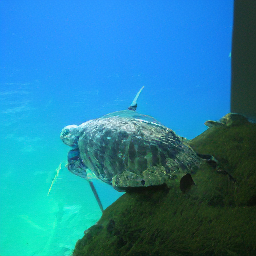} 
        & \includegraphics[width=0.1\textwidth]{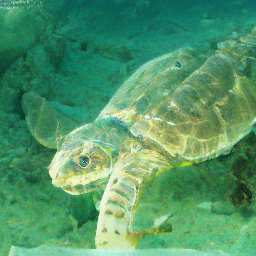} 
        &\includegraphics[width=0.1\textwidth]{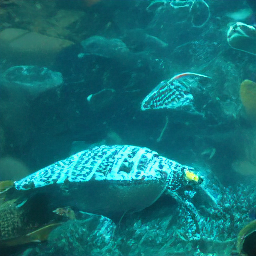}
        &\includegraphics[width=0.1\textwidth]{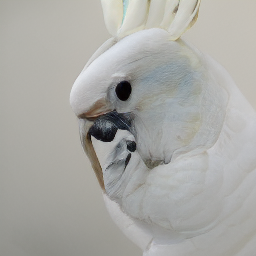} 
        & \includegraphics[width=0.1\textwidth]{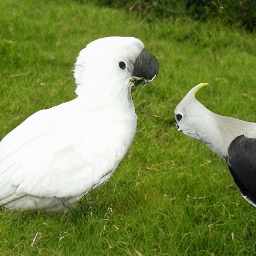} 
        &\includegraphics[width=0.1\textwidth]{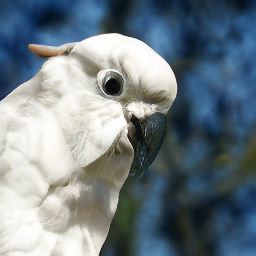}
         &\includegraphics[width=0.1\textwidth]{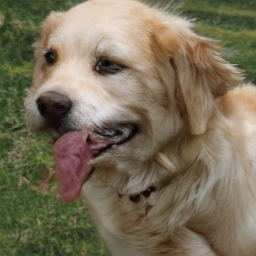} 
        & \includegraphics[width=0.1\textwidth]{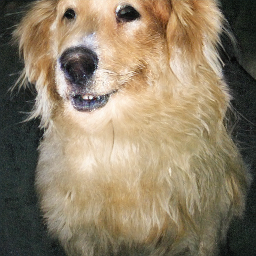} 
        &\includegraphics[width=0.1\textwidth]{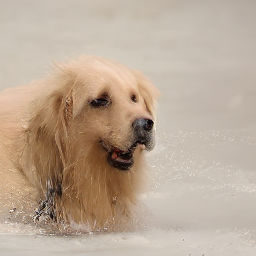}
        \\
        \midrule
        \multicolumn{3}{c|}{\begin{tabular}[c]{@{}c@{}}Husky (250)\end{tabular}} 
        &\multicolumn{3}{c|}{\begin{tabular}[c]{@{}c@{}}Panda (388)\end{tabular}} 
        &\multicolumn{3}{c}{\begin{tabular}[c]{@{}c@{}}Balloon (417)\end{tabular}} \\
         \includegraphics[width=0.1\textwidth]{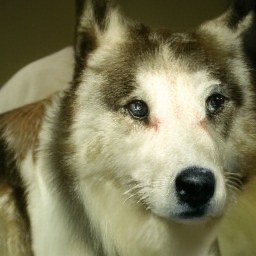} 
        & \includegraphics[width=0.1\textwidth]{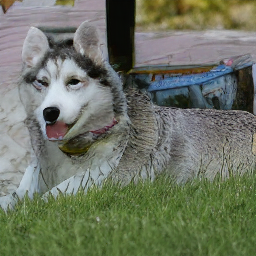} 
        &\includegraphics[width=0.1\textwidth]{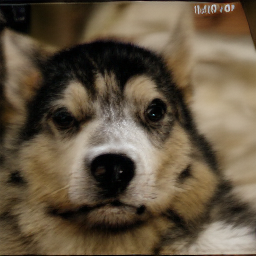}
        &\includegraphics[width=0.1\textwidth]{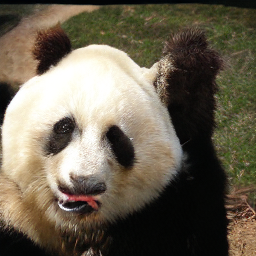} 
        & \includegraphics[width=0.1\textwidth]{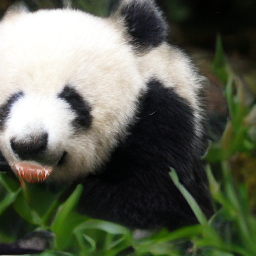} 
        &\includegraphics[width=0.1\textwidth]{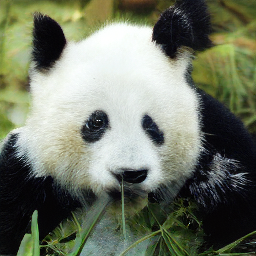}
        &\includegraphics[width=0.1\textwidth]{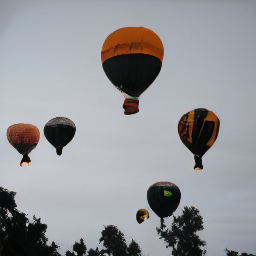} 
        & \includegraphics[width=0.1\textwidth]{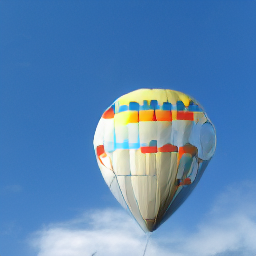} 
        &\includegraphics[width=0.1\textwidth]{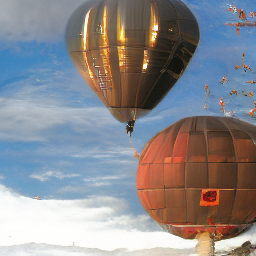}
        \\
        \midrule
        \multicolumn{3}{c|}{\begin{tabular}[c]{@{}c@{}}Baseball (429)\end{tabular}} 
        &\multicolumn{3}{c|}{\begin{tabular}[c]{@{}c@{}}Space shuttle (812)\end{tabular}} 
        &\multicolumn{3}{c}{\begin{tabular}[c]{@{}c@{}}Volcano (980)\end{tabular}} \\
        \includegraphics[width=0.1\textwidth]{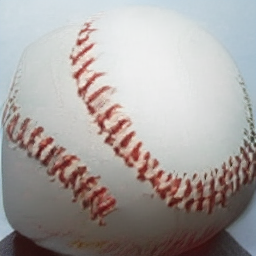} 
        & \includegraphics[width=0.1\textwidth]{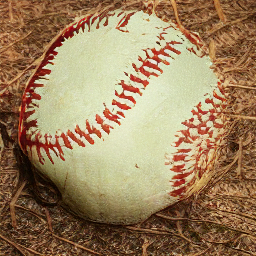} 
        &\includegraphics[width=0.1\textwidth]{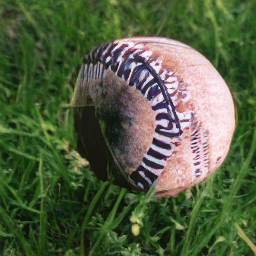}
        &\includegraphics[width=0.1\textwidth]{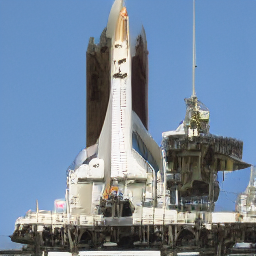} 
        & \includegraphics[width=0.1\textwidth]{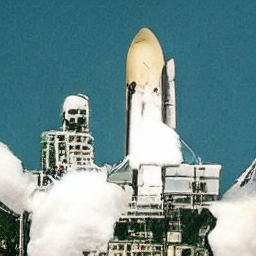} 
        &\includegraphics[width=0.1\textwidth]{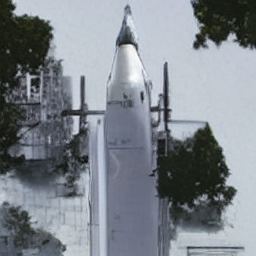}
        &\includegraphics[width=0.1\textwidth]{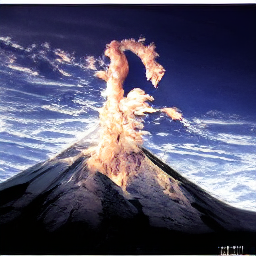} 
        & \includegraphics[width=0.1\textwidth]{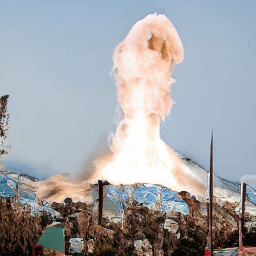} 
        &\includegraphics[width=0.1\textwidth]{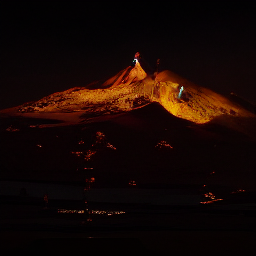}\\
        \bottomrule
    \end{tabular}
\end{table*}

\begin{table*}[ht]
    \centering
    \small
    \caption{Random class-conditional samples generated by SiT-XL/2 trained on the ImageNet dataset using the COAP optimizer.}
    \label{tbl:sit-imagenet}
    \setlength{\tabcolsep}{2pt}
    \begin{tabular}{ccc|ccc|ccc}
        \toprule
        \multicolumn{3}{c|}{\begin{tabular}[c]{@{}c@{}}Loggerhead sea turtle (33)\end{tabular}} 
        &\multicolumn{3}{c|}{\begin{tabular}[c]{@{}c@{}}Sulphur-crested cockatoo (89)\end{tabular}} 
        &\multicolumn{3}{c}{\begin{tabular}[c]{@{}c@{}}Golden retriever (207)\end{tabular}} \\
       \includegraphics[width=0.1\textwidth]{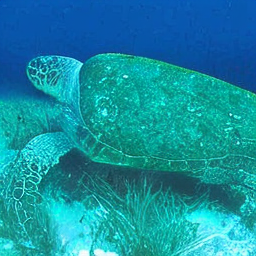} 
        & \includegraphics[width=0.1\textwidth]{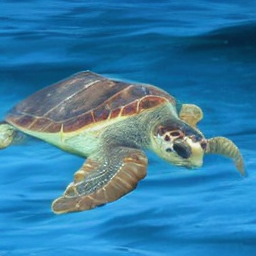} 
        &\includegraphics[width=0.1\textwidth]{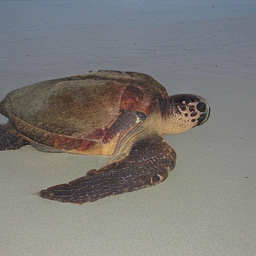}
        &\includegraphics[width=0.1\textwidth]{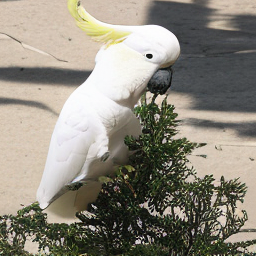} 
        & \includegraphics[width=0.1\textwidth]{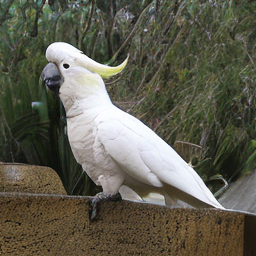} 
        &\includegraphics[width=0.1\textwidth]{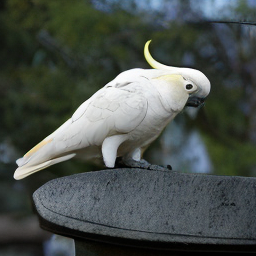}
        &\includegraphics[width=0.1\textwidth]{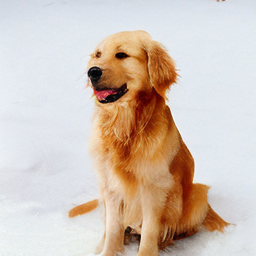} 
        & \includegraphics[width=0.1\textwidth]{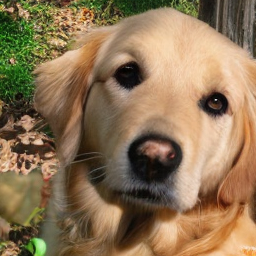} 
        &\includegraphics[width=0.1\textwidth]{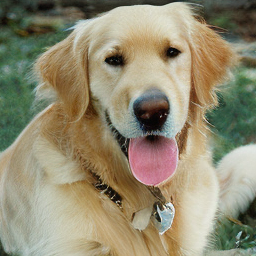}
        \\
        \midrule
        \multicolumn{3}{c|}{\begin{tabular}[c]{@{}c@{}}Husky (250)\end{tabular}} 
        &\multicolumn{3}{c|}{\begin{tabular}[c]{@{}c@{}}Panda (388)\end{tabular}} 
        &\multicolumn{3}{c}{\begin{tabular}[c]{@{}c@{}}Balloon (417)\end{tabular}} \\
        \includegraphics[width=0.1\textwidth]{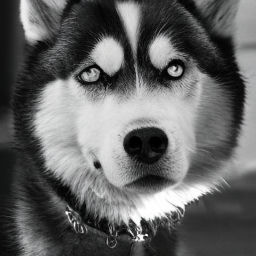} 
        & \includegraphics[width=0.1\textwidth]{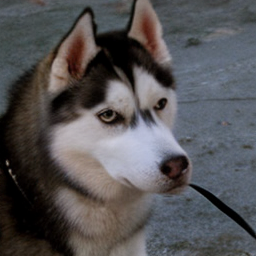} 
        &\includegraphics[width=0.1\textwidth]{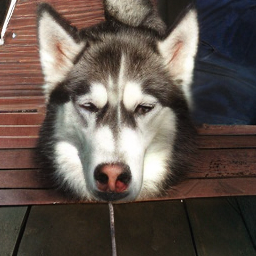}
        &\includegraphics[width=0.1\textwidth]{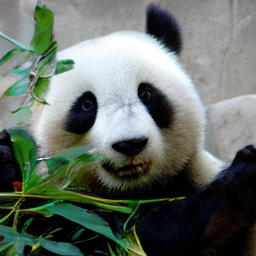} 
        & \includegraphics[width=0.1\textwidth]{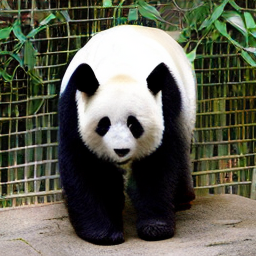} 
        &\includegraphics[width=0.1\textwidth]{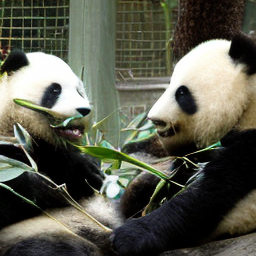}
        &\includegraphics[width=0.1\textwidth]{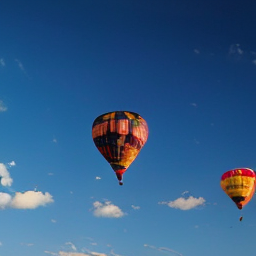} 
        & \includegraphics[width=0.1\textwidth]{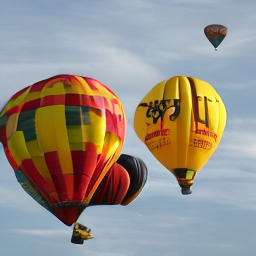} 
        &\includegraphics[width=0.1\textwidth]{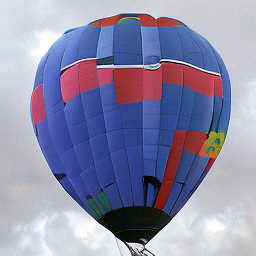}\\
         \midrule
        \multicolumn{3}{c|}{\begin{tabular}[c]{@{}c@{}}Baseball (429)\end{tabular}} 
        &\multicolumn{3}{c|}{\begin{tabular}[c]{@{}c@{}}Space shuttle (812)\end{tabular}} 
        &\multicolumn{3}{c}{\begin{tabular}[c]{@{}c@{}}Volcano (980)\end{tabular}} \\
        \includegraphics[width=0.1\textwidth]{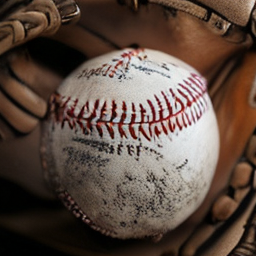} 
        & \includegraphics[width=0.1\textwidth]{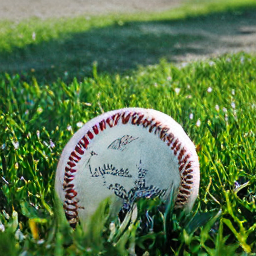} 
        &\includegraphics[width=0.1\textwidth]{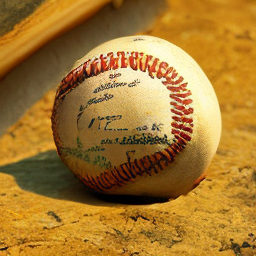}
        &\includegraphics[width=0.1\textwidth]{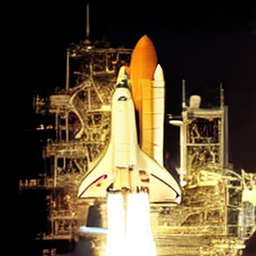} 
        & \includegraphics[width=0.1\textwidth]{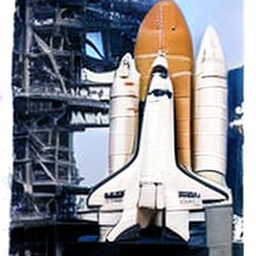} 
        &\includegraphics[width=0.1\textwidth]{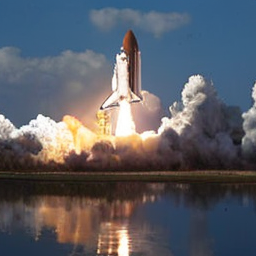}
        &\includegraphics[width=0.1\textwidth]{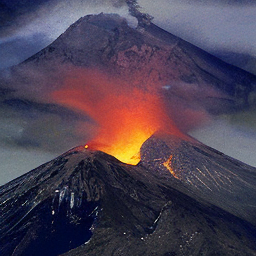} 
        & \includegraphics[width=0.1\textwidth]{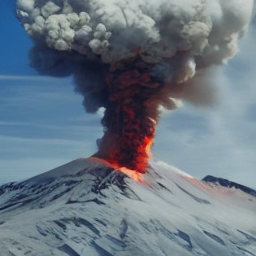} 
        &\includegraphics[width=0.1\textwidth]{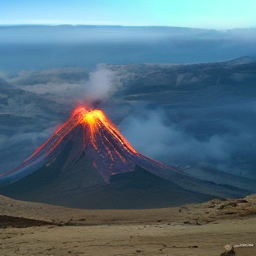}
        \\
        \bottomrule
    \end{tabular}
\end{table*}

\begin{table*}[ht]
    \centering
    \small
    \setlength{\tabcolsep}{2pt}
    \caption{Comparison of images generated at different training steps (20K, 40K, 80K) with ControlNet-XL trained under various optimizers. DDIM with a guidance scale of 5.0 is applied for image generation, with the number of inference steps set to 50.}
    \label{tbl:controlnet-xl-pose}
    \begin{tabular}{ccc|ccc}
        \toprule
        \multicolumn{6}{c}{Prompt: a young woman with a yellow flower crown on her head} \\
        \midrule
        \multirow{2}{*}{\begin{tabular}[c]{@{}c@{}}Human\\ pose\end{tabular}} & \multirow{2}{*}{\begin{tabular}[c]{@{}c@{}}Reference\end{tabular}} && 20K & 40K & 80K\\
        \cmidrule{4-6}
      &&&
        \multicolumn{3}{c}{\begin{tabular}[c]{@{}c@{}}Adafactor (5.1 GB)\end{tabular}} \\
         \includegraphics[width=0.15\textwidth]{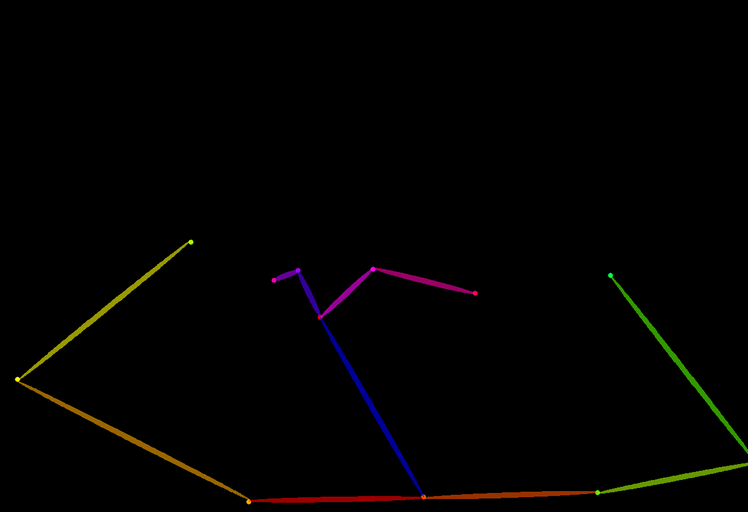} 
        & \includegraphics[width=0.15\textwidth]{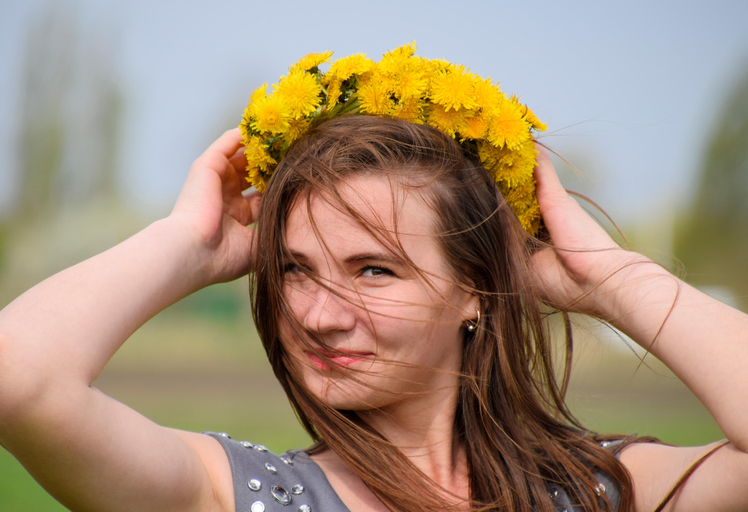} &
        &\includegraphics[width=0.15\textwidth]{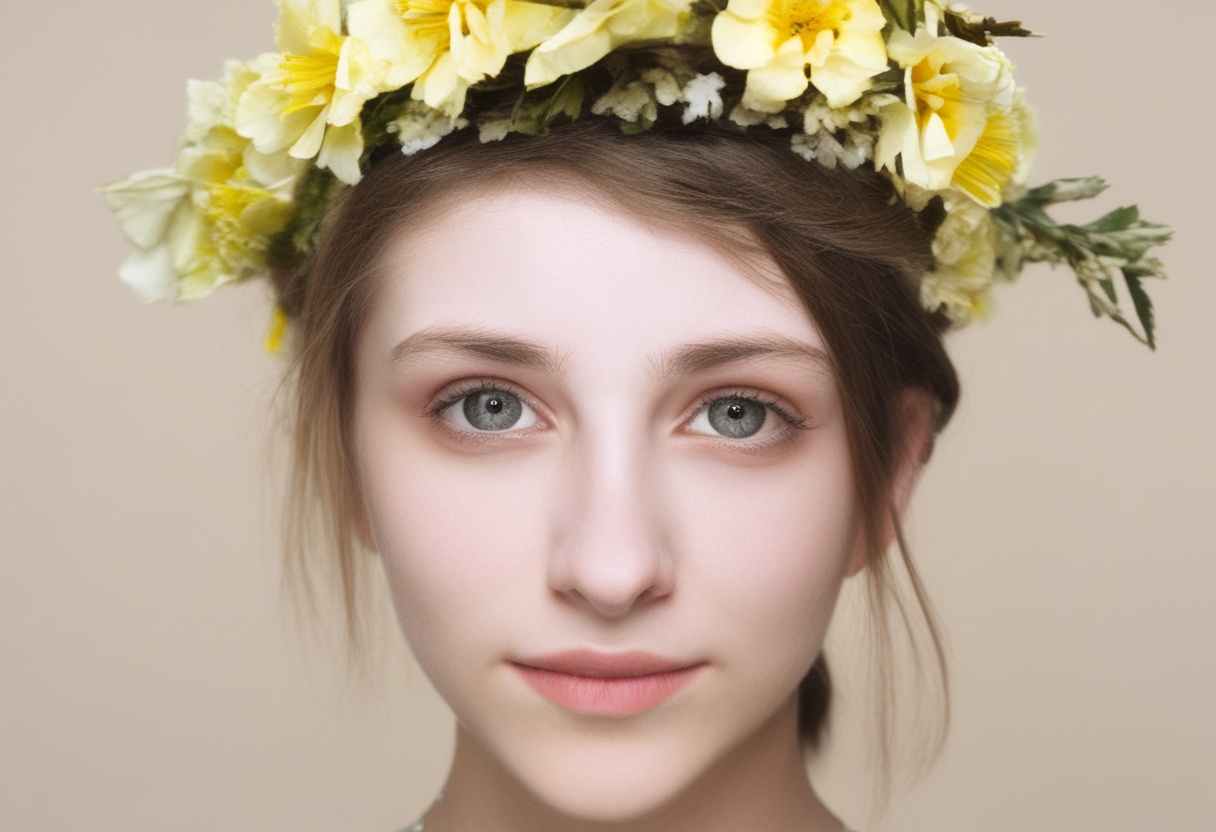} 
        & \includegraphics[width=0.15\textwidth]{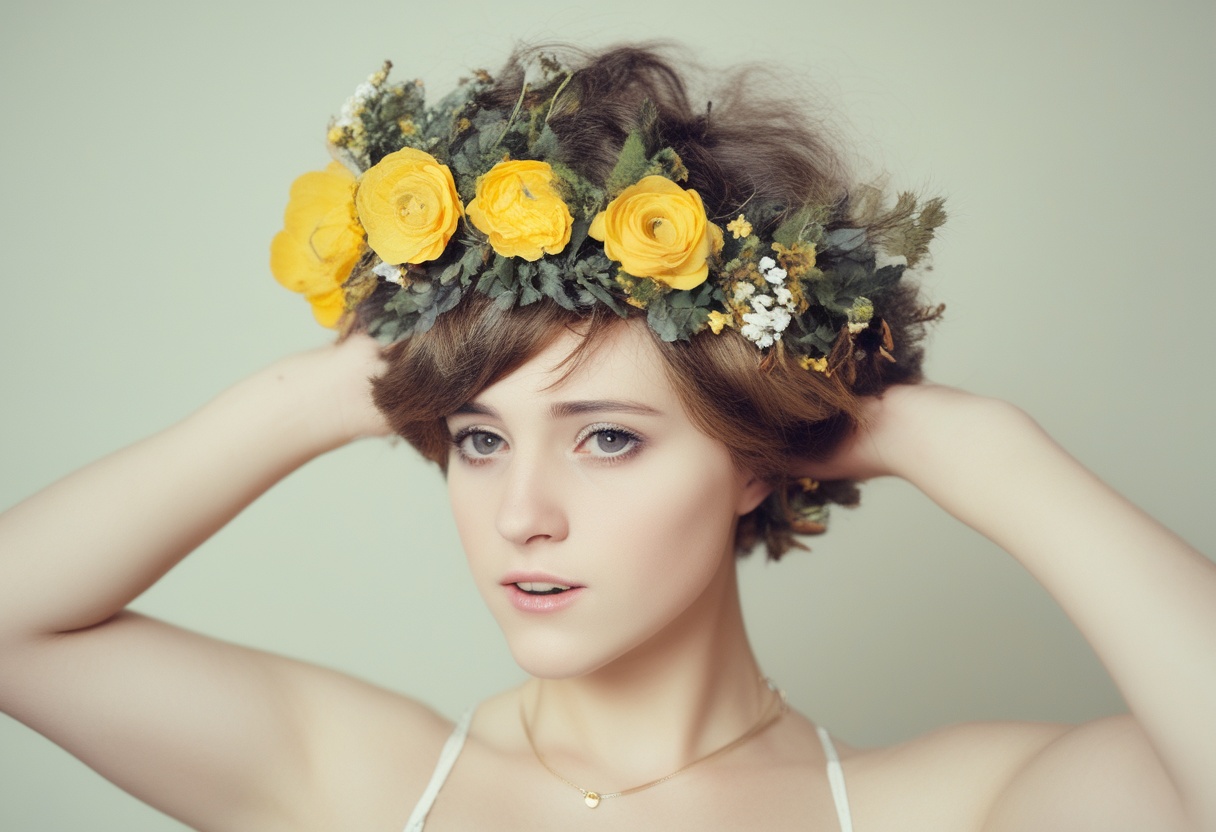} 
        & \includegraphics[width=0.15\textwidth]{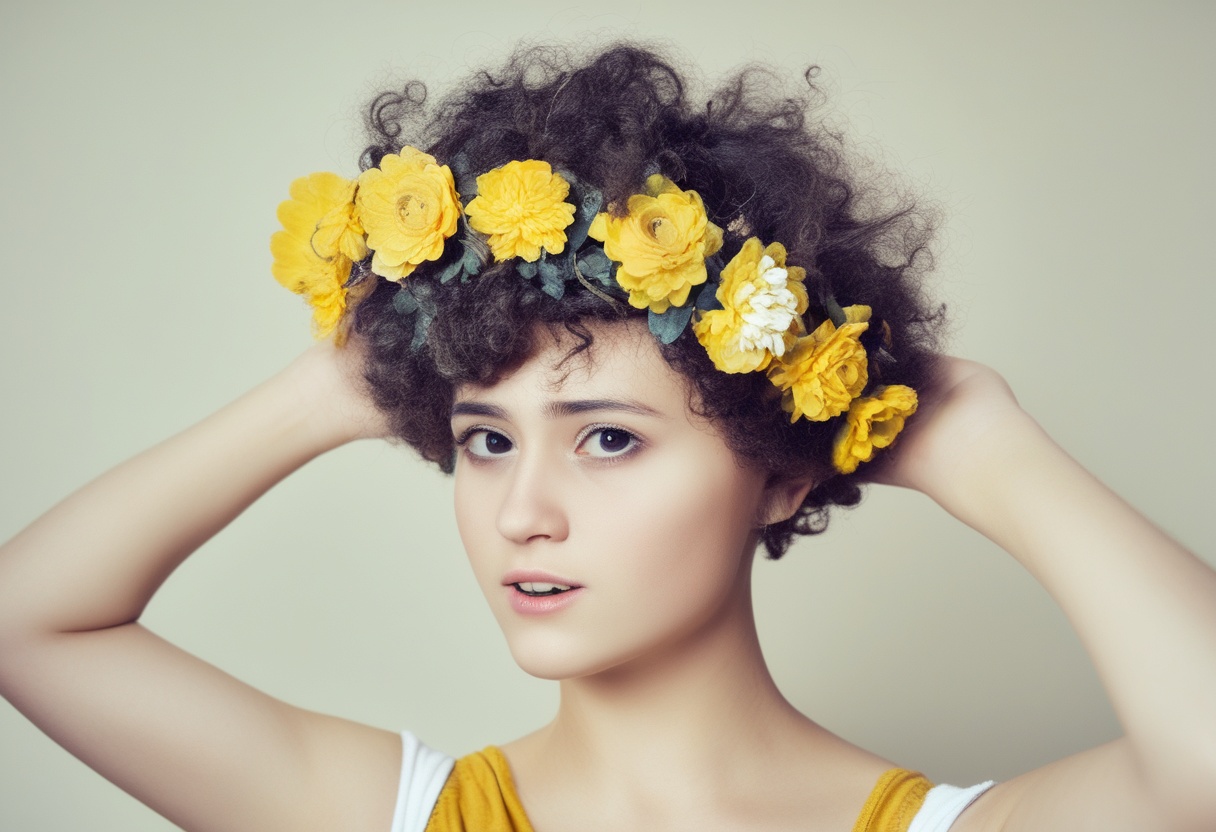}\\
        \midrule
        \multicolumn{3}{c|}{\begin{tabular}[c]{@{}c@{}}GaLore (4.7 GB)\end{tabular}} &\multicolumn{3}{c}{\begin{tabular}[c]{@{}c@{}}\textbf{COAP (3.6 GB)}\end{tabular}} \\
        \includegraphics[width=0.15\textwidth]{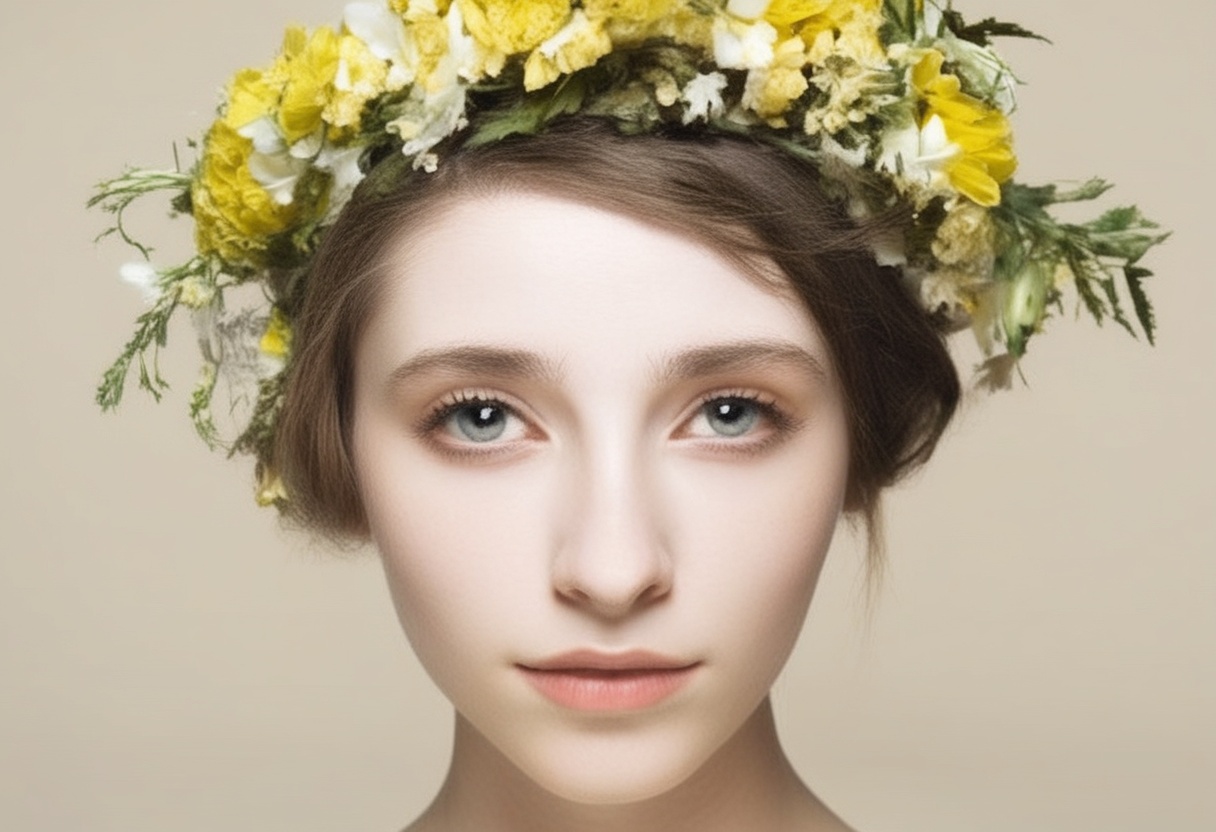} 
        & \includegraphics[width=0.15\textwidth]{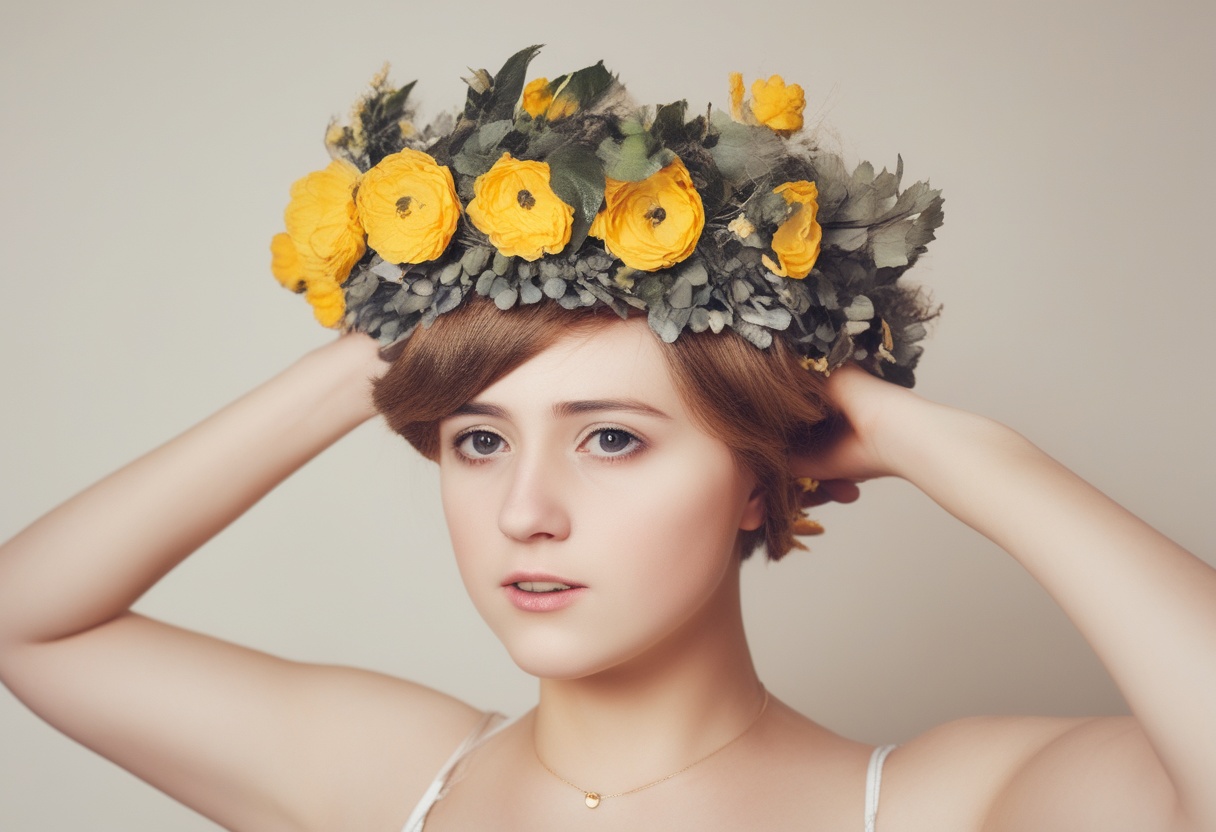} 
        & \includegraphics[width=0.15\textwidth]{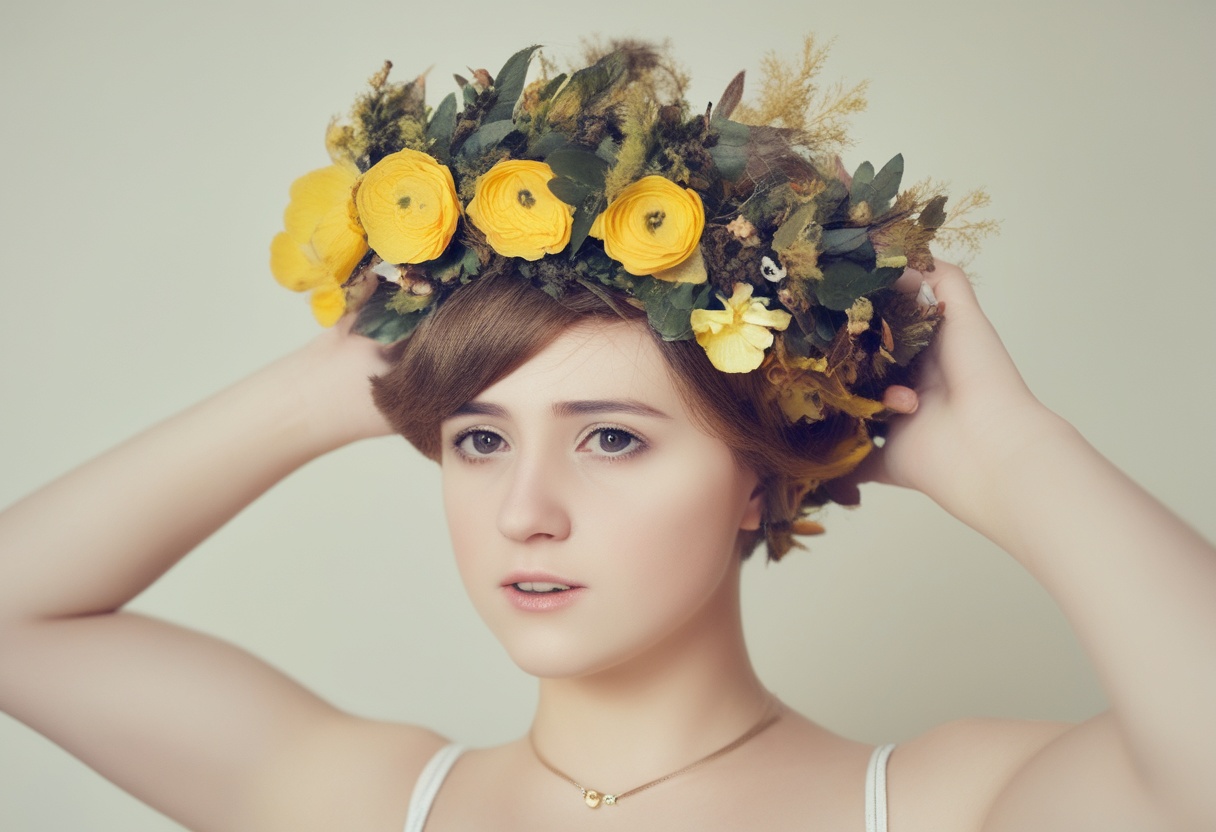}
        &\includegraphics[width=0.15\textwidth]{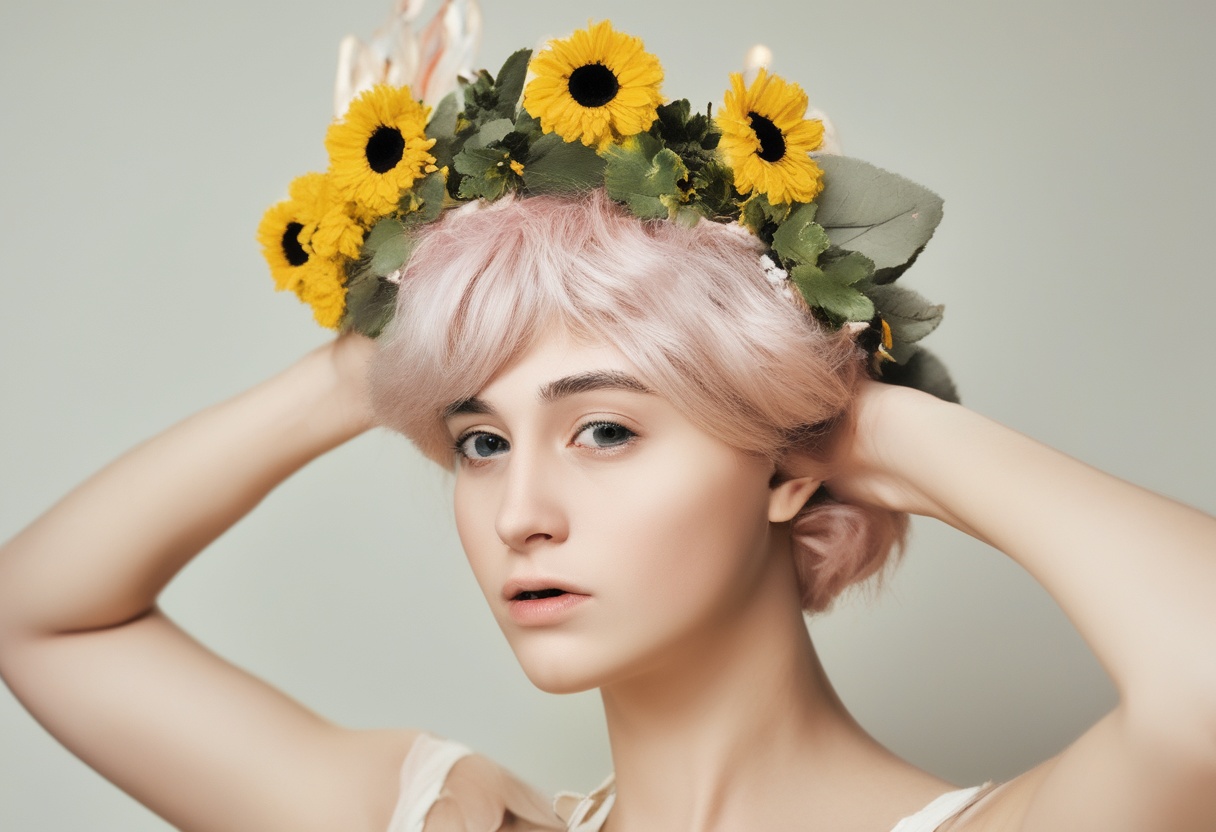} 
        & \includegraphics[width=0.15\textwidth]{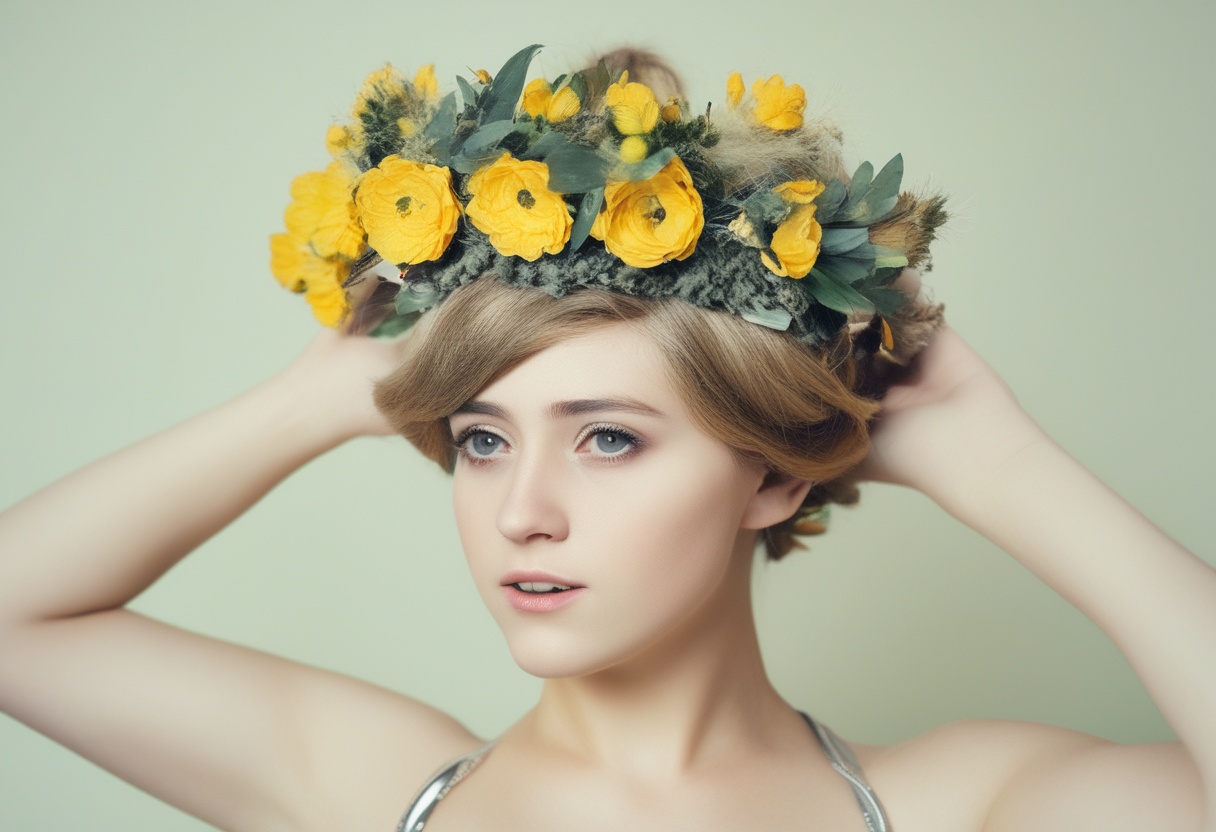} 
        & \includegraphics[width=0.15\textwidth]{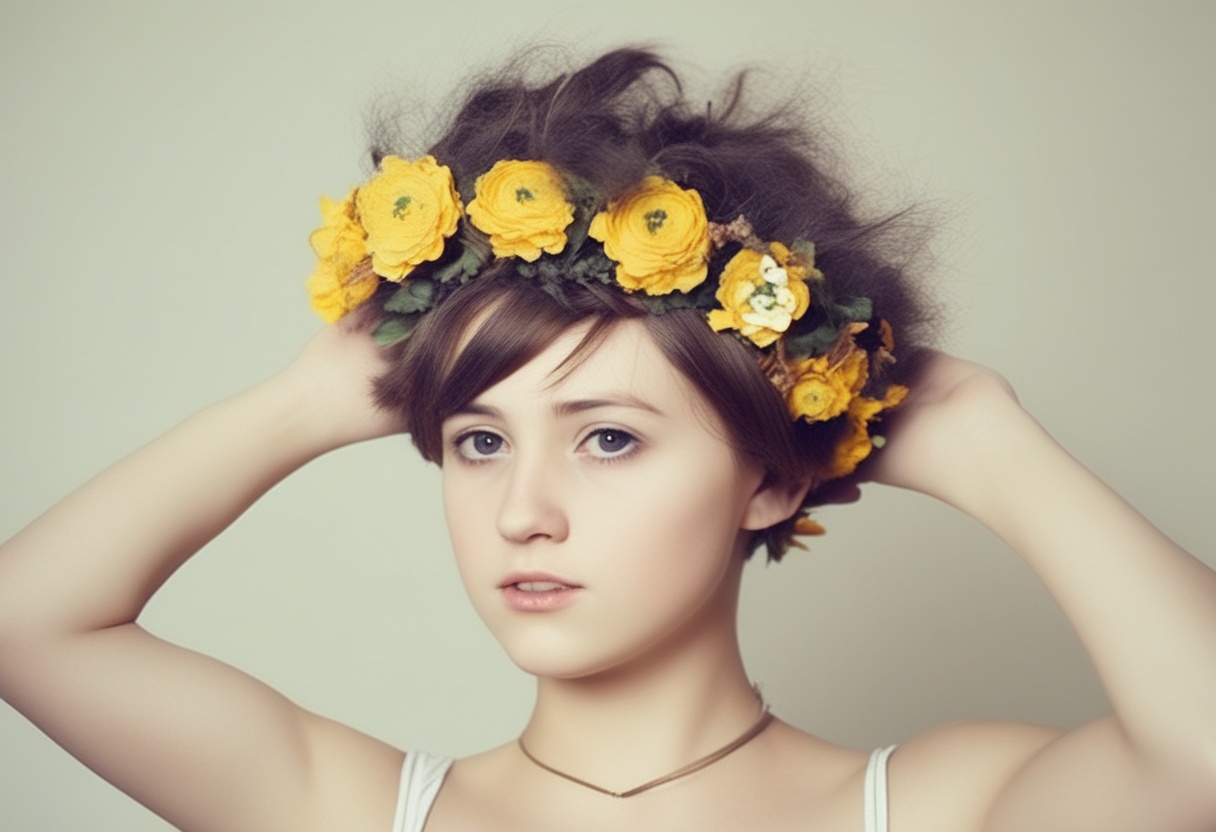}
 \\
 \midrule
 \multicolumn{3}{c|}{\begin{tabular}[c]{@{}c@{}} 8-bit GaLore (2.4GB)\end{tabular}} &\multicolumn{3}{c}{\begin{tabular}[c]{@{}c@{}}\textbf{8-bit COAP (0.5 GB)}\end{tabular}} \\
        \includegraphics[width=0.15\textwidth]{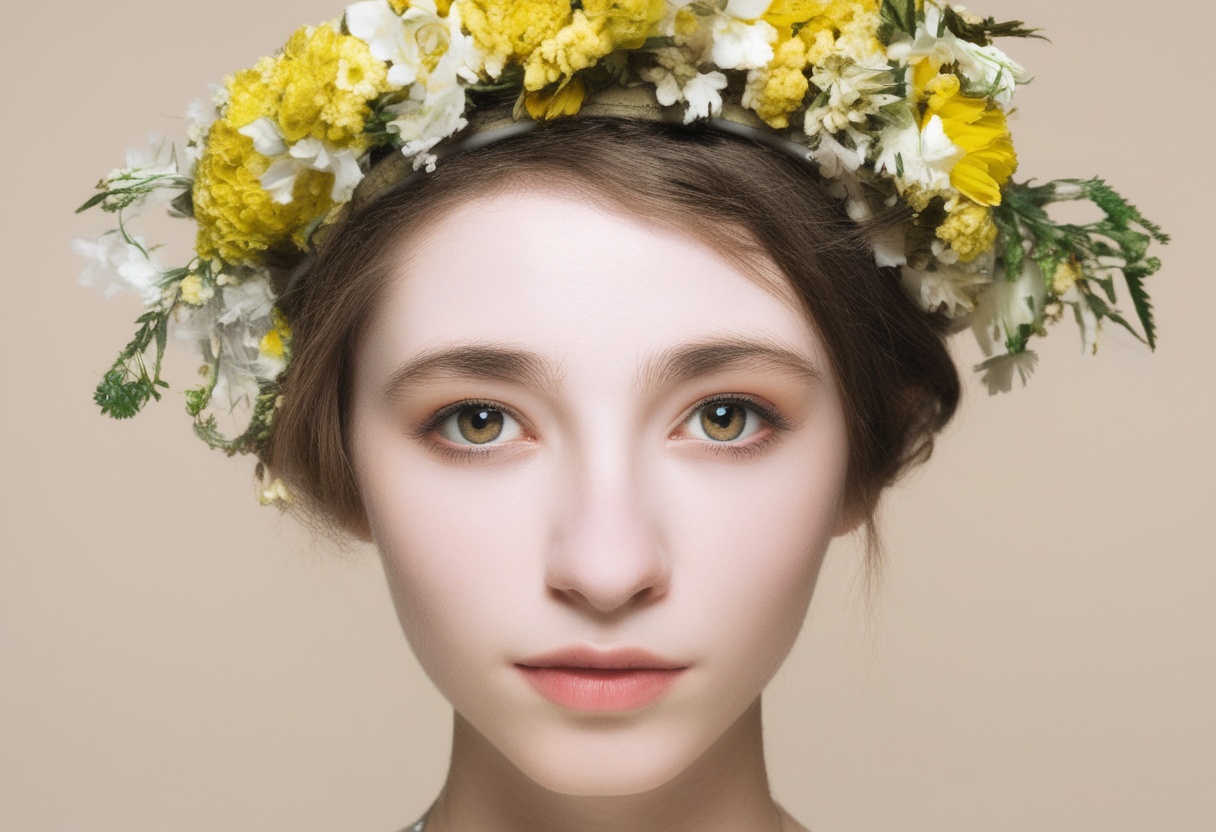} 
        & \includegraphics[width=0.15\textwidth]{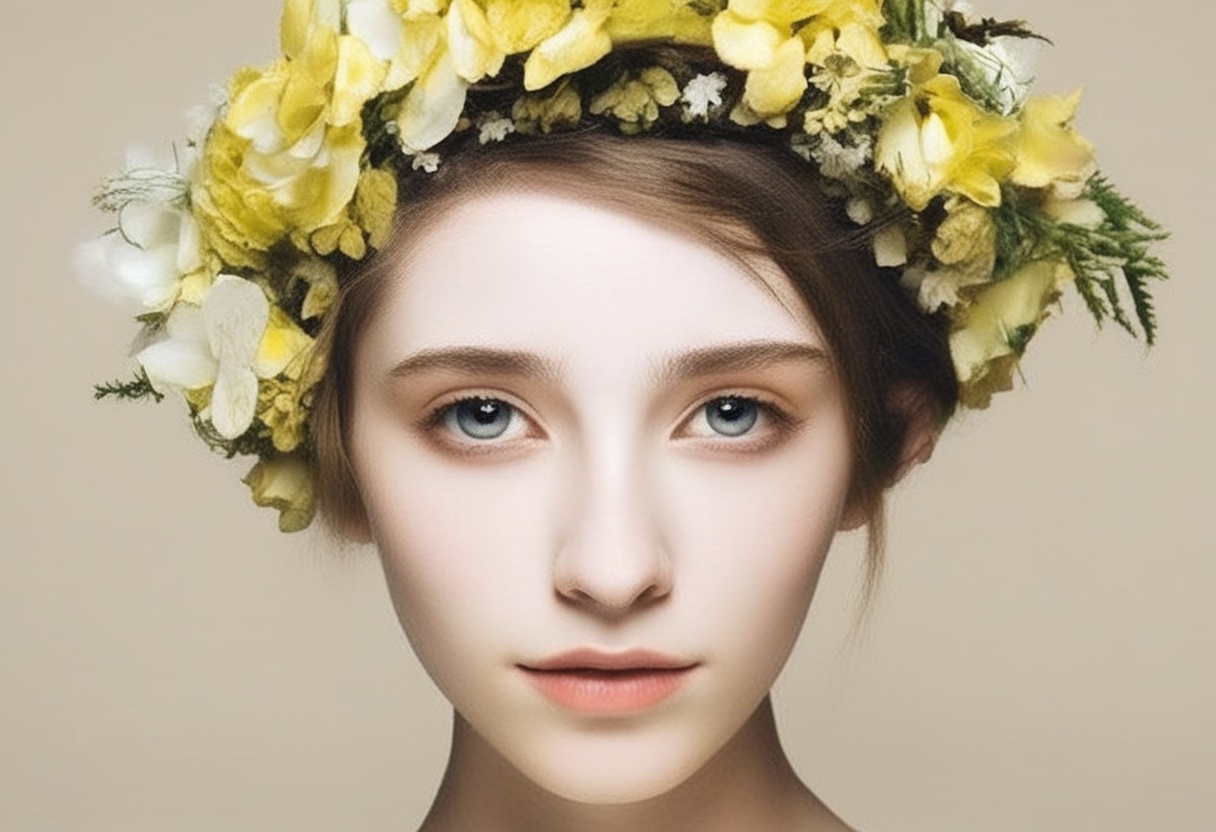} 
        & \includegraphics[width=0.15\textwidth]{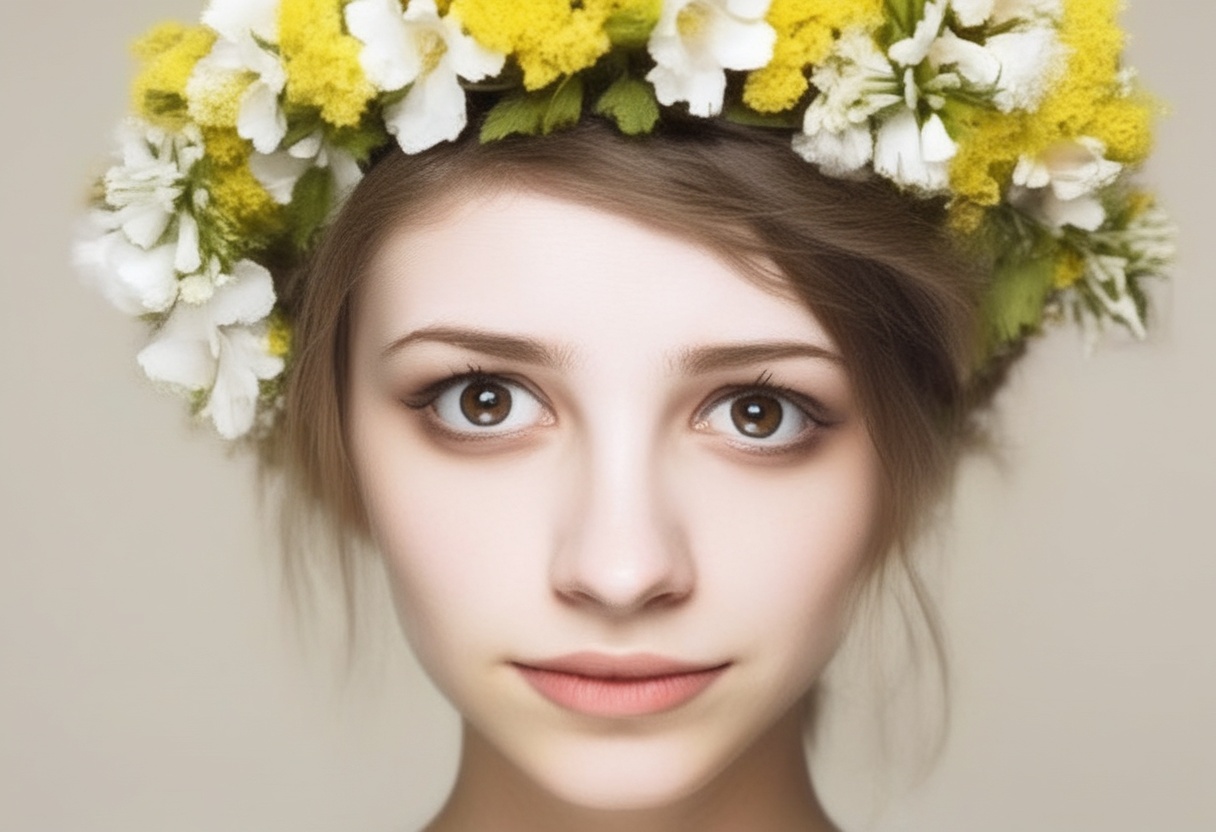}
        &\includegraphics[width=0.15\textwidth]{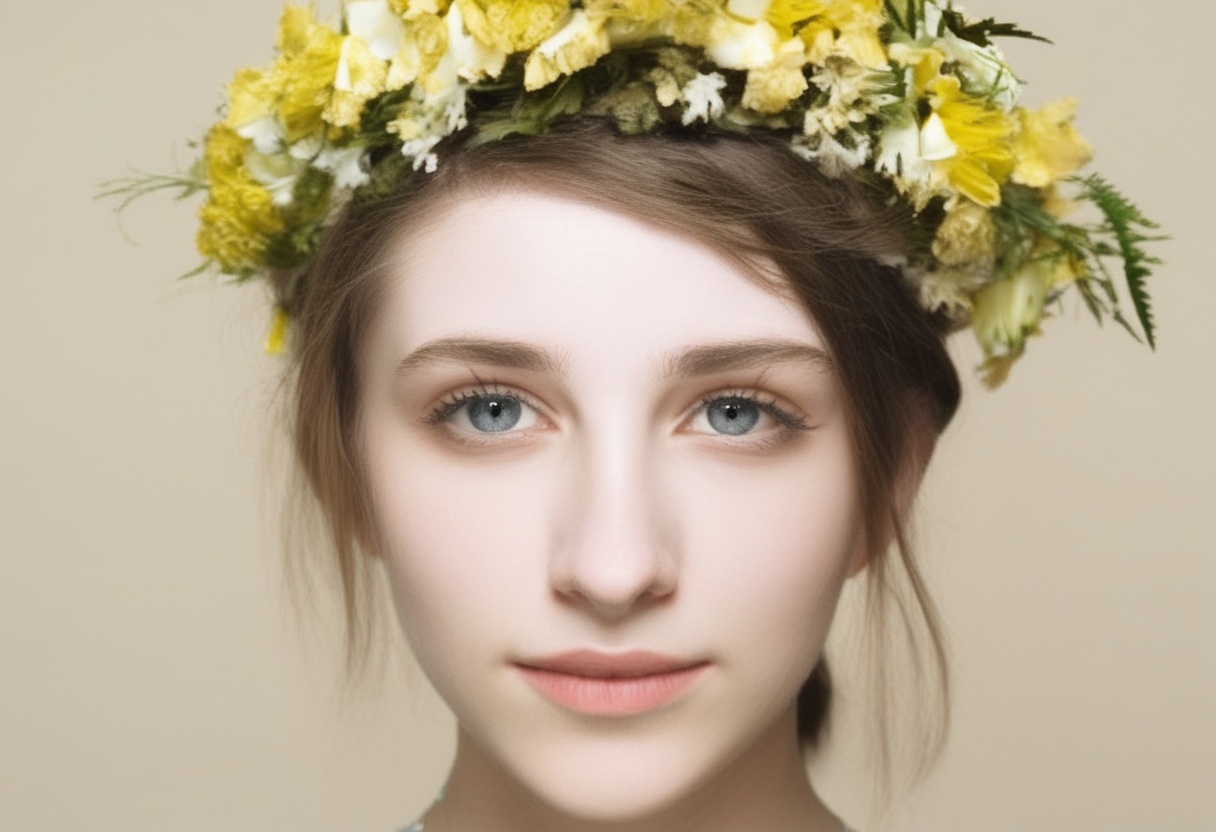} 
        & \includegraphics[width=0.15\textwidth]{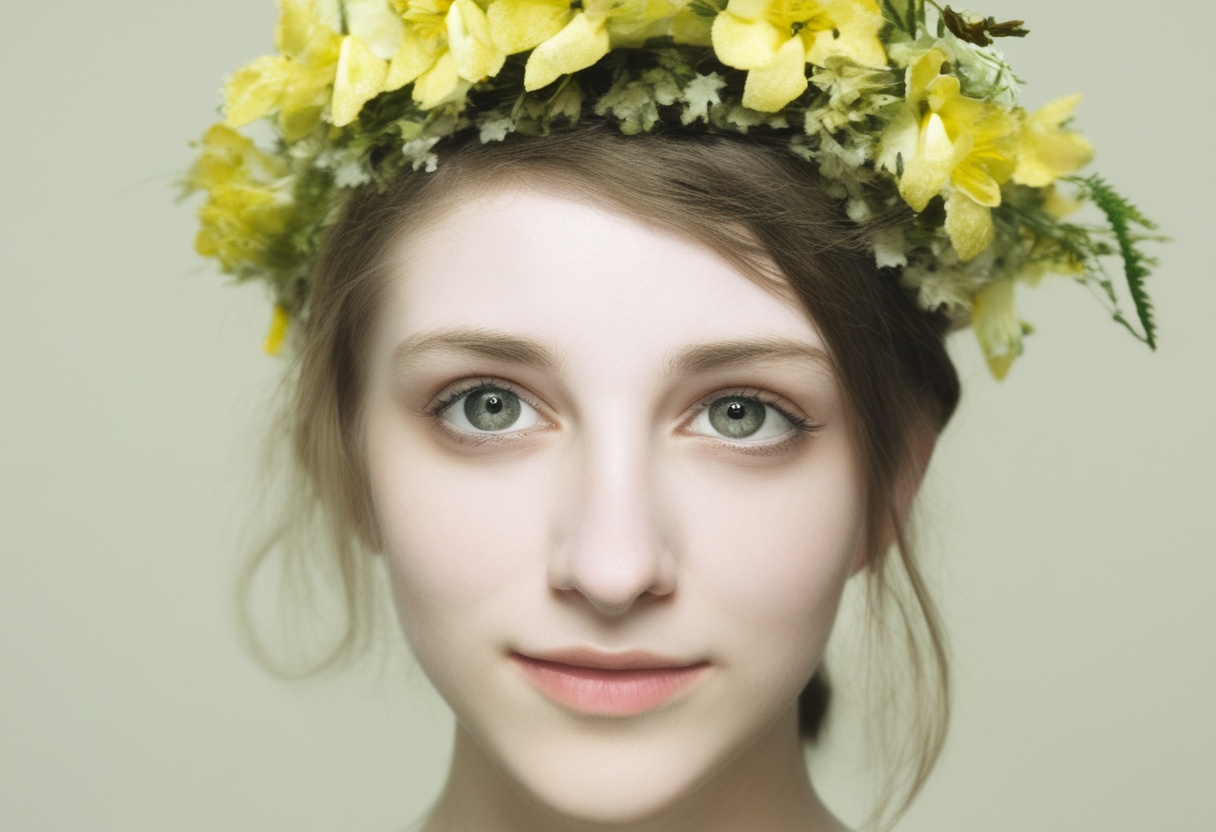} 
        & \includegraphics[width=0.15\textwidth]{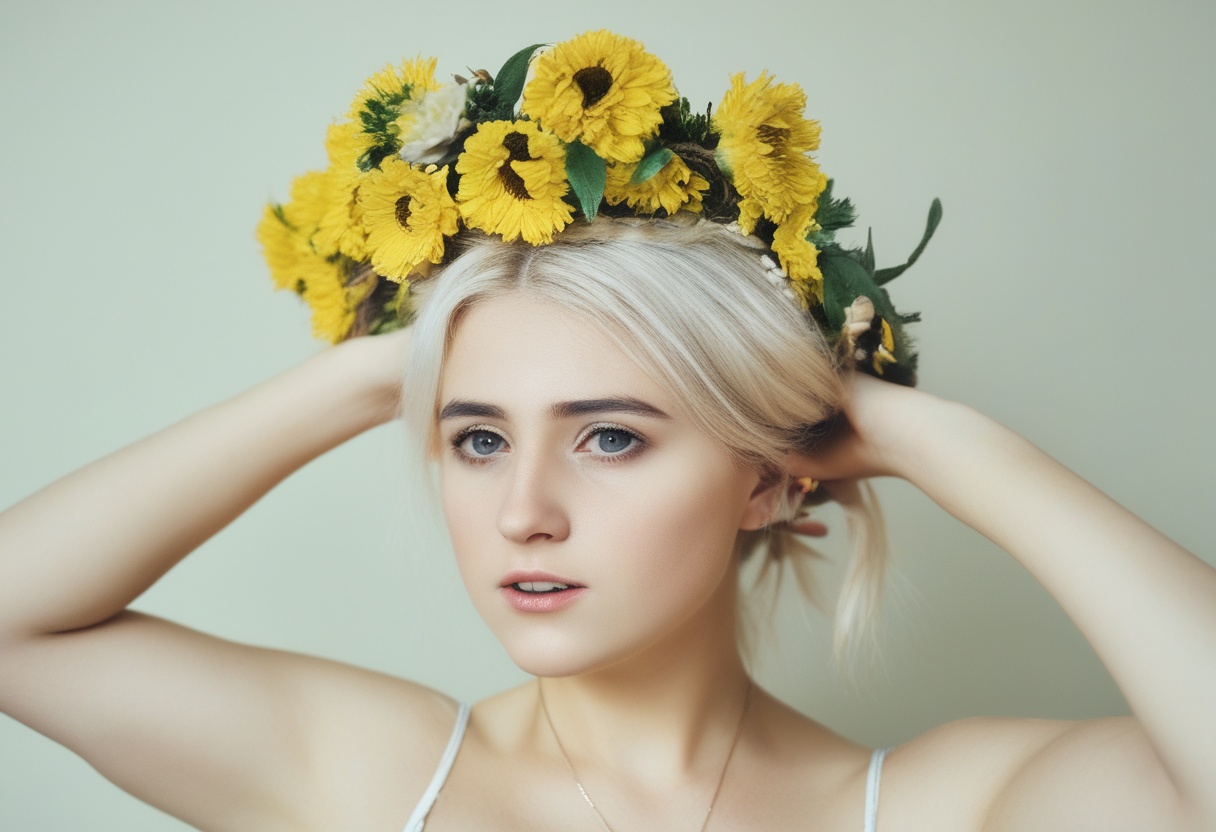} \\
        \bottomrule
    \end{tabular}
\end{table*}

\begin{table*}[ht]
    \centering
    \small
    \setlength{\tabcolsep}{2pt}
    \begin{tabular}{ccc|ccc}
        \toprule
        \multicolumn{6}{c}{Prompt: beautiful african american woman with curly hair on blue background} \\
        \midrule
        \multirow{2}{*}{\begin{tabular}[c]{@{}c@{}}Human\\ pose\end{tabular}} & \multirow{2}{*}{\begin{tabular}[c]{@{}c@{}}Reference\end{tabular}} && 20K & 40K & 80K\\
        \cmidrule{4-6}
      &&&
        \multicolumn{3}{c}{\begin{tabular}[c]{@{}c@{}}Adafactor (5.1 GB)\end{tabular}} \\
         \includegraphics[width=0.15\textwidth]{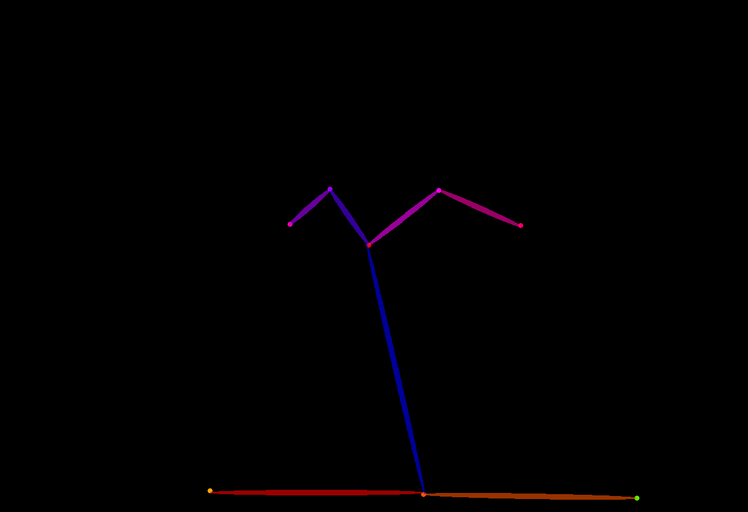} 
        & \includegraphics[width=0.15\textwidth]{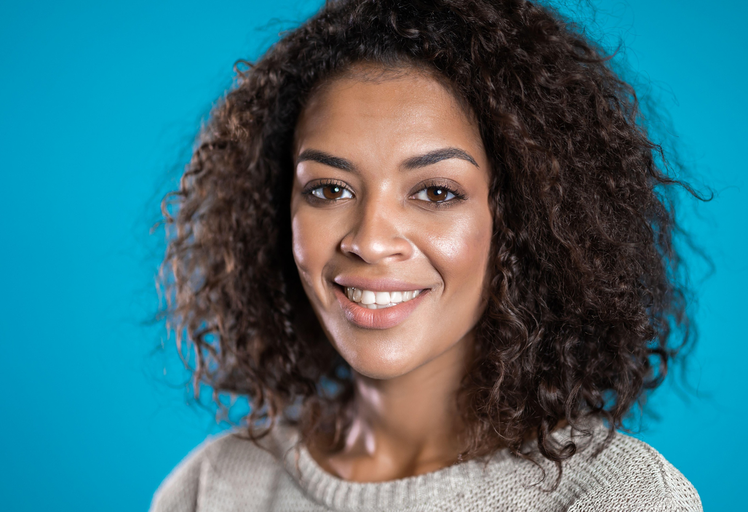} &
        &\includegraphics[width=0.15\textwidth]{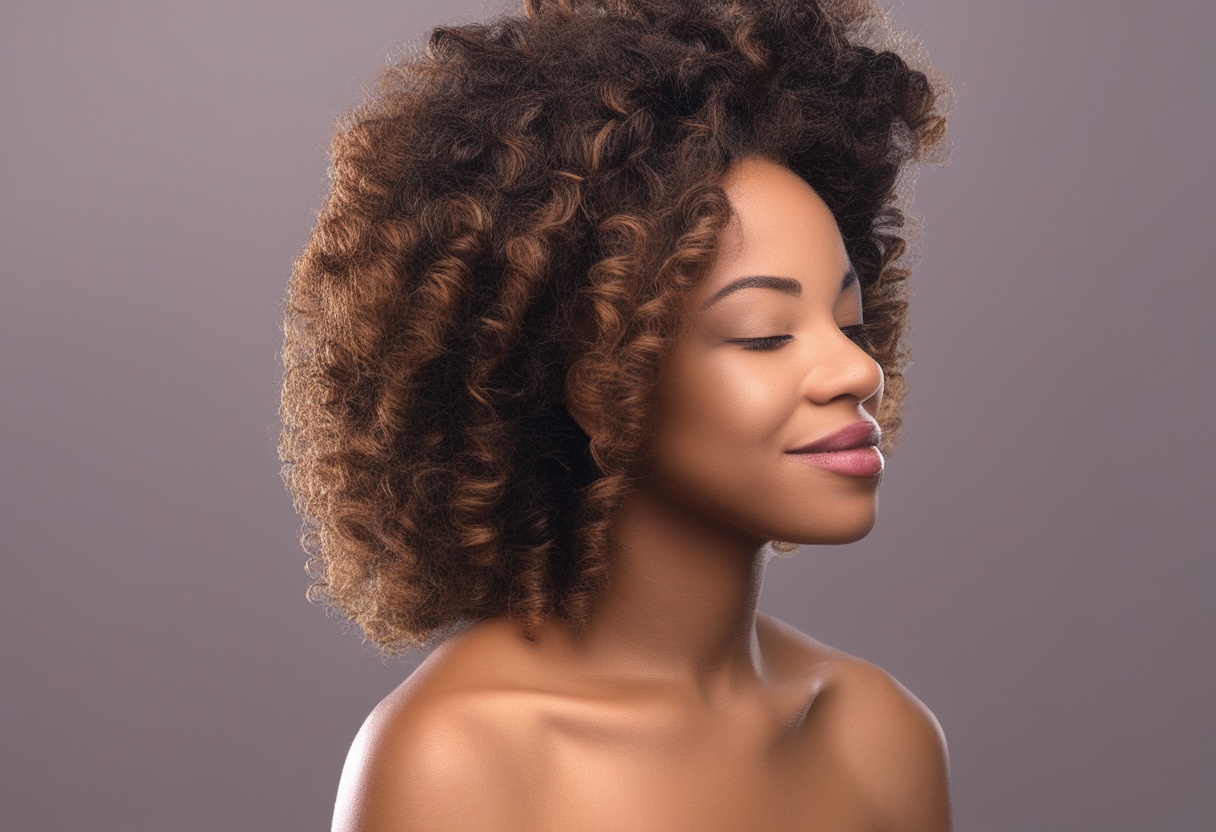} 
        & \includegraphics[width=0.15\textwidth]{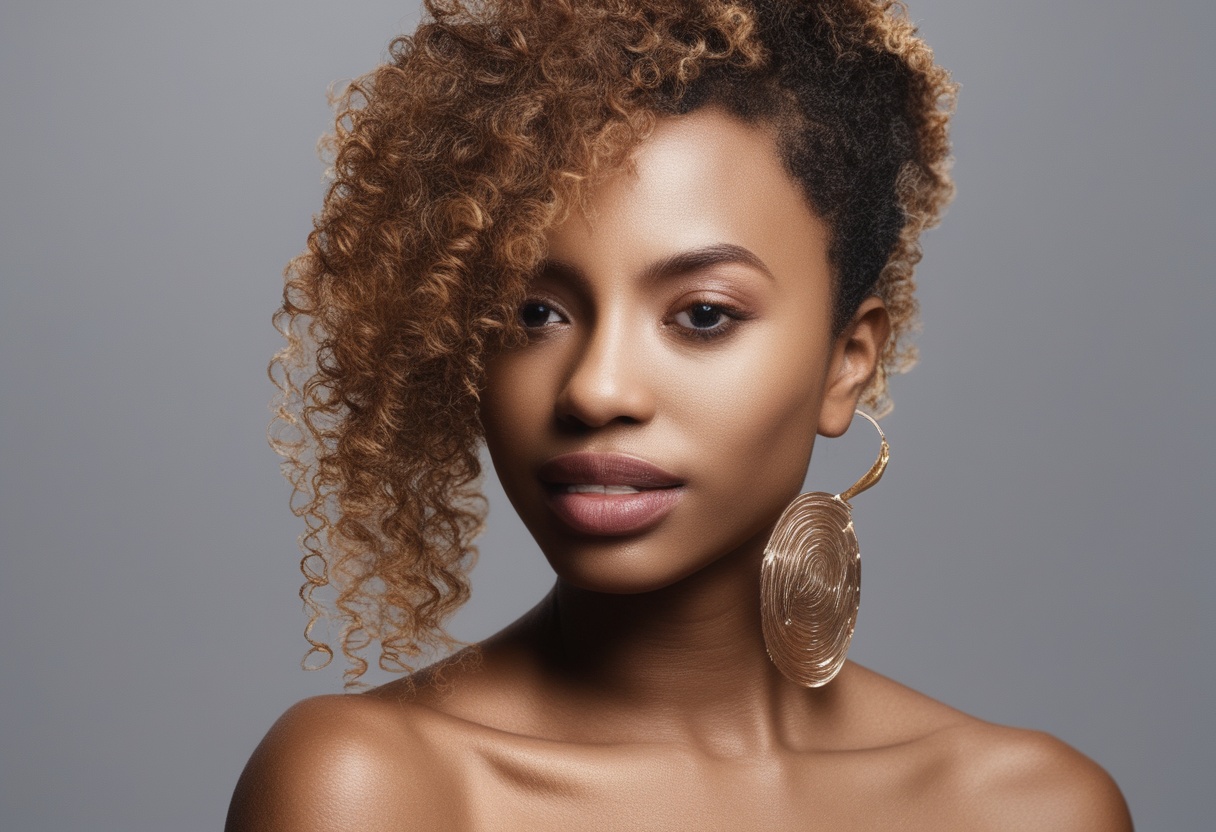} 
        & \includegraphics[width=0.15\textwidth]{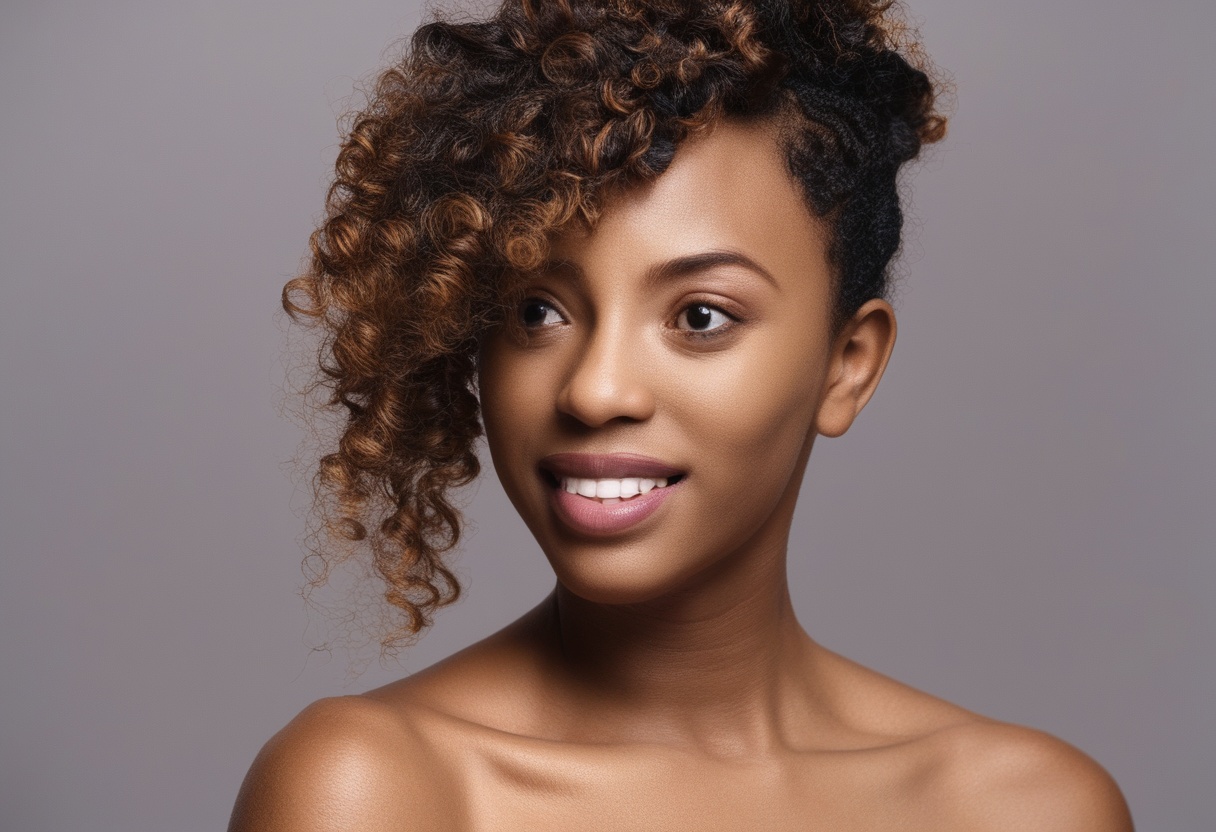}\\
        \midrule
        \multicolumn{3}{c|}{\begin{tabular}[c]{@{}c@{}}GaLore (4.7 GB)\end{tabular}} &\multicolumn{3}{c}{\begin{tabular}[c]{@{}c@{}}\textbf{COAP (3.6 GB)}\end{tabular}} \\
        \includegraphics[width=0.15\textwidth]{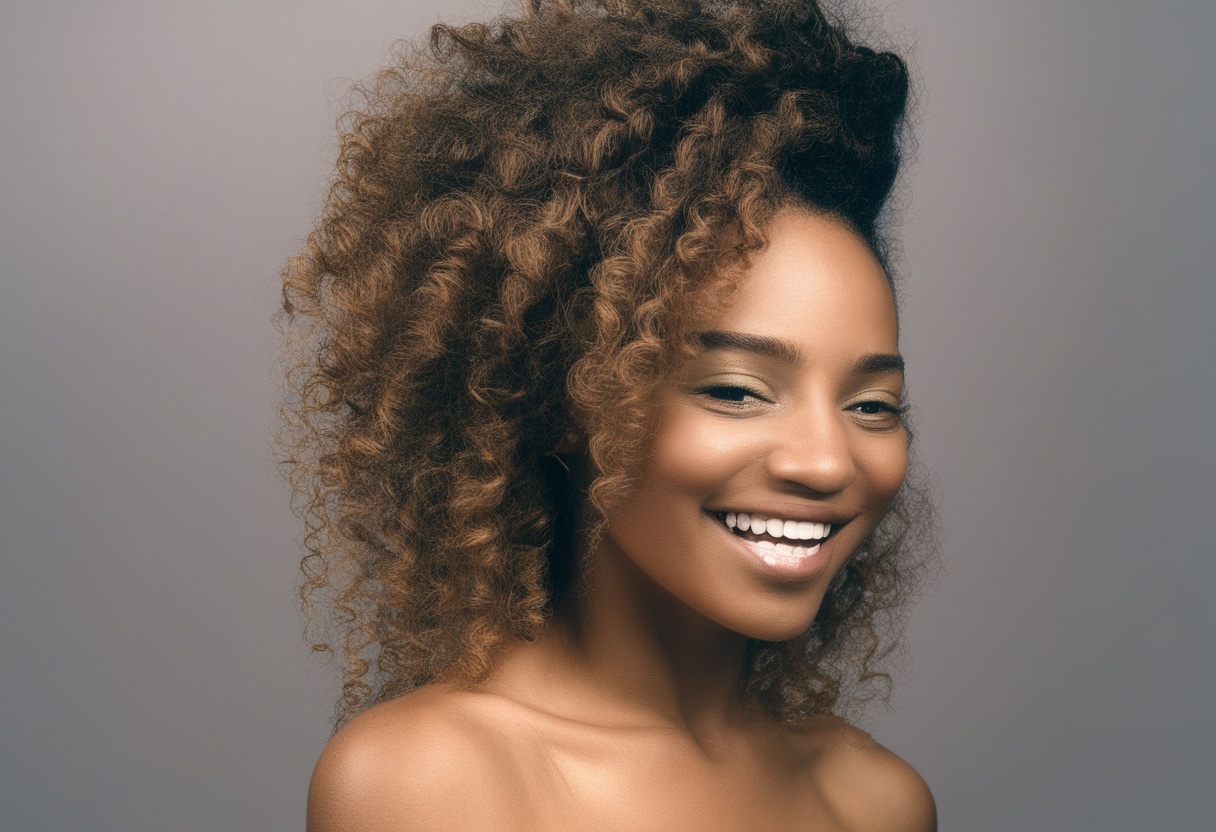} 
        & \includegraphics[width=0.15\textwidth]{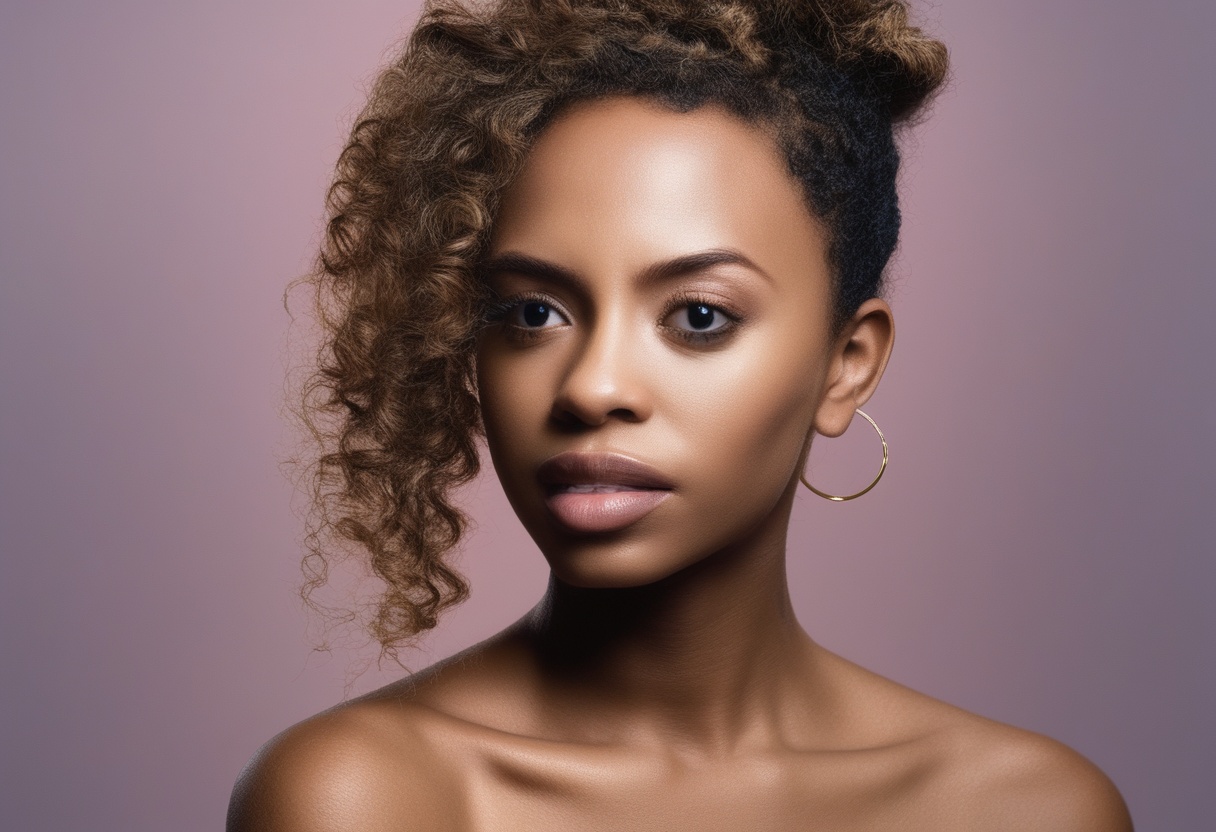} 
        & \includegraphics[width=0.15\textwidth]{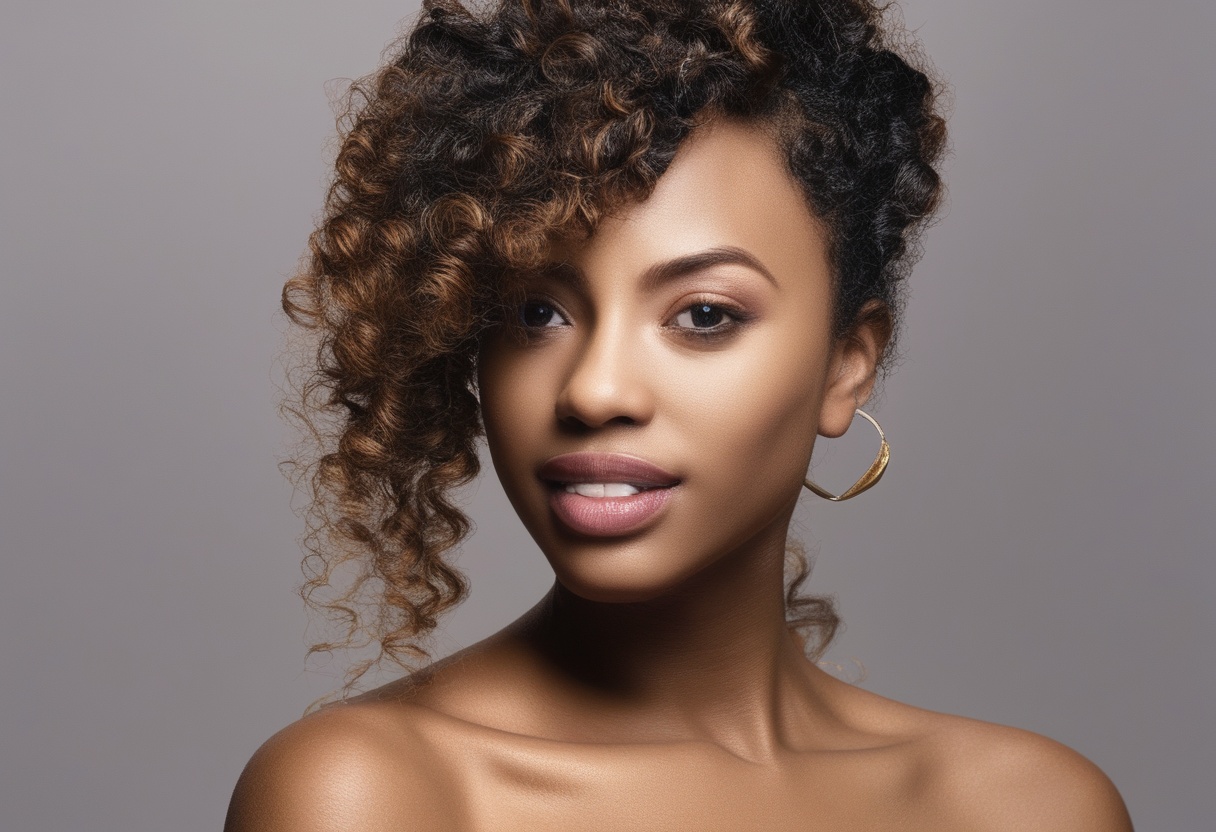}
        &\includegraphics[width=0.15\textwidth]{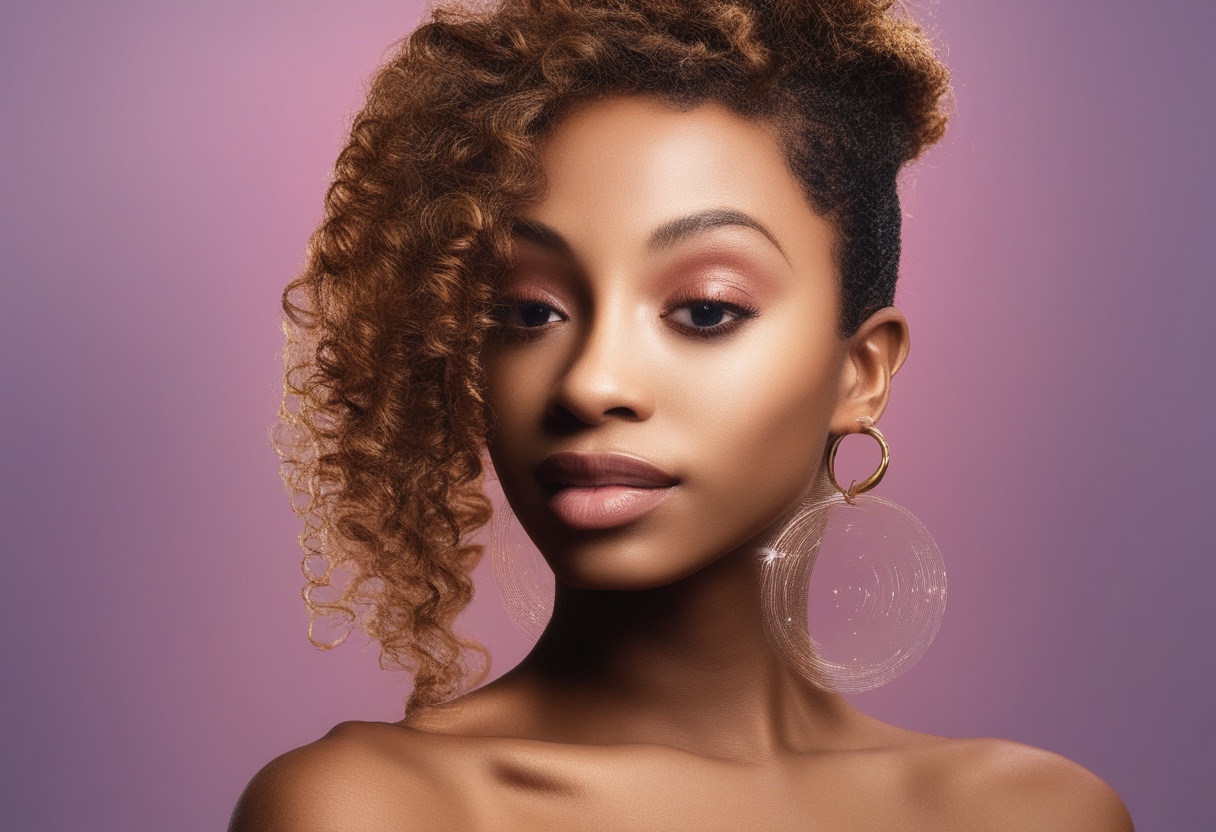} 
        & \includegraphics[width=0.15\textwidth]{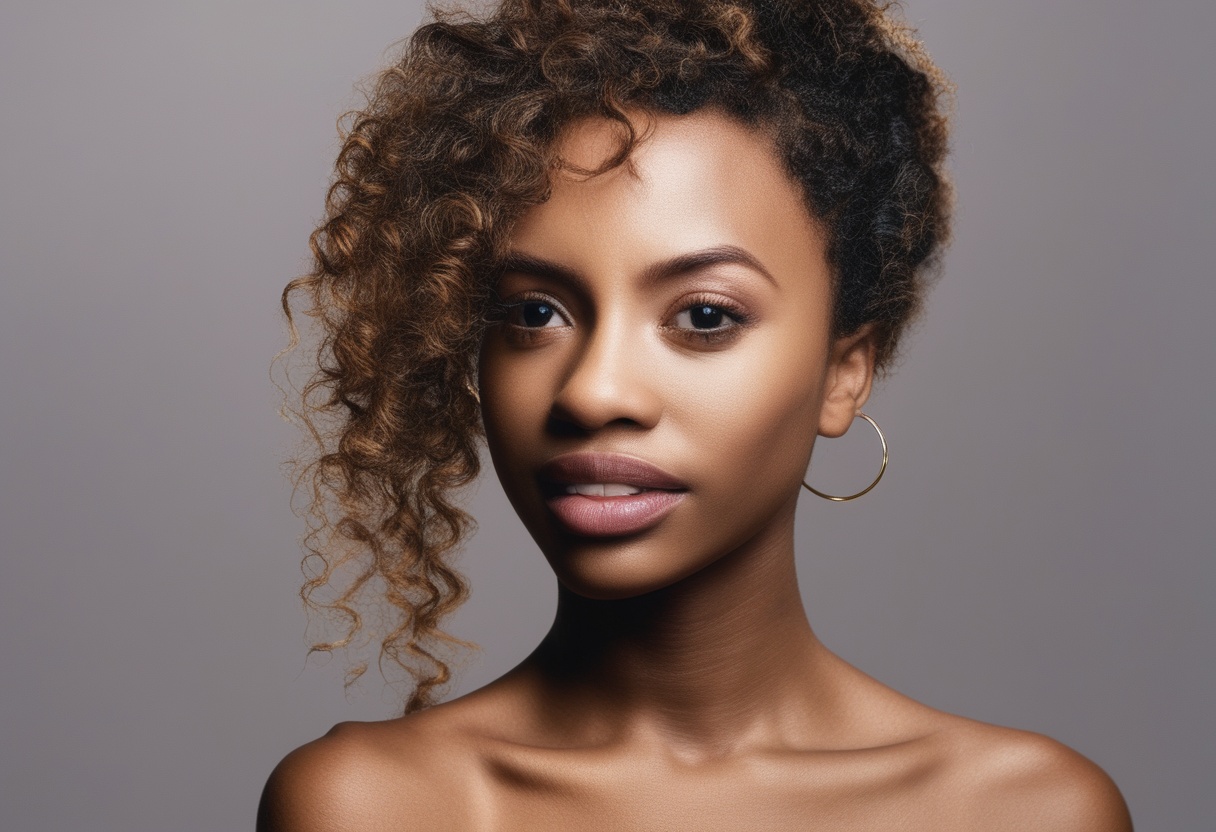} 
        & \includegraphics[width=0.15\textwidth]{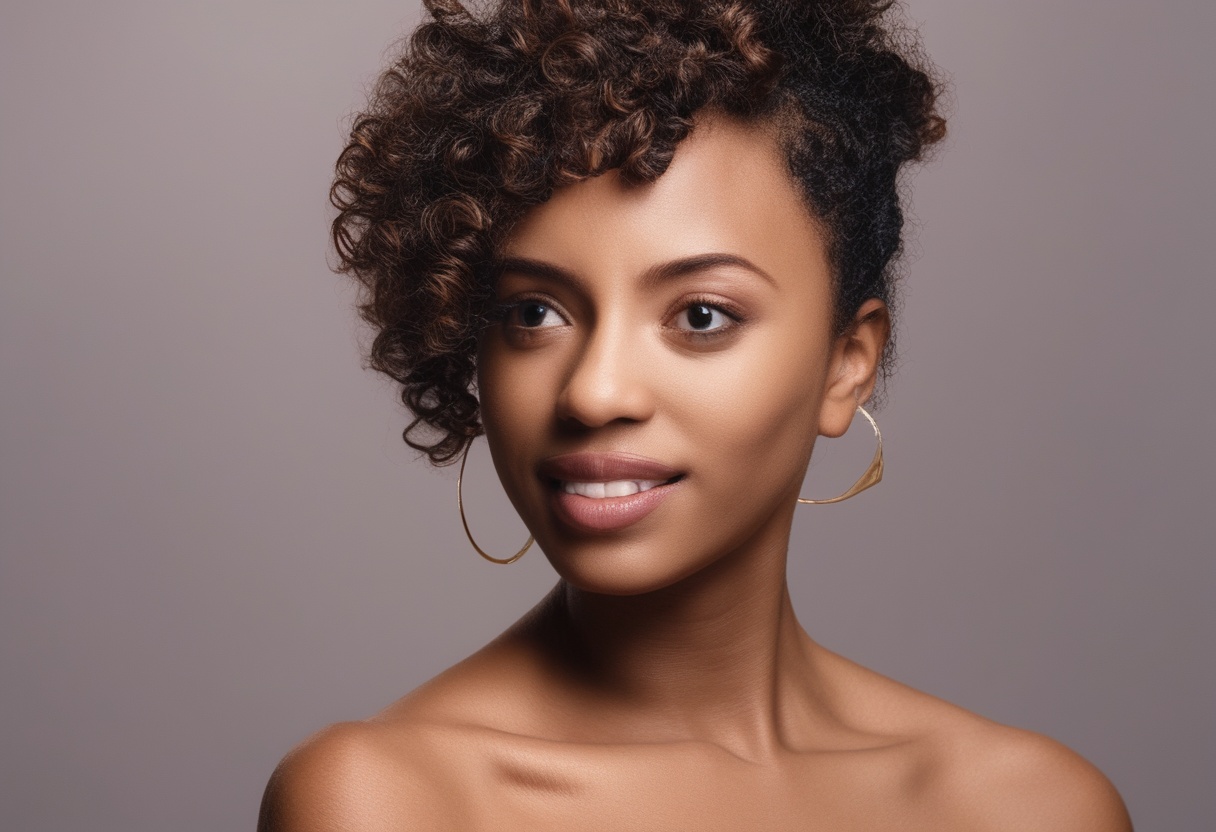}
 \\
 \midrule
 \multicolumn{3}{c|}{\begin{tabular}[c]{@{}c@{}} 8-bit GaLore (2.4GB)\end{tabular}} &\multicolumn{3}{c}{\begin{tabular}[c]{@{}c@{}}\textbf{8-bit COAP (0.5 GB)}\end{tabular}} \\
        \includegraphics[width=0.15\textwidth]{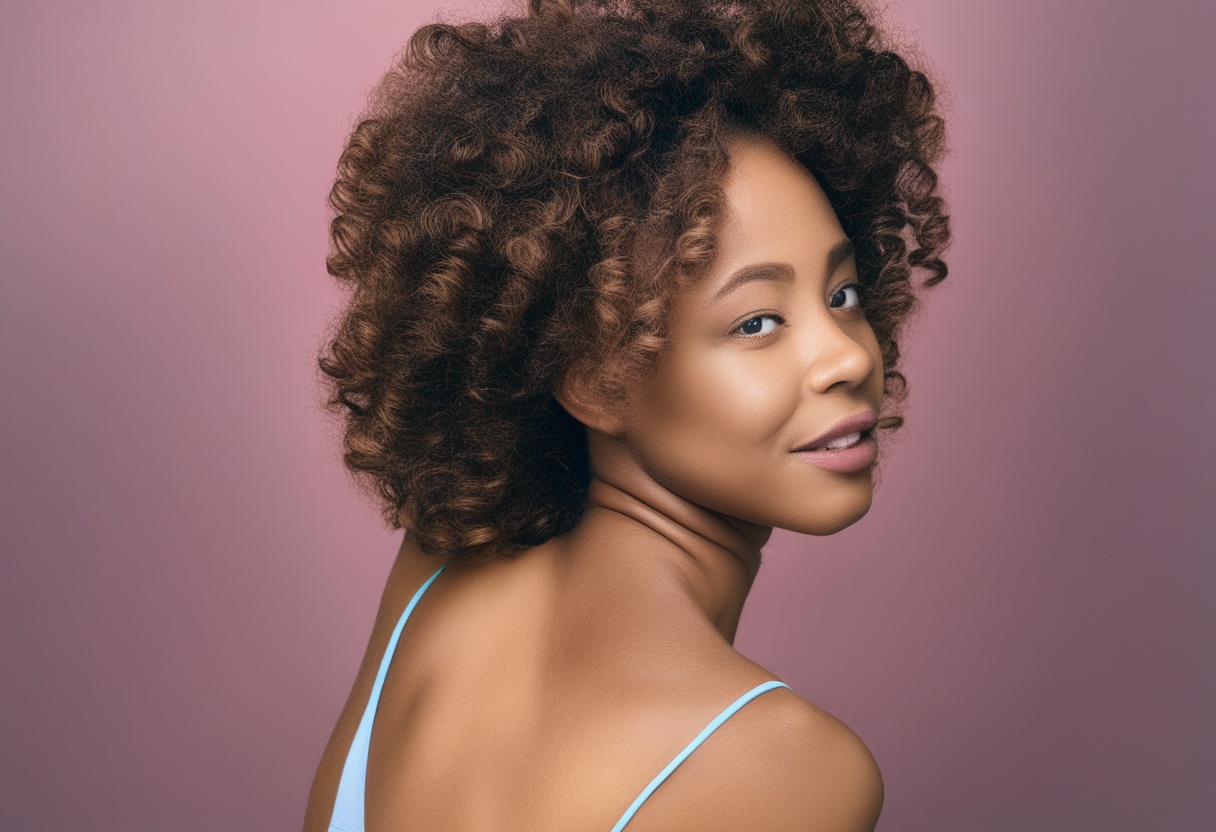} 
        & \includegraphics[width=0.15\textwidth]{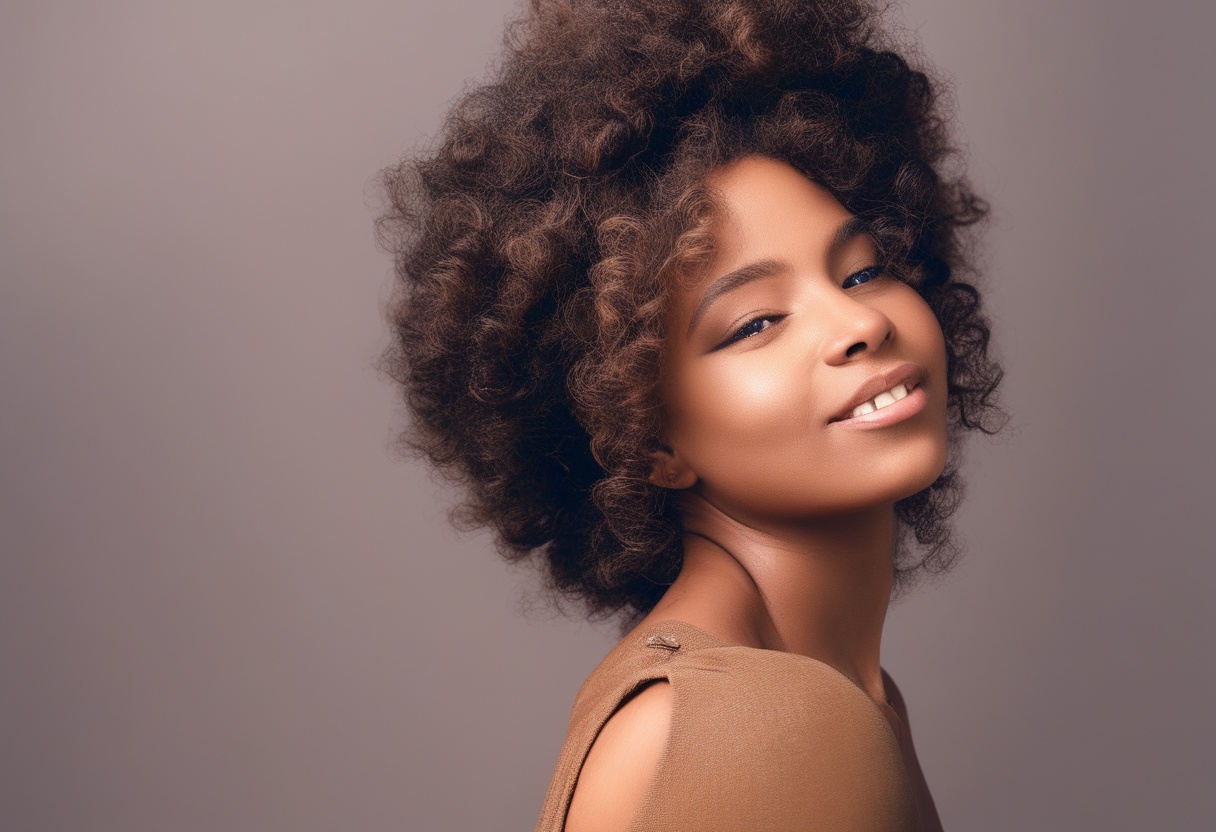} 
        & \includegraphics[width=0.15\textwidth]{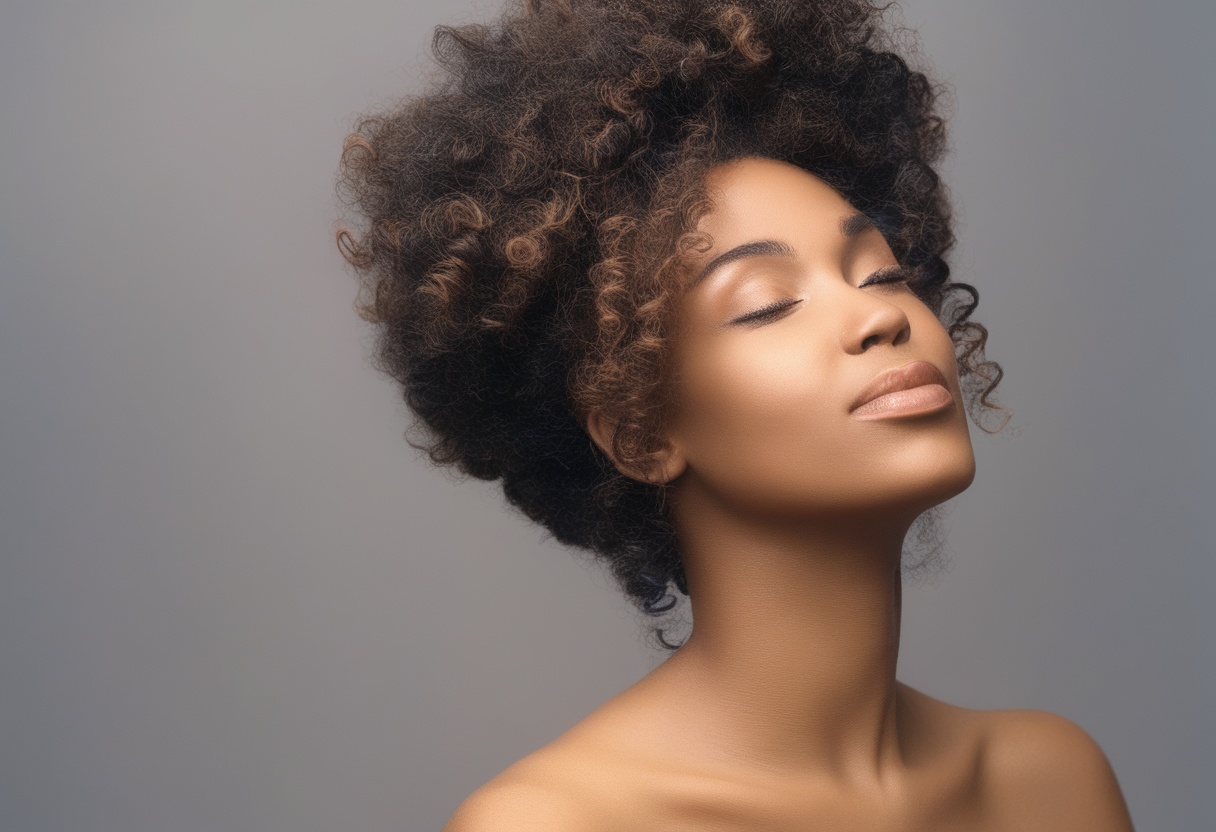}
        &\includegraphics[width=0.15\textwidth]{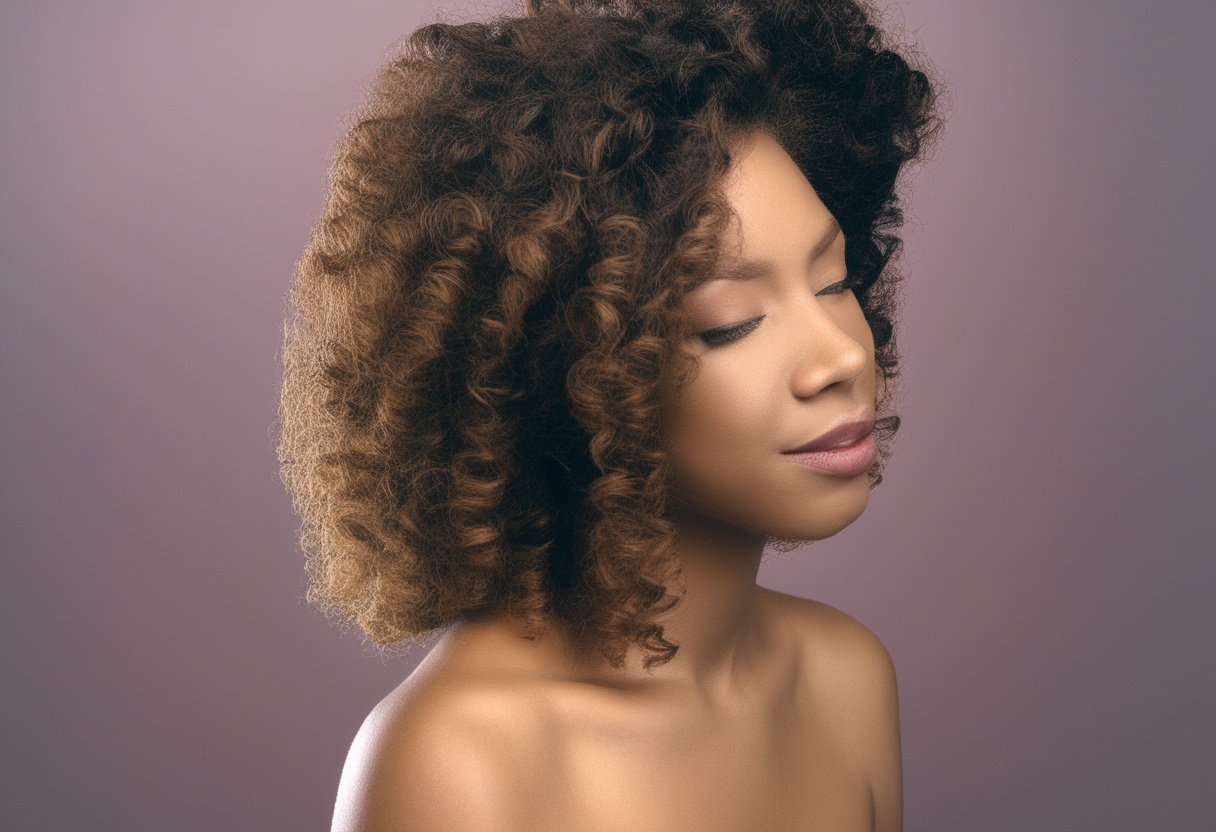} 
        & \includegraphics[width=0.15\textwidth]{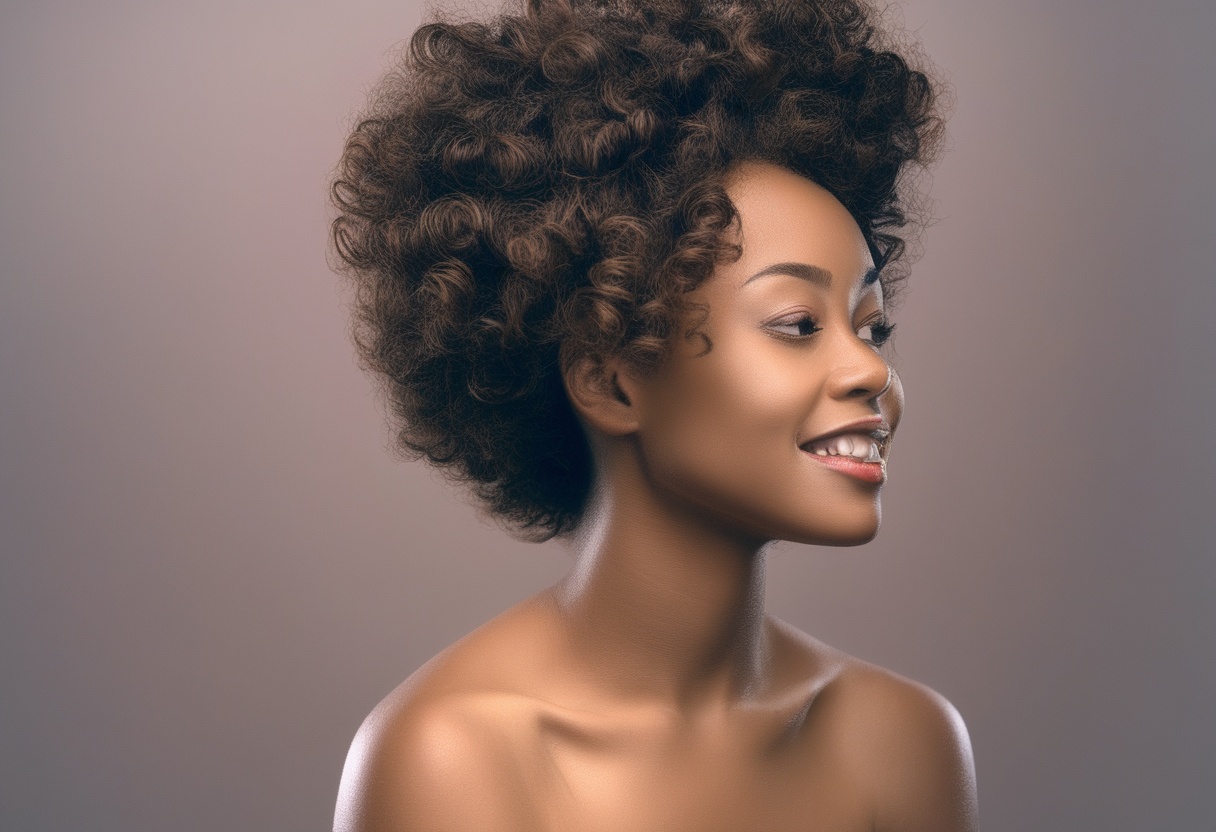} 
        & \includegraphics[width=0.15\textwidth]{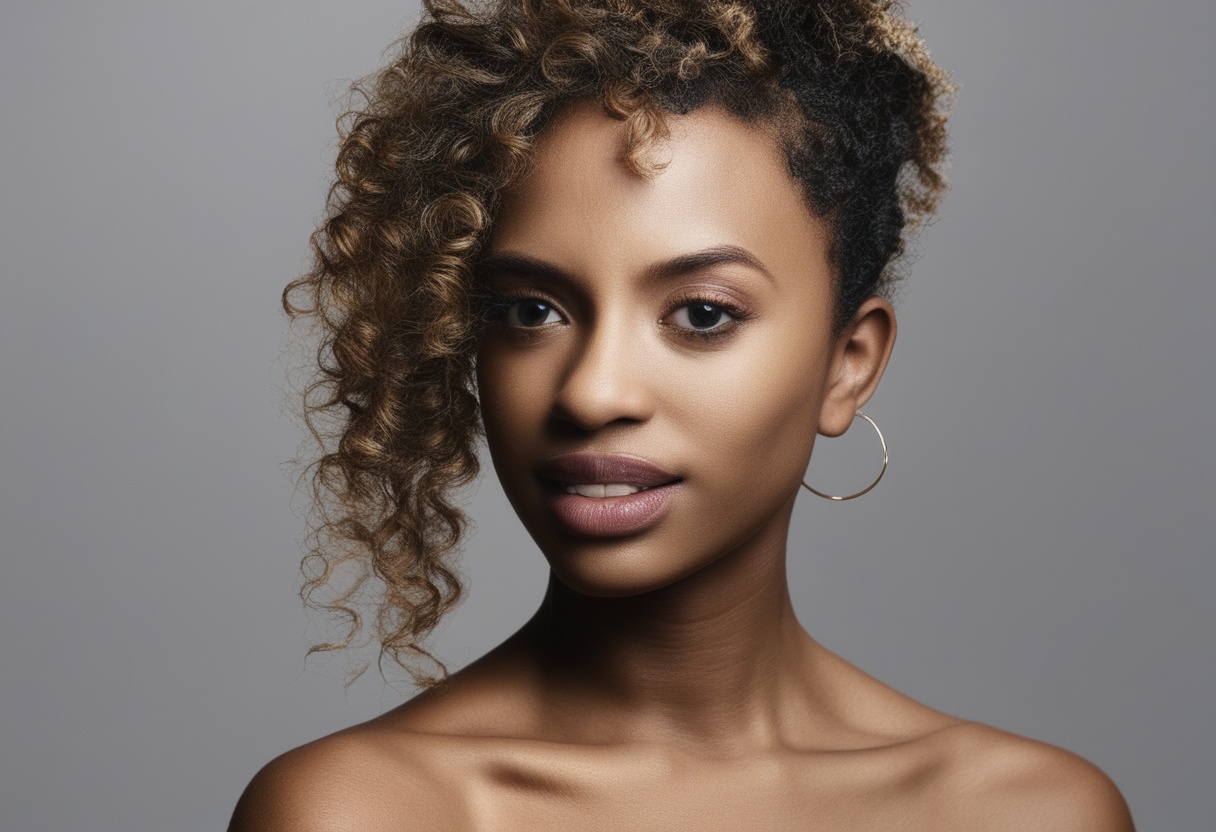} \\
        \bottomrule
    \end{tabular}
\end{table*}

\begin{table*}[ht]
    \centering
    \small
    \setlength{\tabcolsep}{2pt}
    \begin{tabular}{ccc|ccc}
        \toprule
        \multicolumn{6}{c}{Prompt: a man in a scarf and sweater leaning against a brick wall} \\
        \midrule
        \multirow{2}{*}{\begin{tabular}[c]{@{}c@{}}Human\\ pose\end{tabular}} & \multirow{2}{*}{\begin{tabular}[c]{@{}c@{}}Reference\end{tabular}} && 20K & 40K & 80K\\
        \cmidrule{4-6}
      &&&
        \multicolumn{3}{c}{\begin{tabular}[c]{@{}c@{}}Adafactor (5.1 GB)\end{tabular}} \\
         \includegraphics[width=0.15\textwidth]{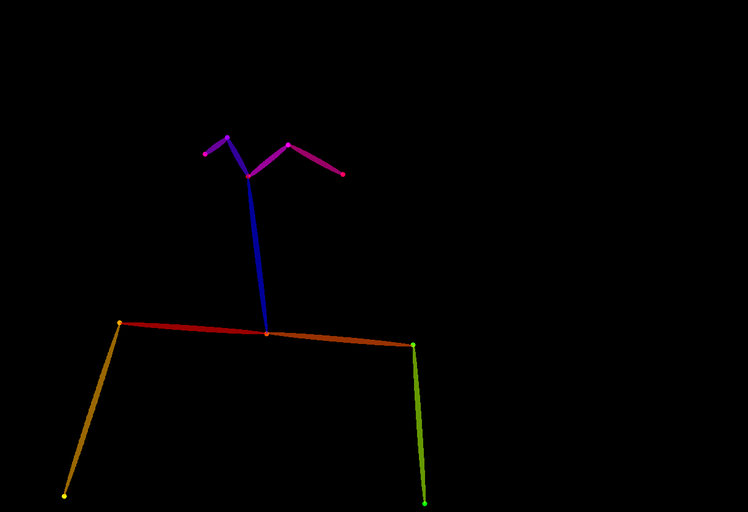} 
        & \includegraphics[width=0.15\textwidth]{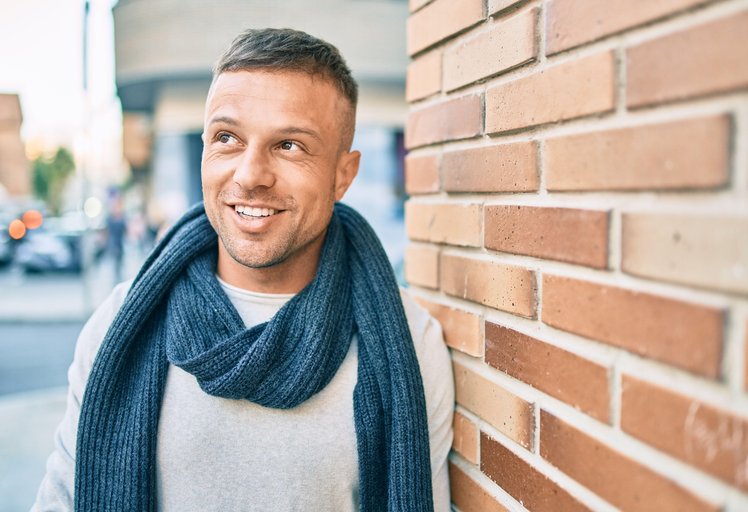} &
        &\includegraphics[width=0.15\textwidth]{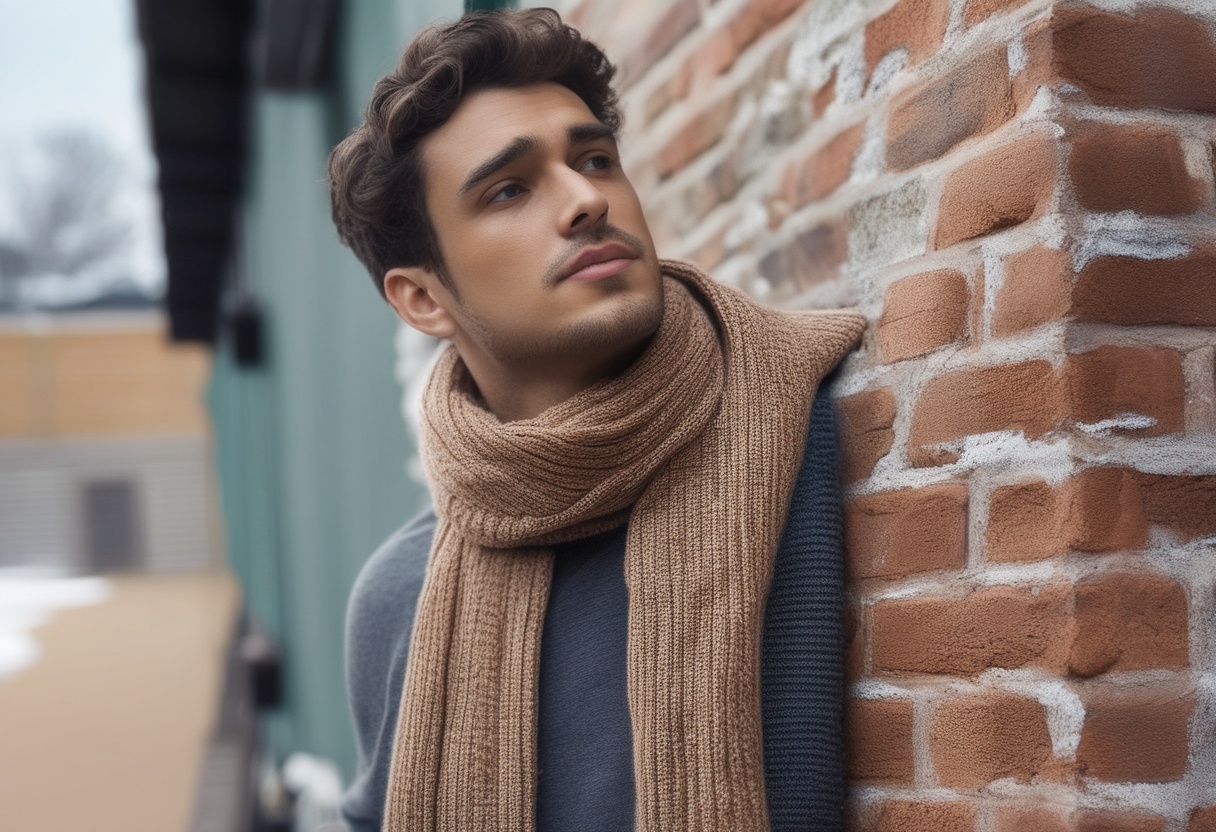} 
        & \includegraphics[width=0.15\textwidth]{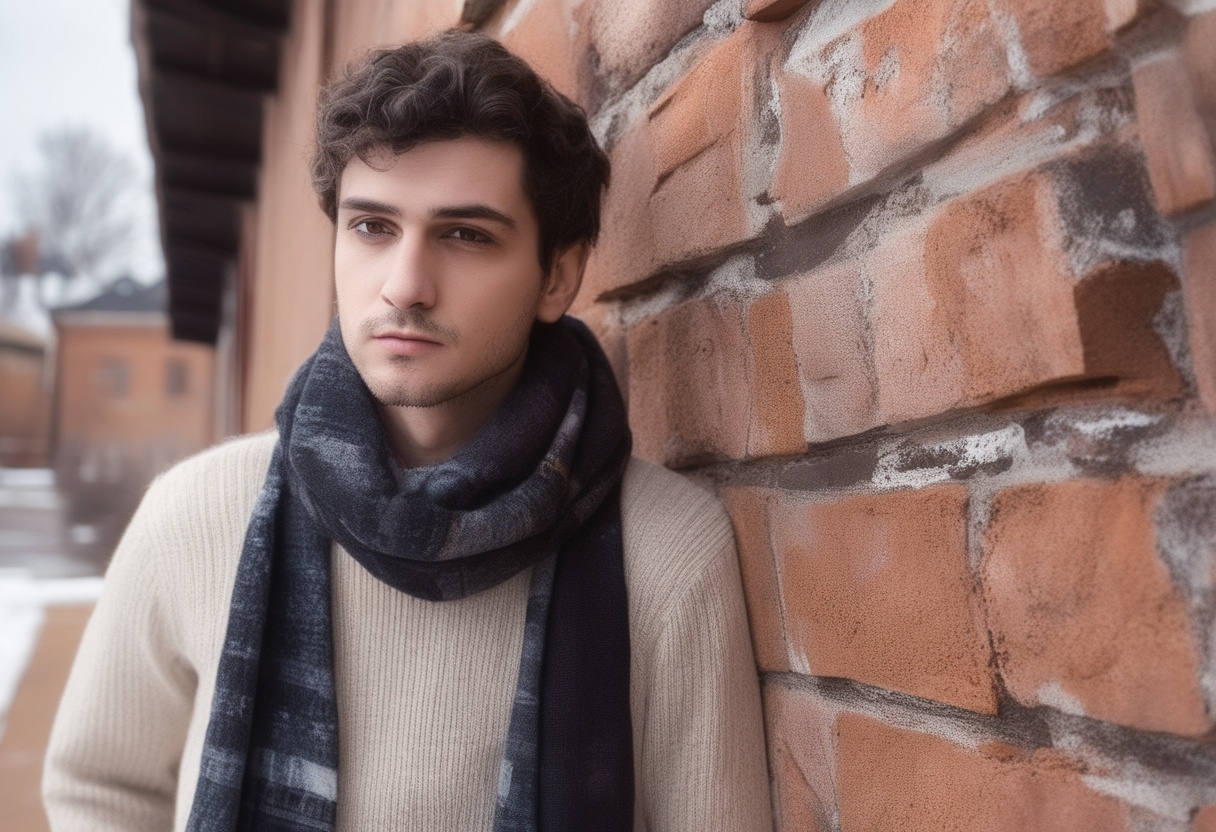} 
        & \includegraphics[width=0.15\textwidth]{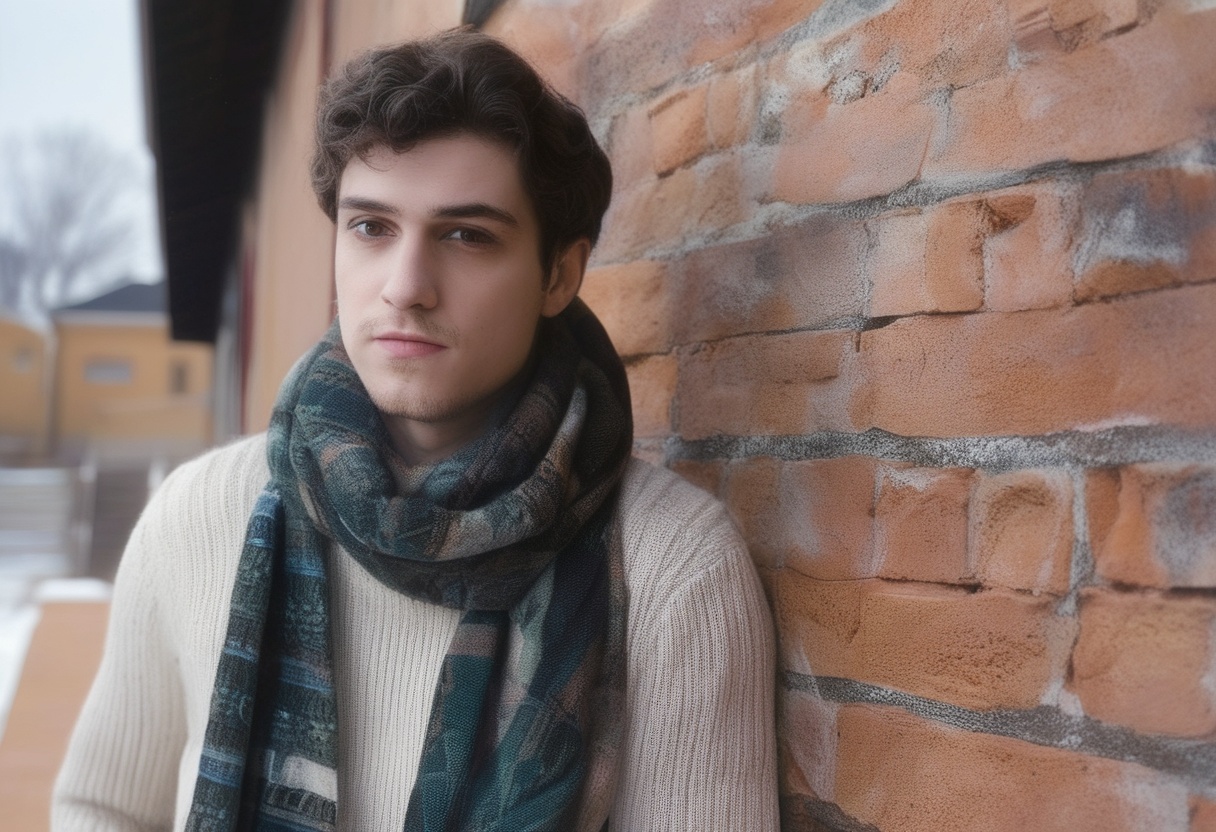}\\
        \midrule
        \multicolumn{3}{c|}{\begin{tabular}[c]{@{}c@{}}GaLore (4.7 GB)\end{tabular}} &\multicolumn{3}{c}{\begin{tabular}[c]{@{}c@{}}\textbf{COAP (3.6 GB)}\end{tabular}} \\
        \includegraphics[width=0.15\textwidth]{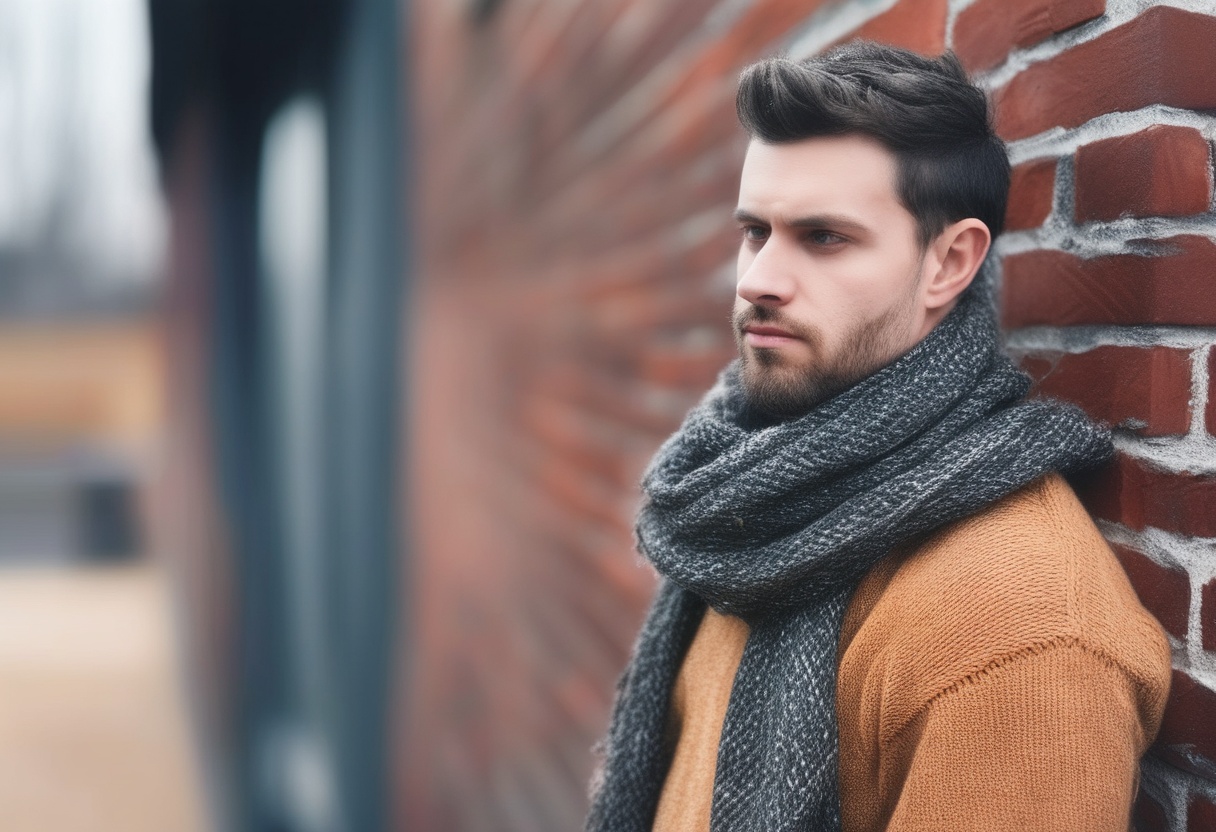} 
        & \includegraphics[width=0.15\textwidth]{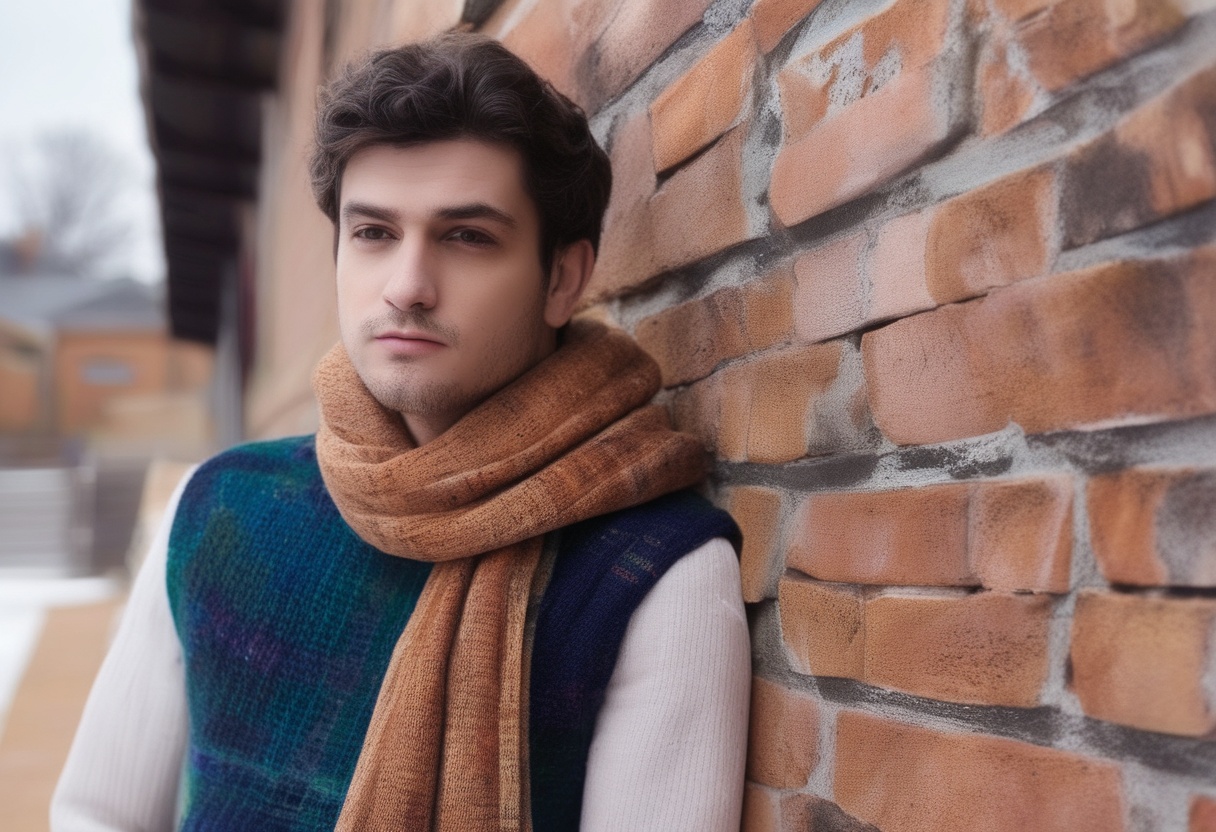} 
        & \includegraphics[width=0.15\textwidth]{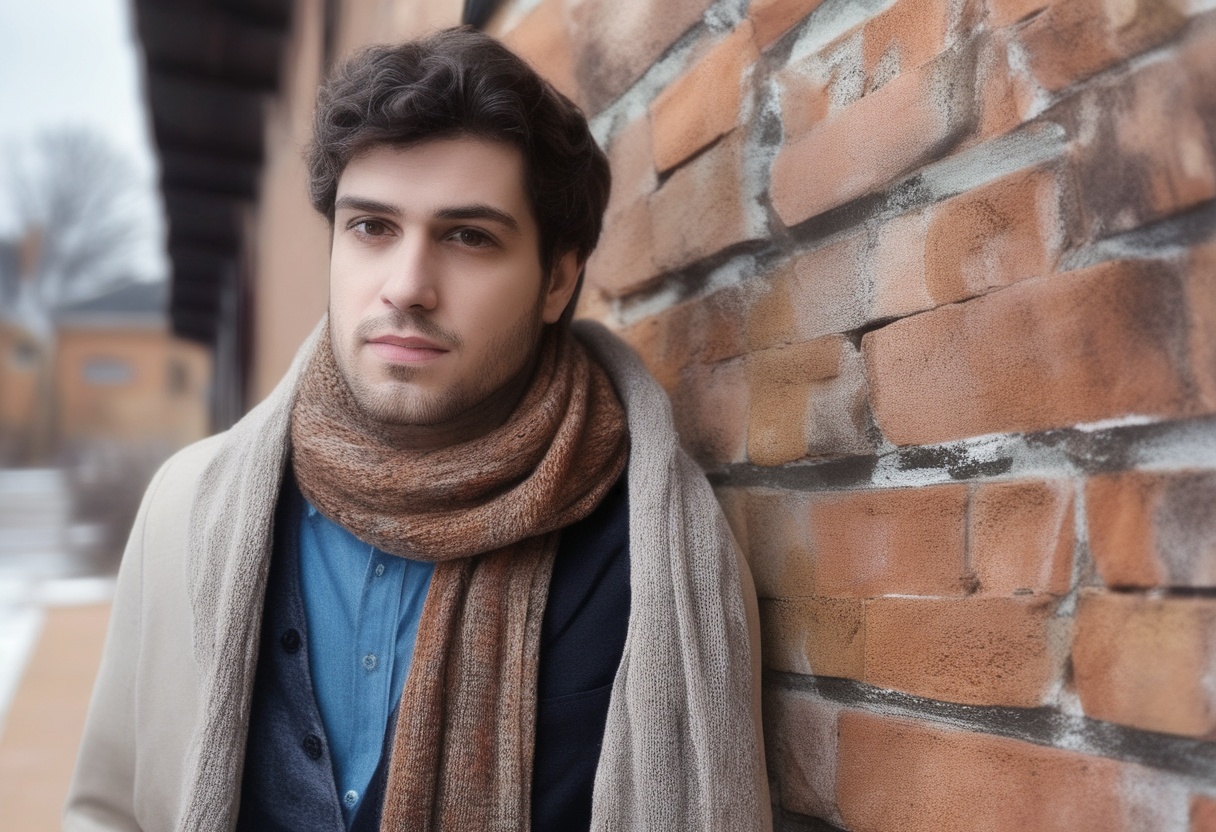}
        &\includegraphics[width=0.15\textwidth]{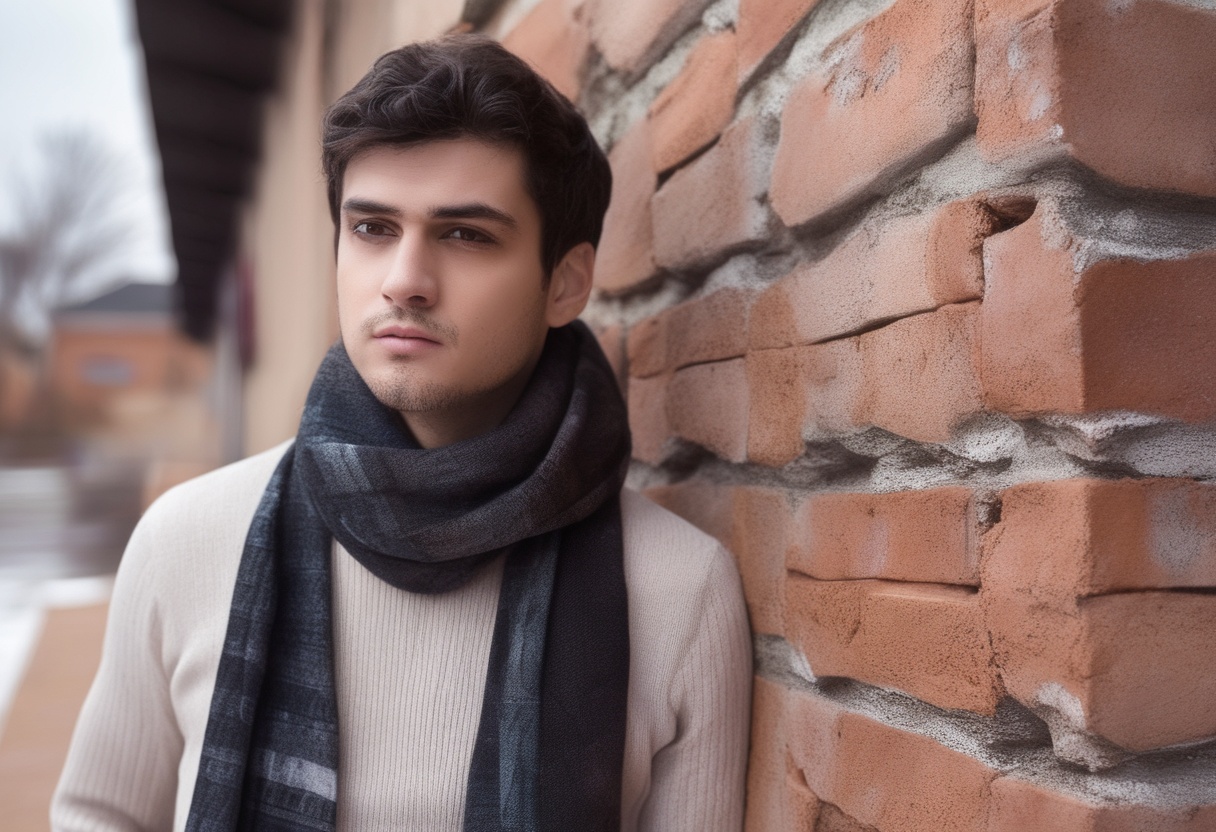} 
        & \includegraphics[width=0.15\textwidth]{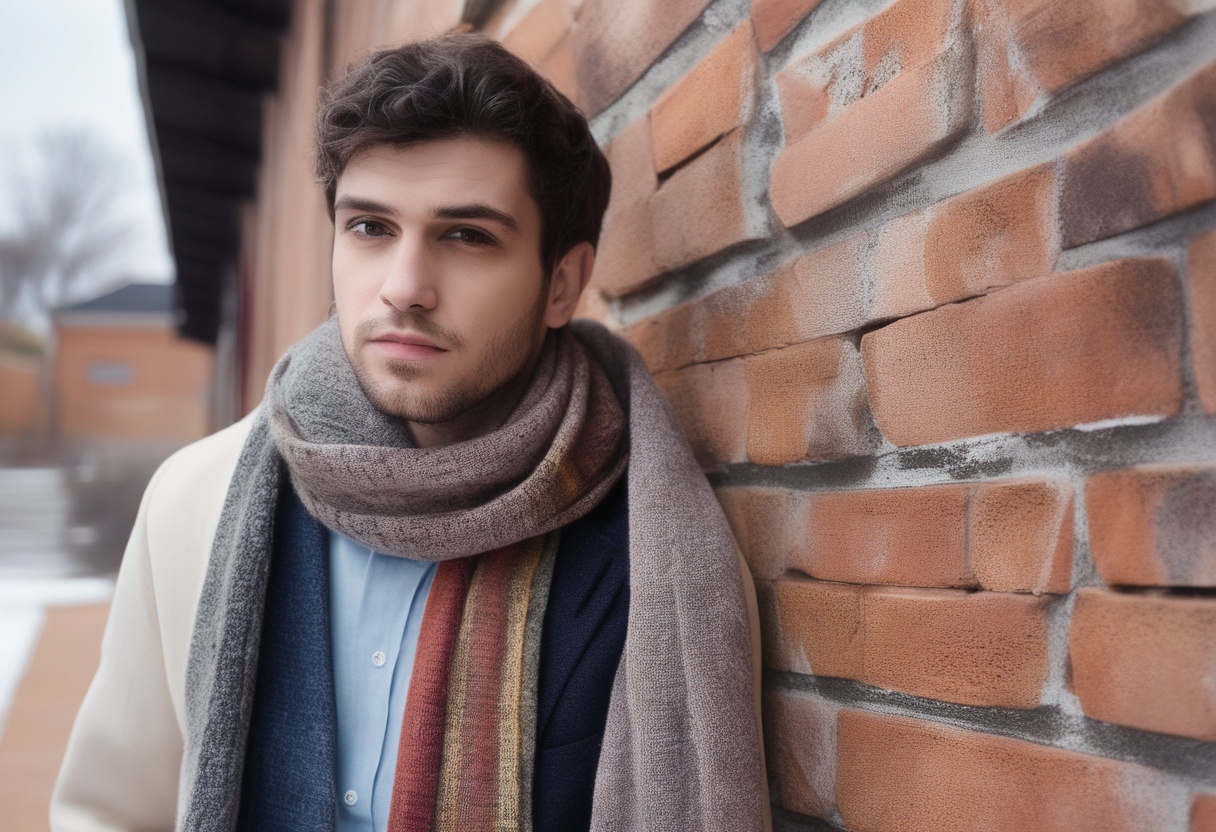} 
        & \includegraphics[width=0.15\textwidth]{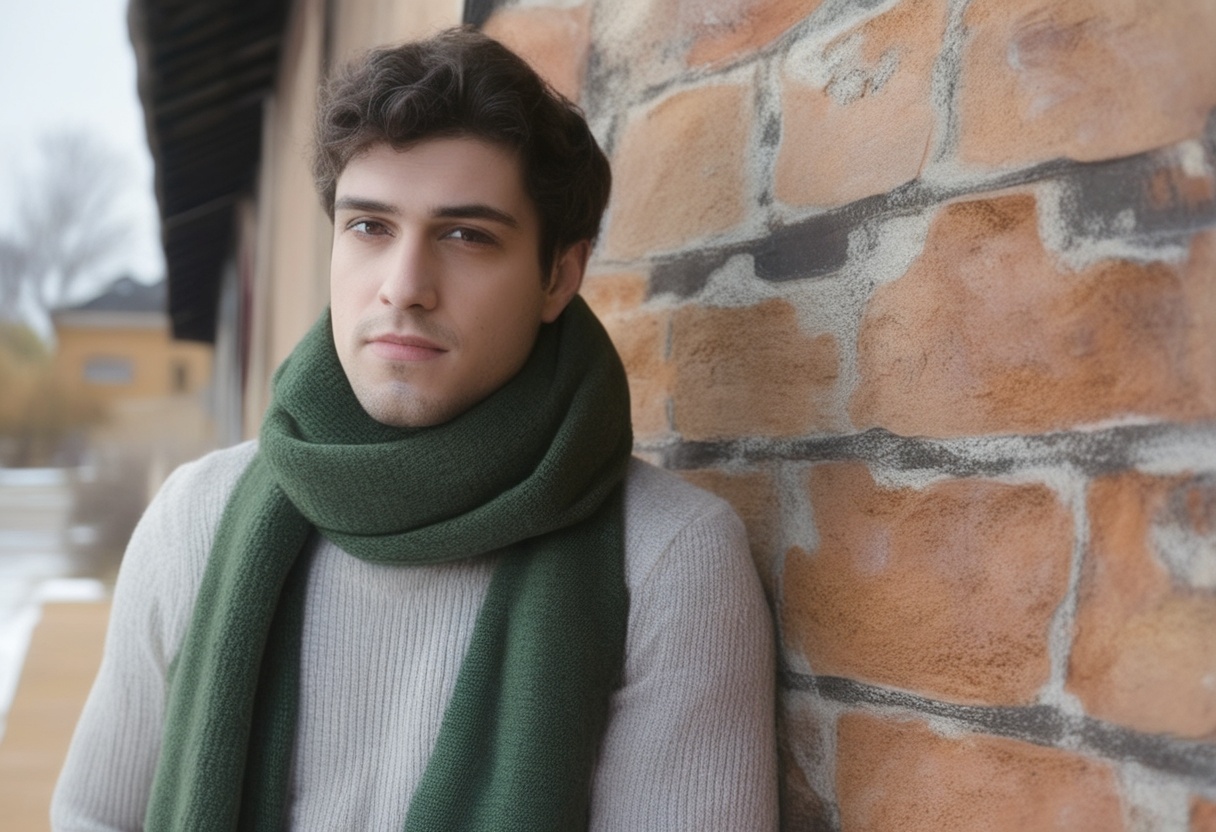}
 \\
 \midrule
 \multicolumn{3}{c|}{\begin{tabular}[c]{@{}c@{}} 8-bit GaLore (2.4GB)\end{tabular}} &\multicolumn{3}{c}{\begin{tabular}[c]{@{}c@{}}\textbf{8-bit COAP (0.5 GB)}\end{tabular}} \\
        \includegraphics[width=0.15\textwidth]{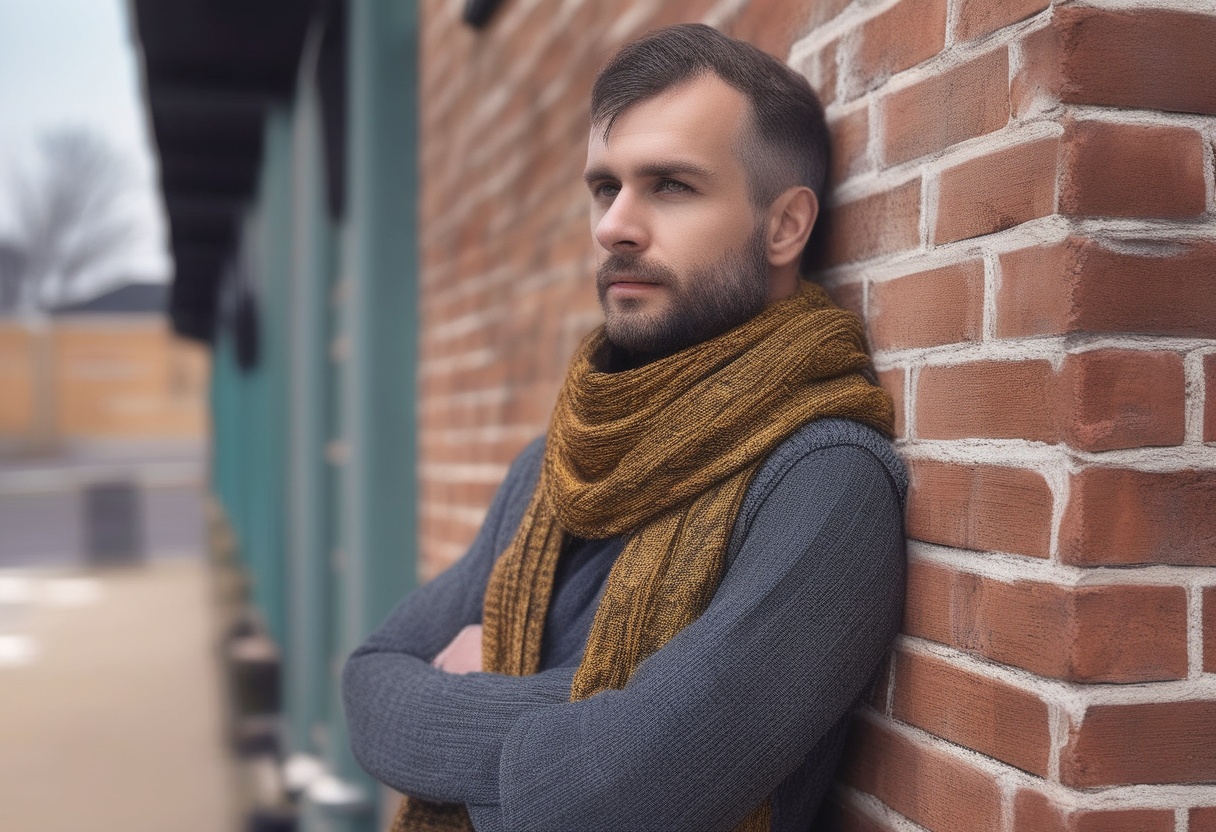} 
        & \includegraphics[width=0.15\textwidth]{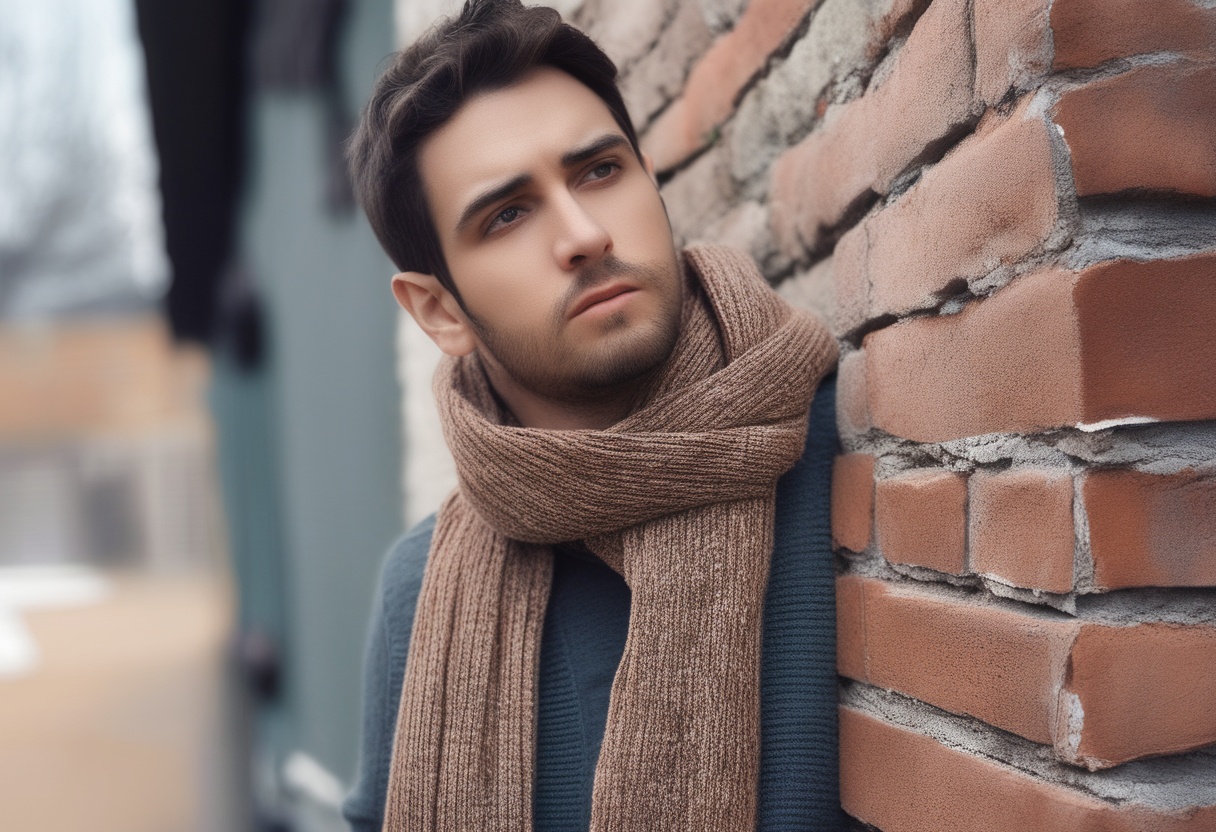} 
        & \includegraphics[width=0.15\textwidth]{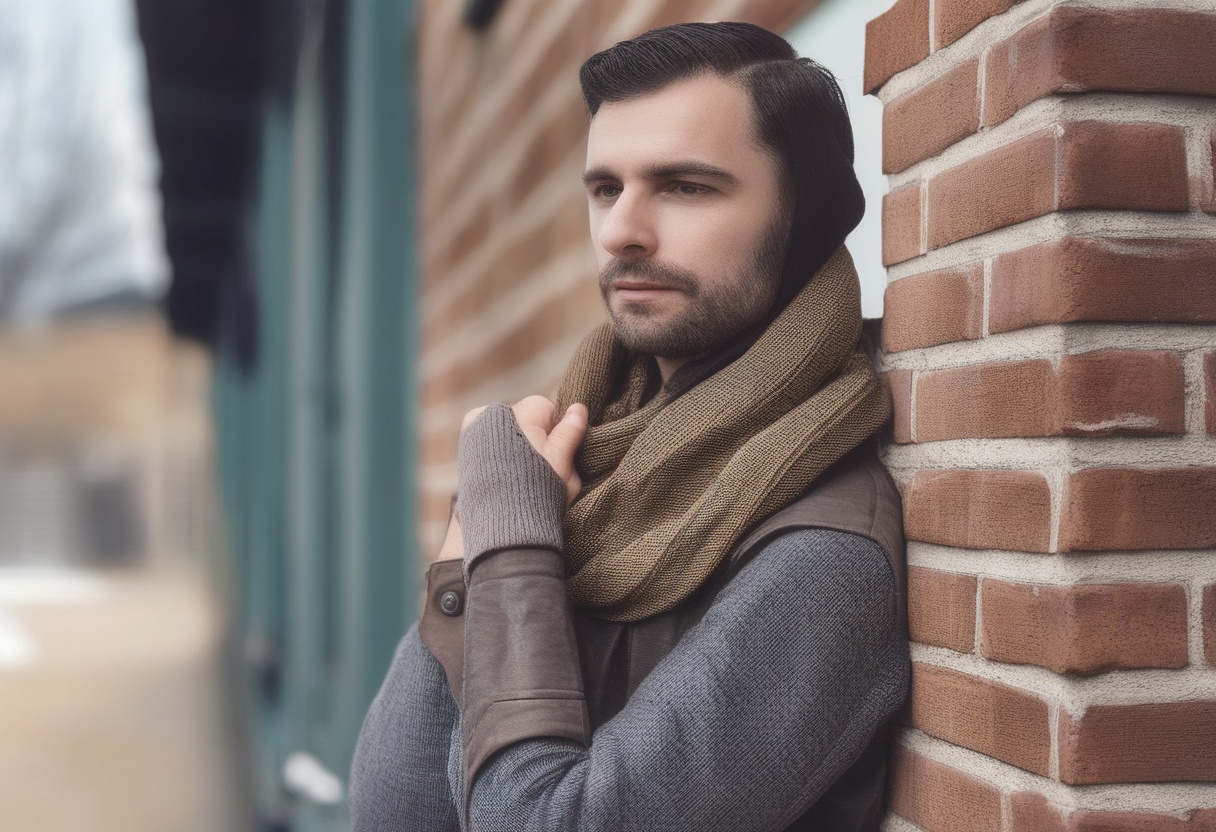}
        &\includegraphics[width=0.15\textwidth]{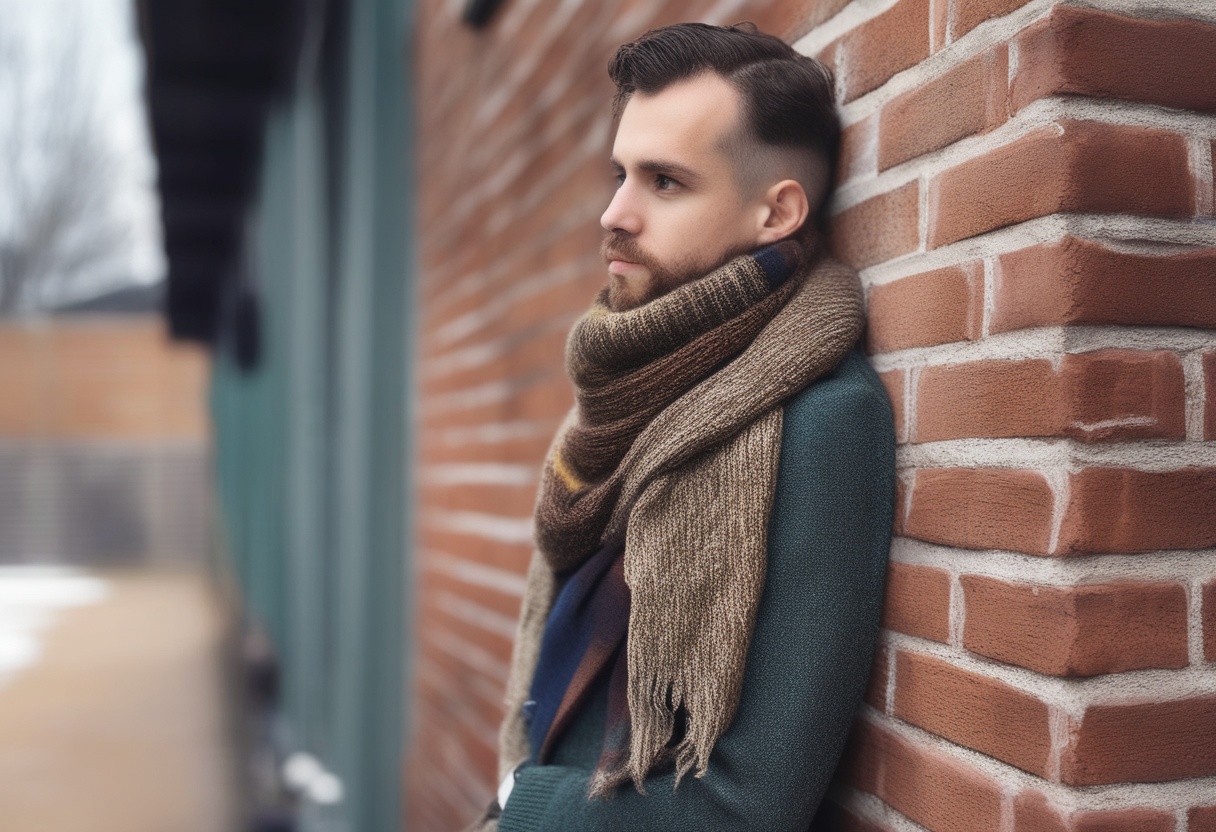} 
        & \includegraphics[width=0.15\textwidth]{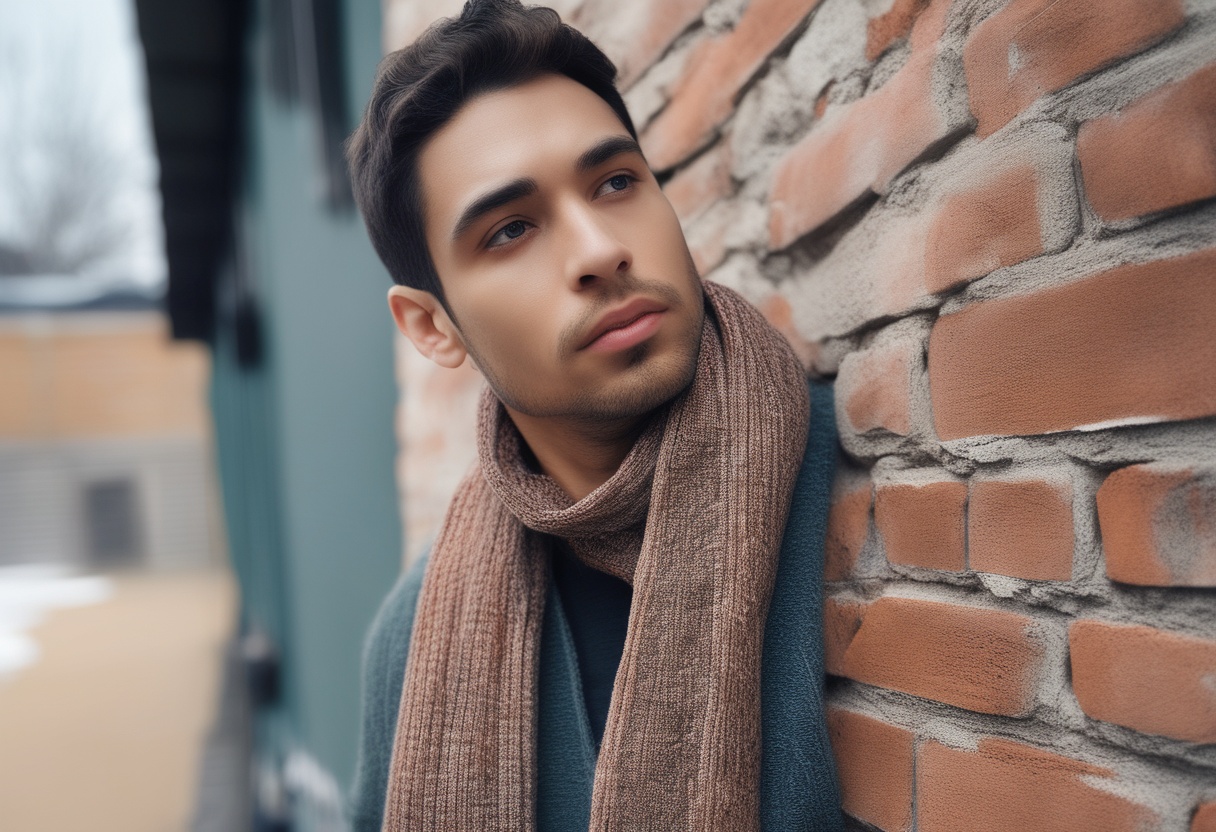} 
        & \includegraphics[width=0.15\textwidth]{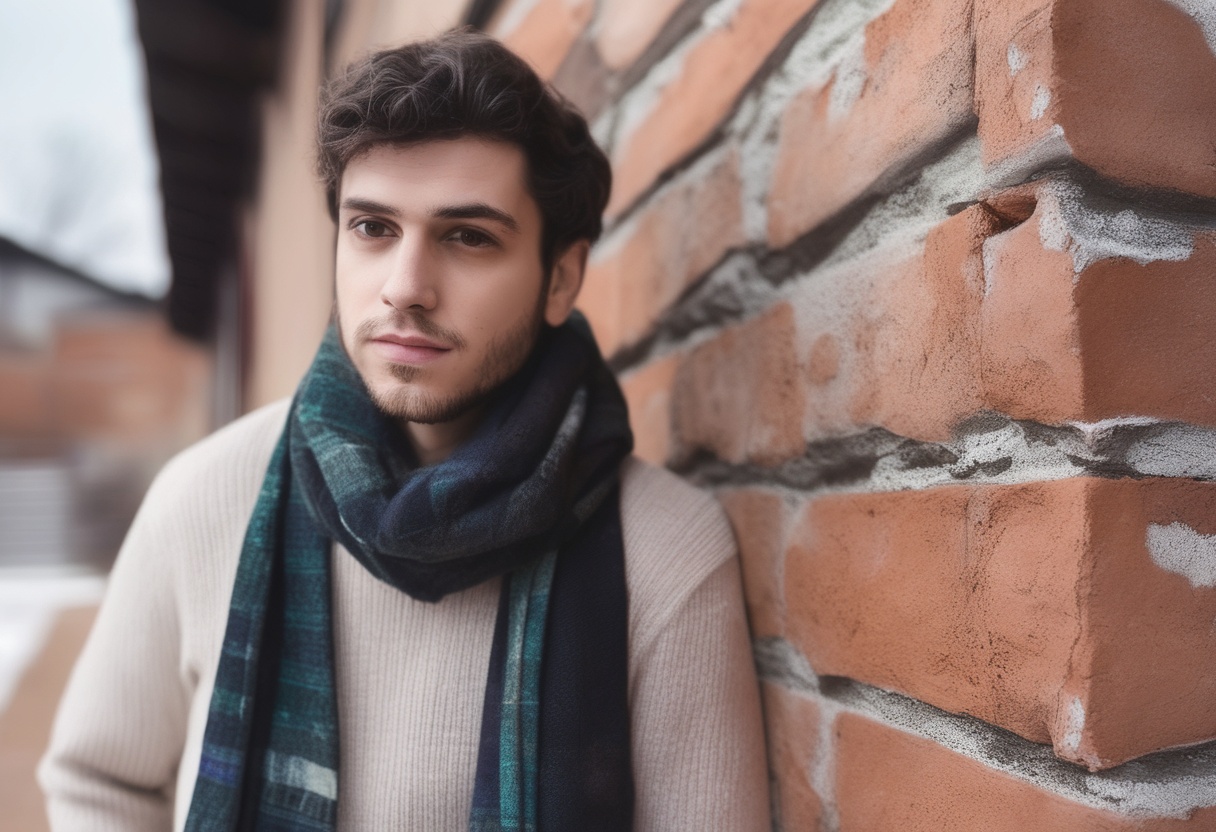} \\
        \bottomrule
    \end{tabular}
\end{table*}

\begin{table*}[ht]
    \centering
    \small
    \setlength{\tabcolsep}{2pt}
    \begin{tabular}{ccc|ccc}
        \toprule
        \multicolumn{6}{c}{Prompt: a young girl with her hands painted with colorful paint} \\
        \midrule
        \multirow{2}{*}{\begin{tabular}[c]{@{}c@{}}Human\\ pose\end{tabular}} & \multirow{2}{*}{\begin{tabular}[c]{@{}c@{}}Reference\end{tabular}} && 20K & 40K & 80K\\
        \cmidrule{4-6}
      &&&
        \multicolumn{3}{c}{\begin{tabular}[c]{@{}c@{}}Adafactor (5.1 GB)\end{tabular}} \\
         \includegraphics[width=0.15\textwidth]{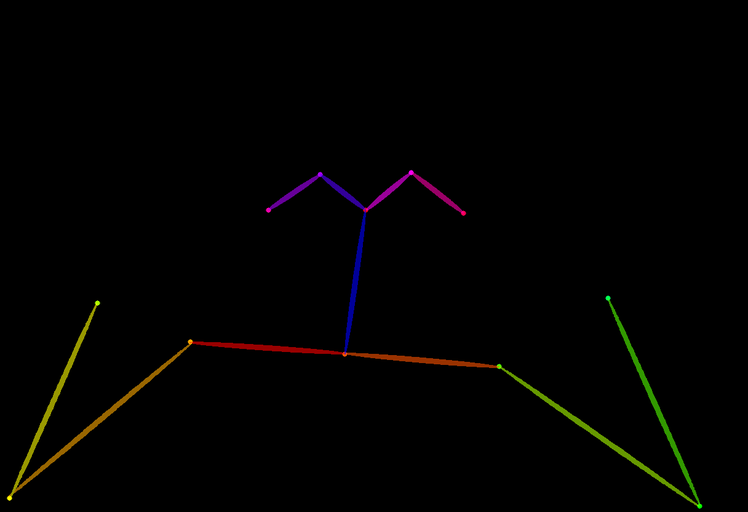} 
        & \includegraphics[width=0.15\textwidth]{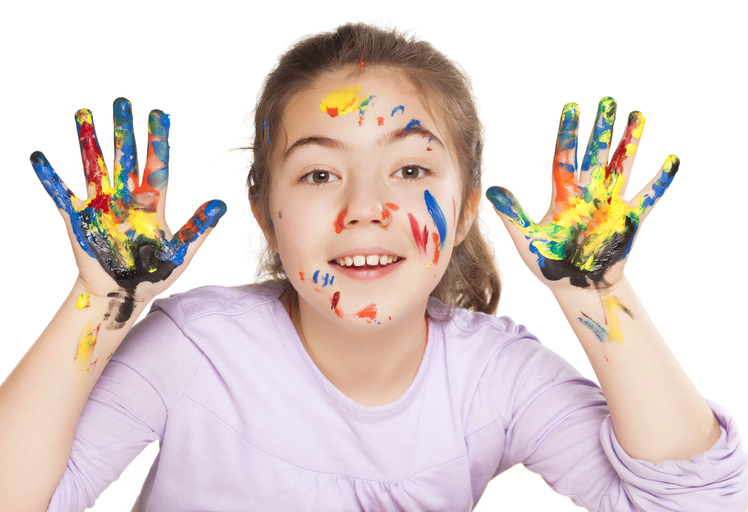} &
        &\includegraphics[width=0.15\textwidth]{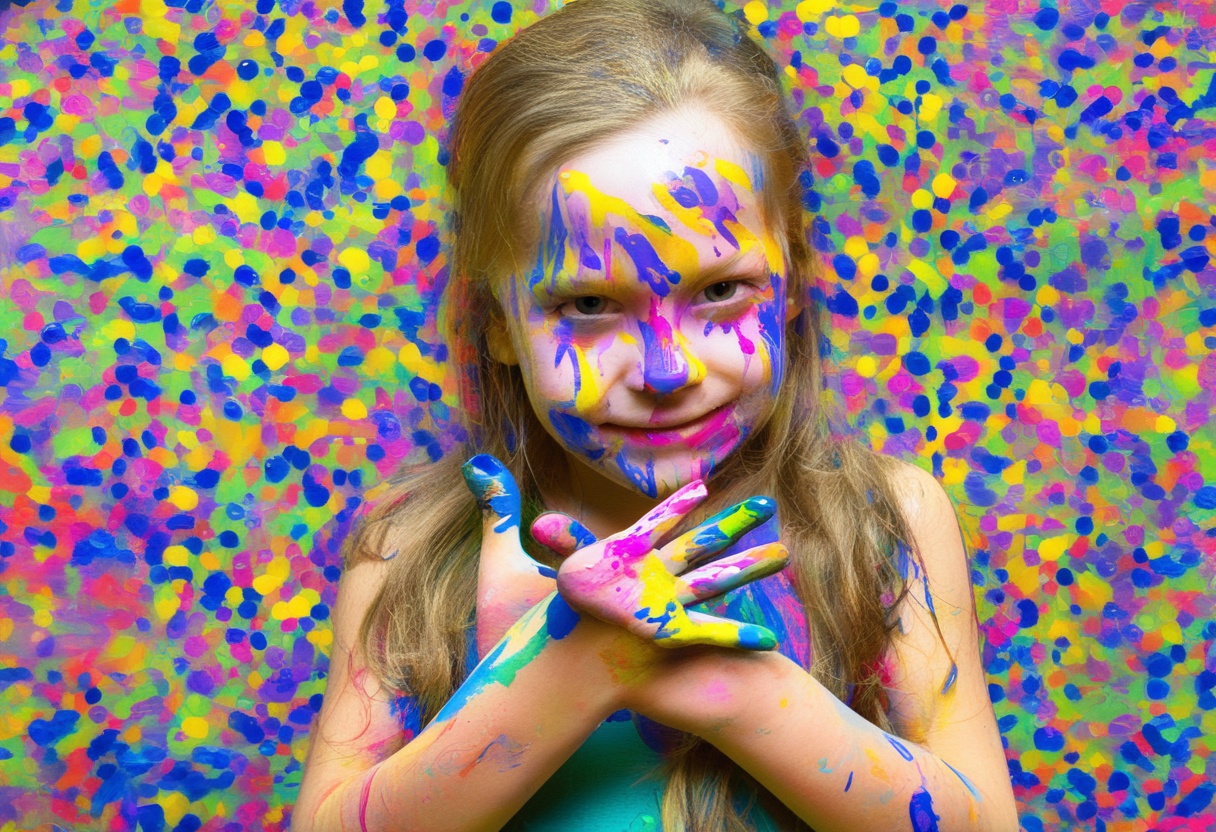} 
        & \includegraphics[width=0.15\textwidth]{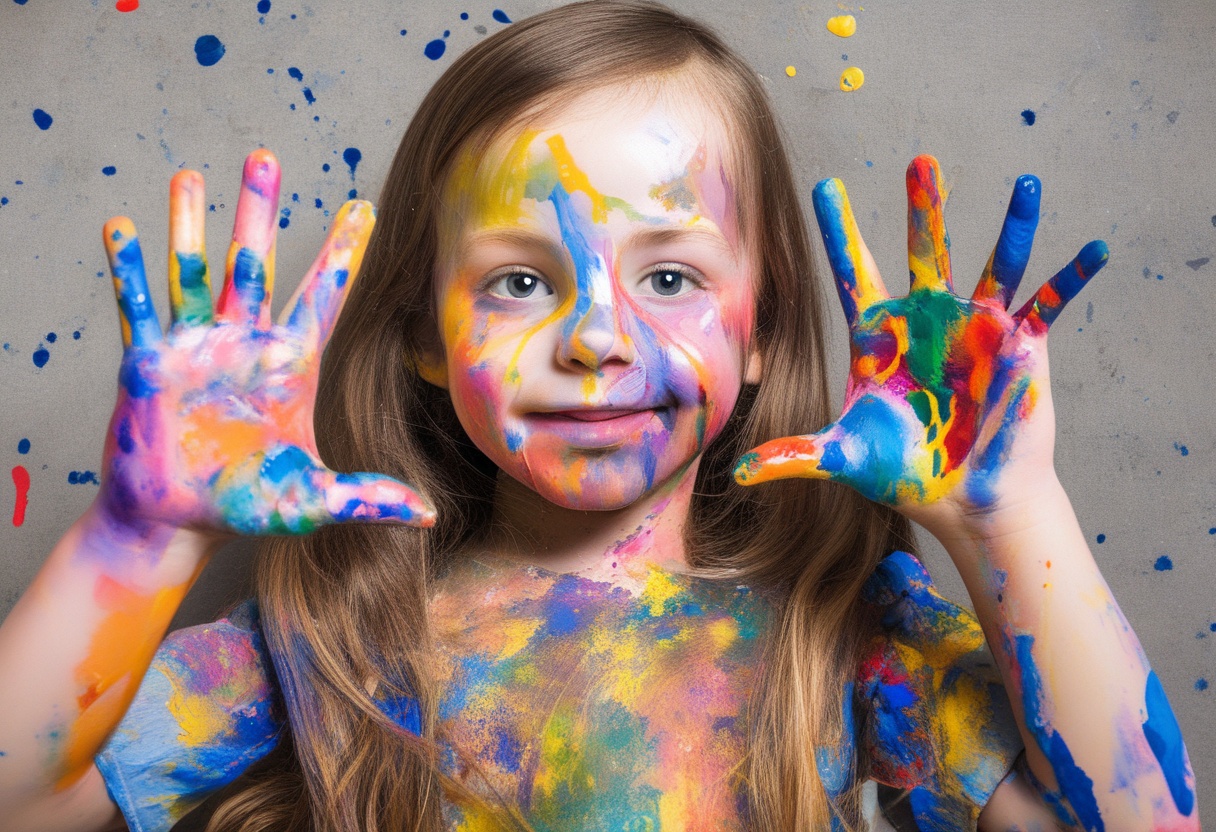} 
        & \includegraphics[width=0.15\textwidth]{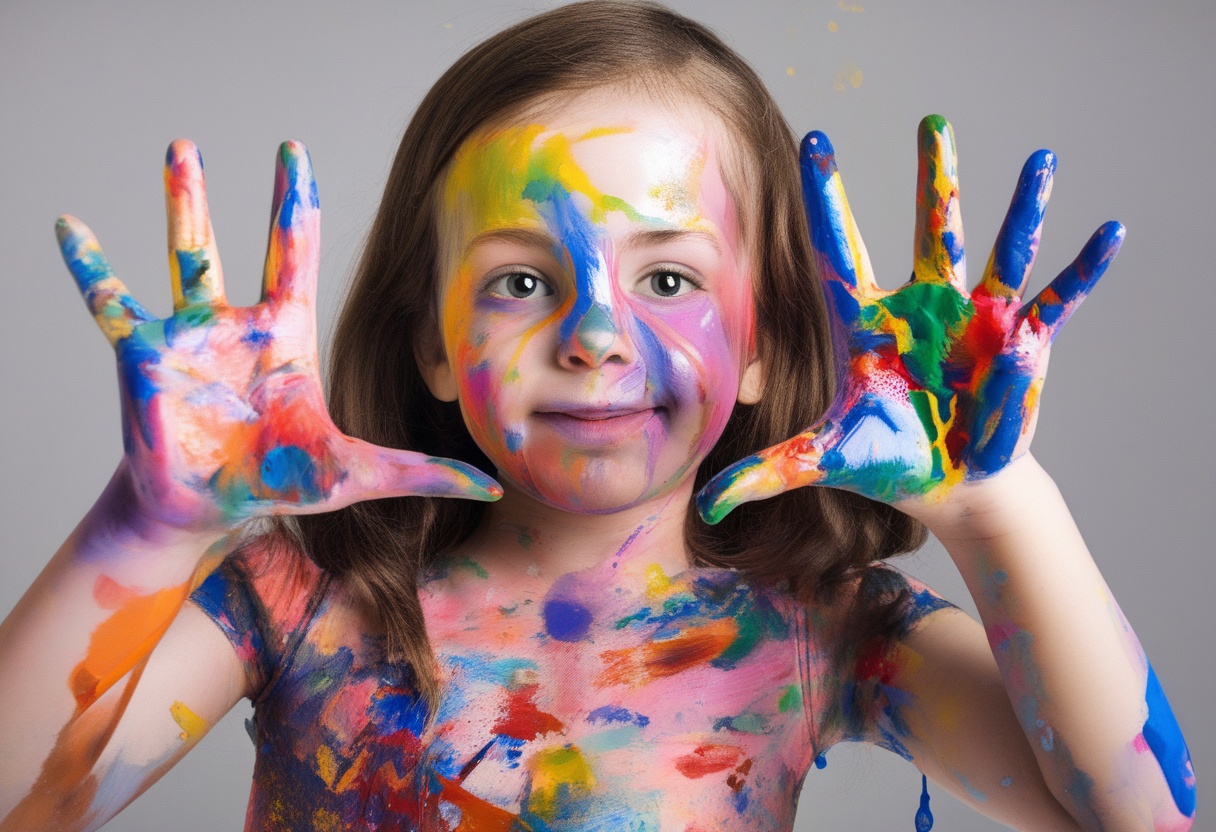}\\
        \midrule
        \multicolumn{3}{c|}{\begin{tabular}[c]{@{}c@{}}GaLore (4.7 GB)\end{tabular}} &\multicolumn{3}{c}{\begin{tabular}[c]{@{}c@{}}\textbf{COAP (3.6 GB)}\end{tabular}} \\
        \includegraphics[width=0.15\textwidth]{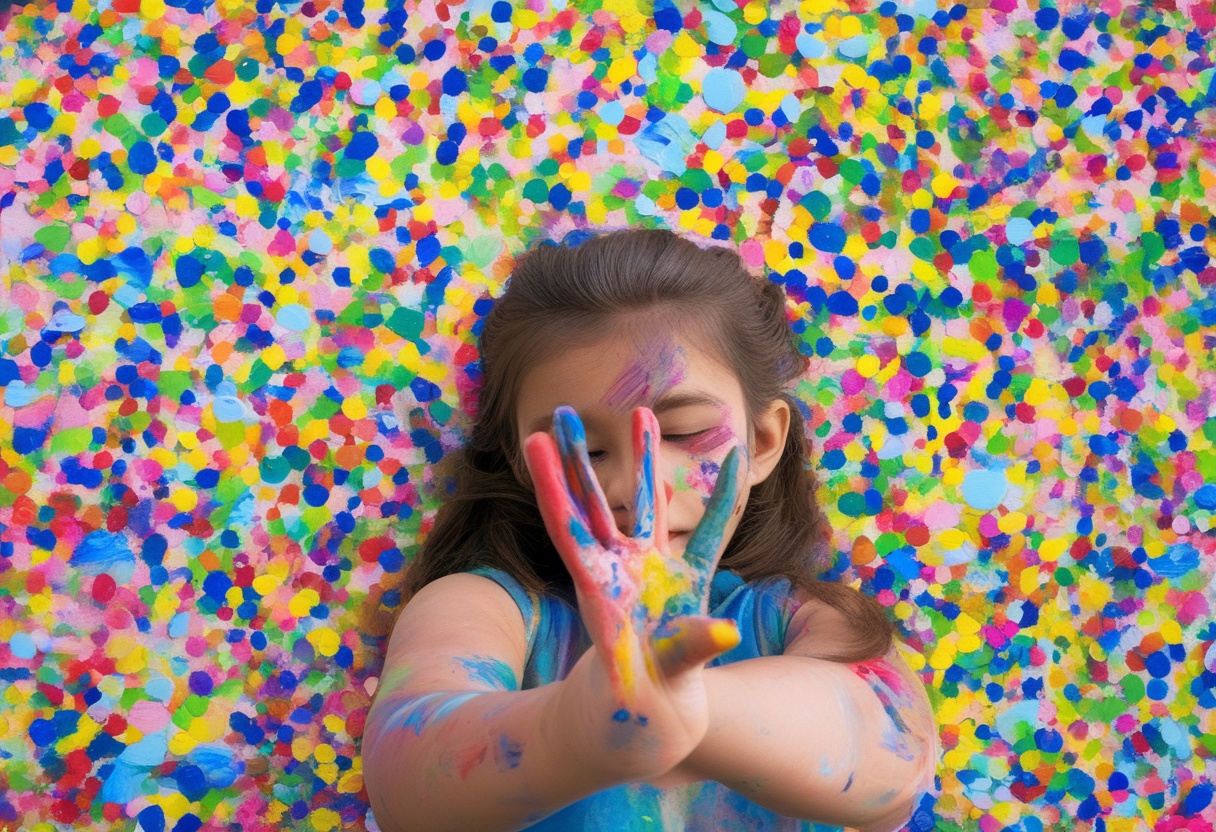} 
        & \includegraphics[width=0.15\textwidth]{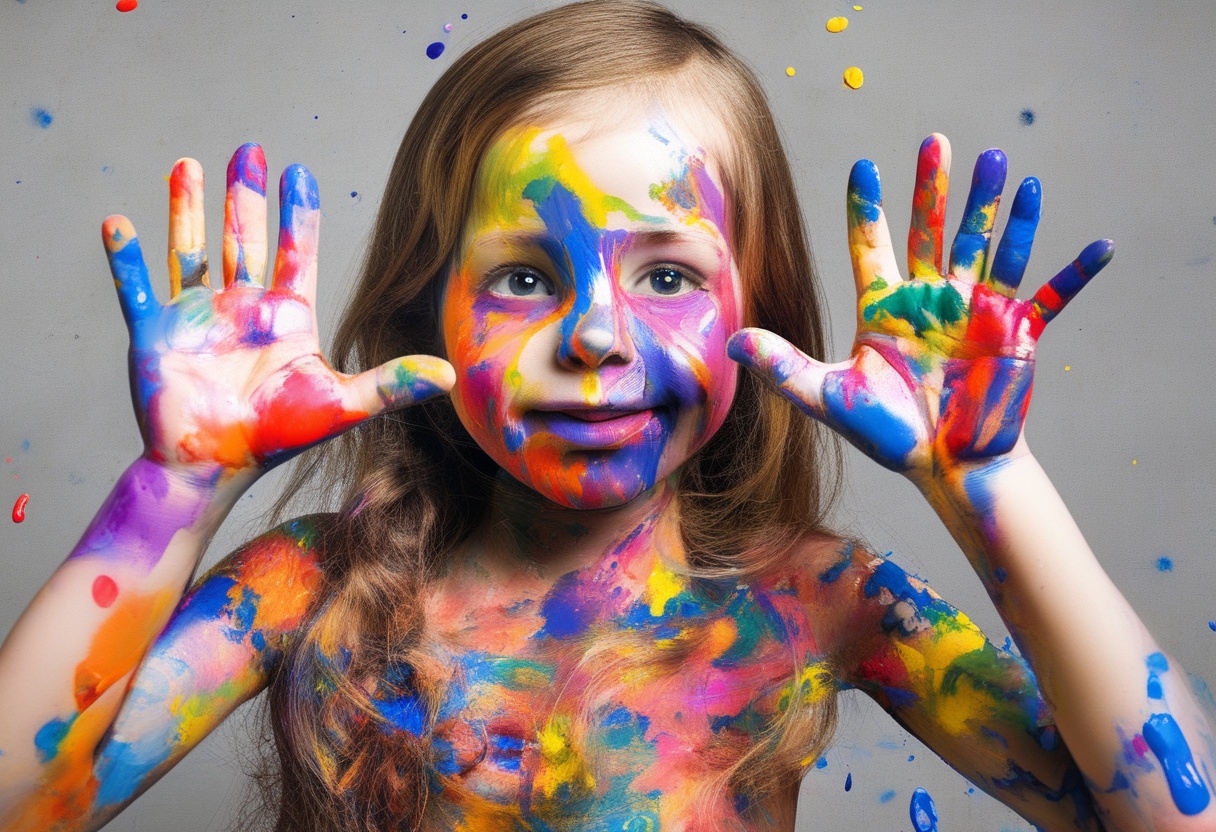} 
        & \includegraphics[width=0.15\textwidth]{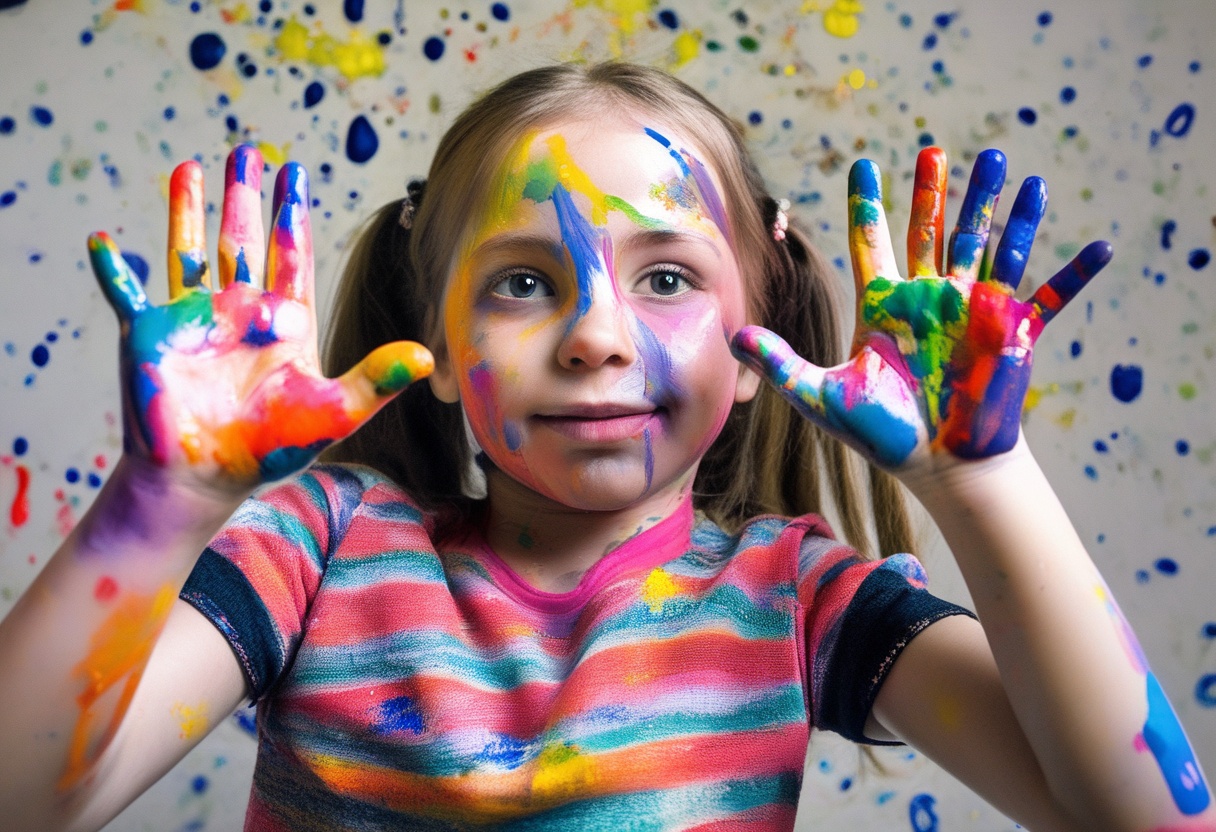}
        &\includegraphics[width=0.15\textwidth]{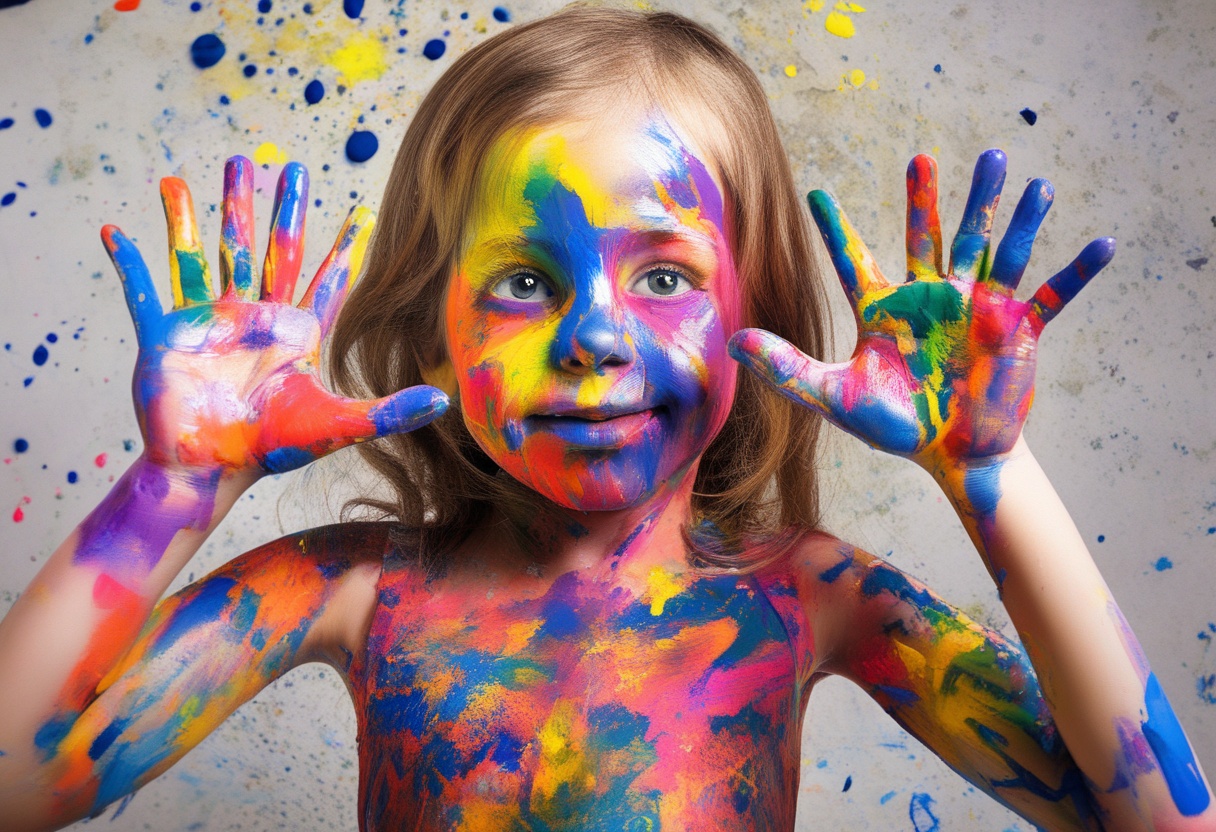} 
        & \includegraphics[width=0.15\textwidth]{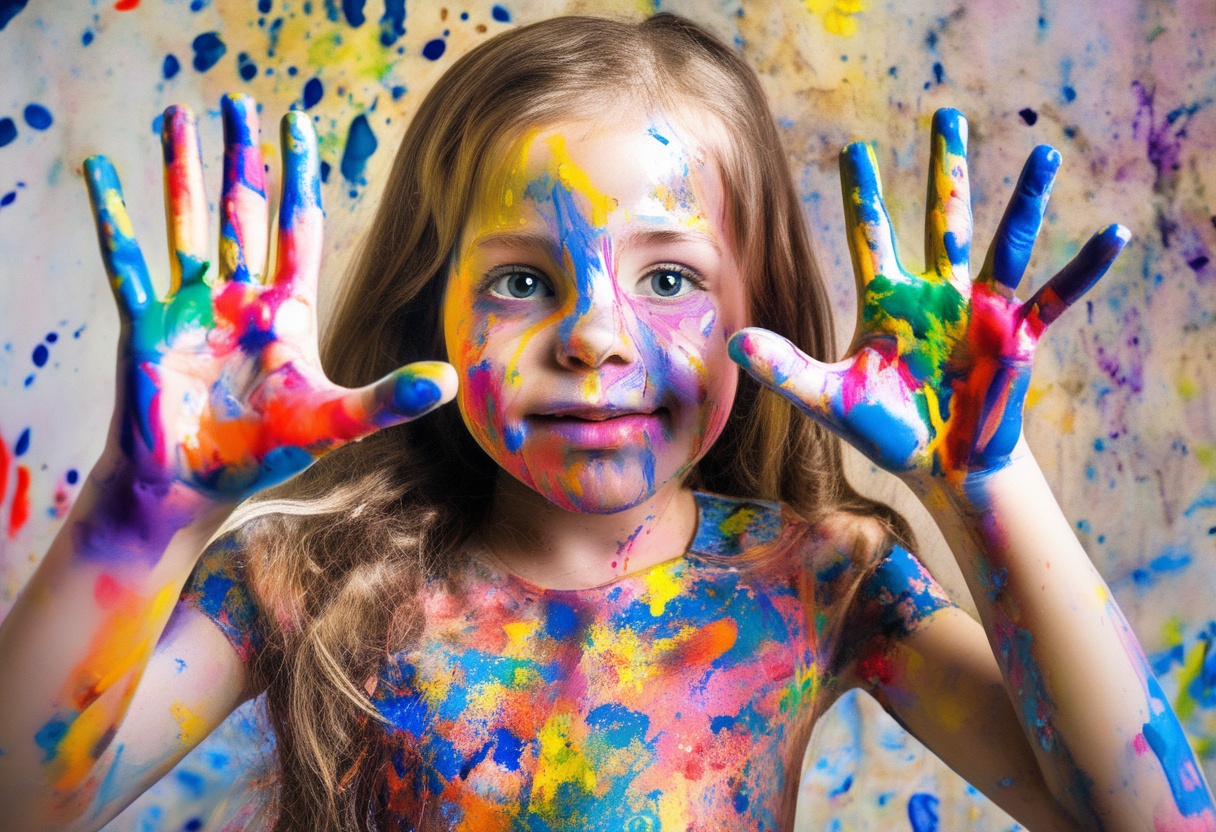} 
        & \includegraphics[width=0.15\textwidth]{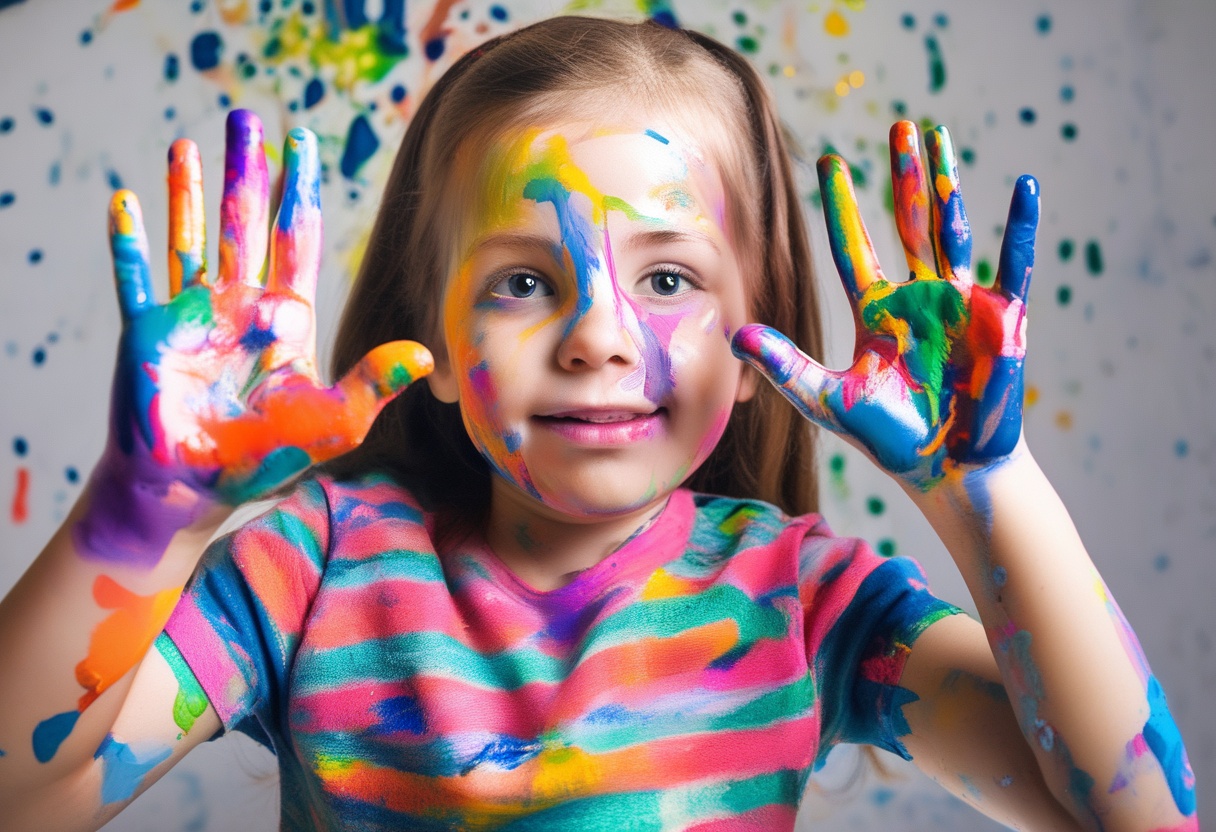}
 \\
 \midrule
 \multicolumn{3}{c|}{\begin{tabular}[c]{@{}c@{}} 8-bit GaLore (2.4GB)\end{tabular}} &\multicolumn{3}{c}{\begin{tabular}[c]{@{}c@{}}\textbf{8-bit COAP (0.5 GB)}\end{tabular}} \\
        \includegraphics[width=0.15\textwidth]{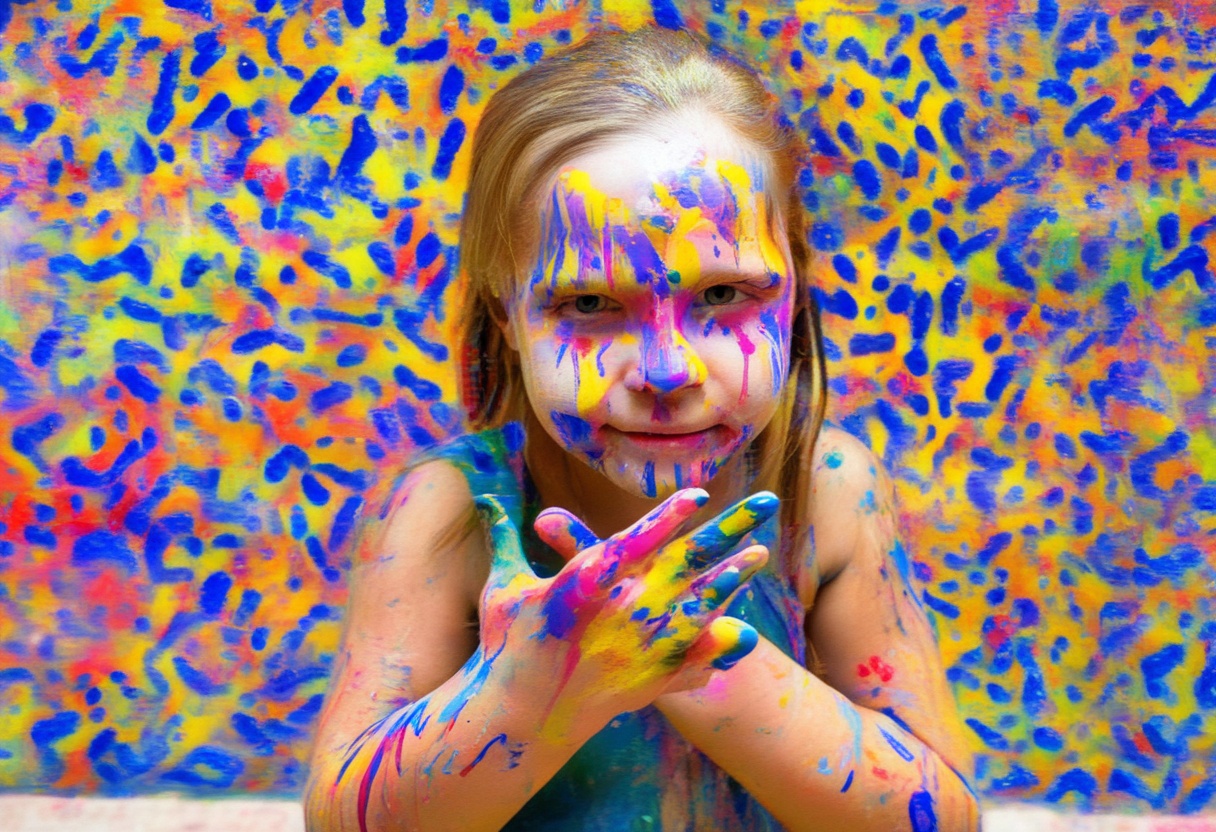} 
        & \includegraphics[width=0.15\textwidth]{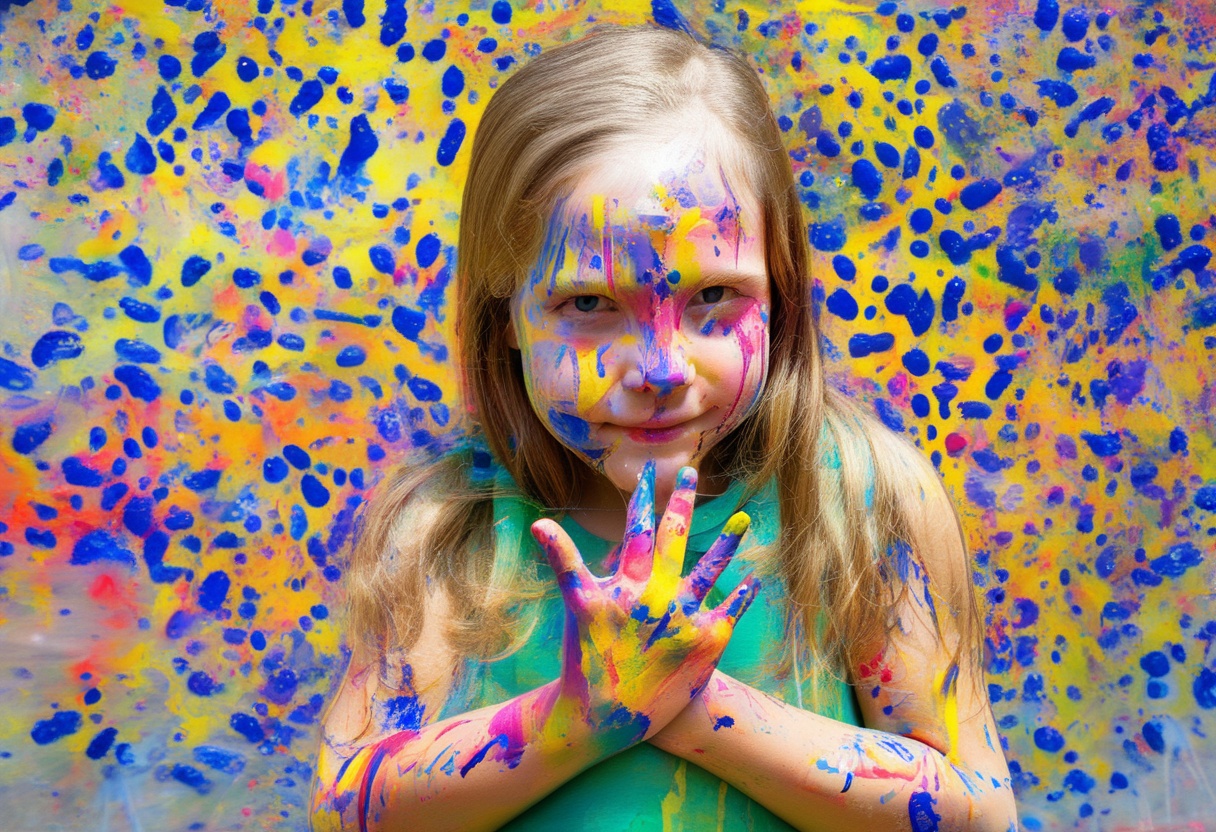} 
        & \includegraphics[width=0.15\textwidth]{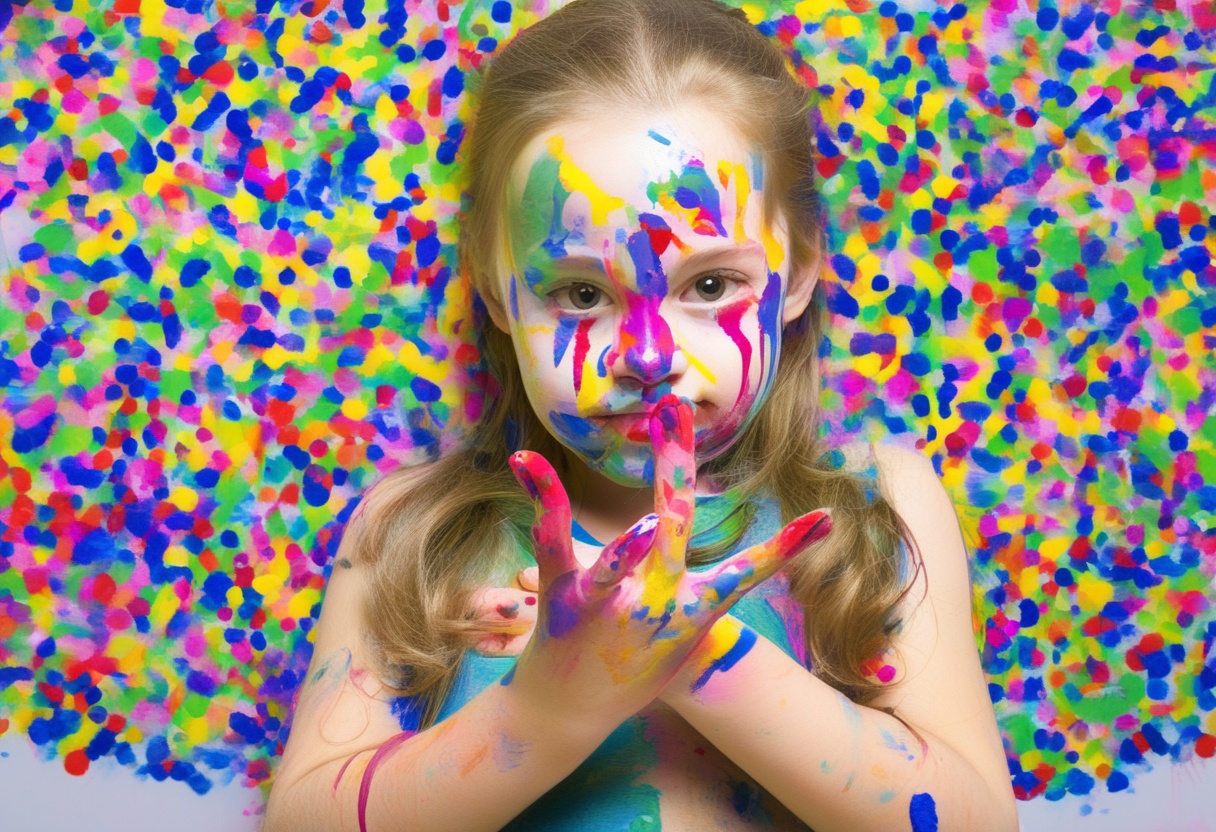}
        &\includegraphics[width=0.15\textwidth]{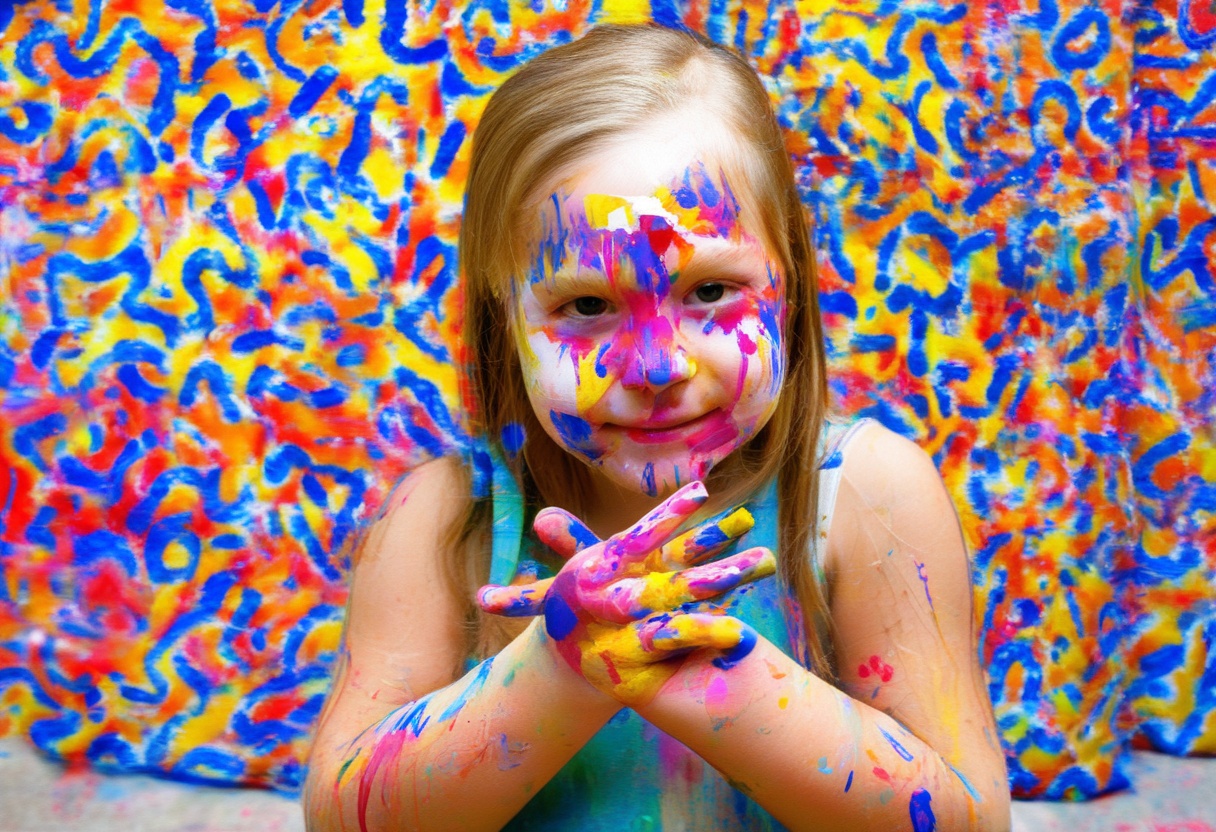} 
        & \includegraphics[width=0.15\textwidth]{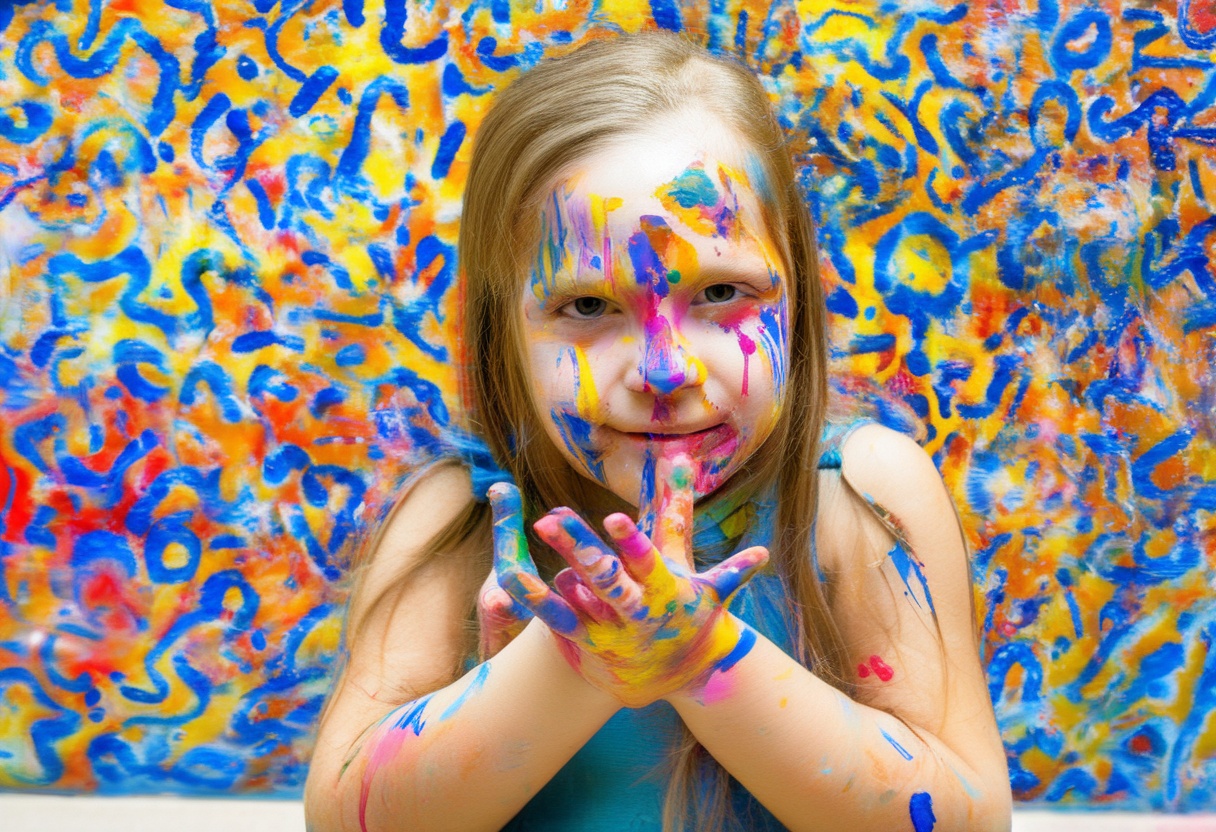} 
        & \includegraphics[width=0.15\textwidth]{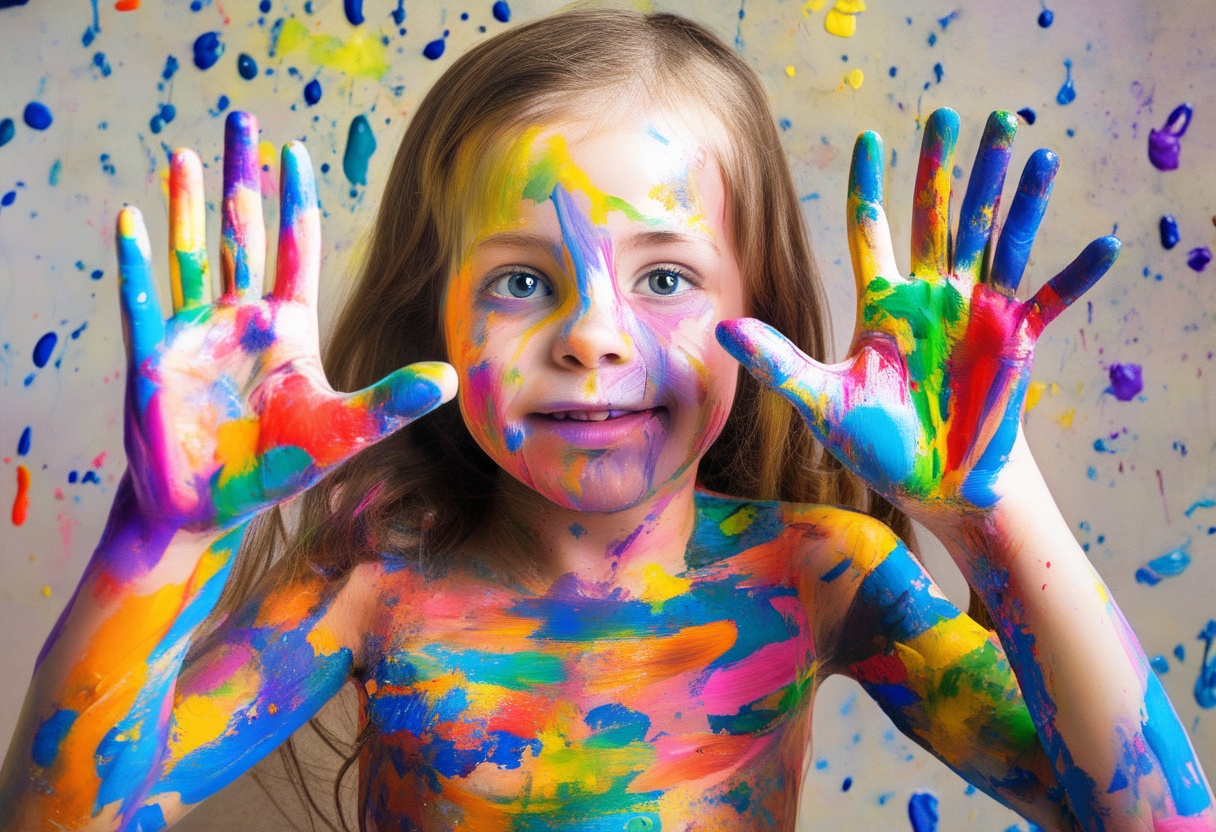} \\
        \bottomrule
    \end{tabular}
\end{table*}

\end{document}